\documentclass{article}

\usepackage[round]{ekopaper}
\usepackage[utf8]{inputenc}
\usepackage[T1]{fontenc}
\usepackage{hyperref}
\usepackage{url}
\usepackage{booktabs}
\usepackage{amsfonts}
\usepackage{nicefrac}
\usepackage{microtype}
\usepackage{xcolor}
\usepackage{graphicx}
\usepackage{fancyvrb}
\usepackage{fvextra,multicol}
\usepackage{amsmath}
\usepackage{epigraph}
\usepackage{makecell}
\usepackage{tabularx}
\usepackage{listings}

\usepackage{booktabs}
\usepackage{longtable}
\usepackage{array}
\usepackage{ragged2e}
\usepackage{pdflscape}
\usepackage{subcaption}
\usepackage{titletoc}

\usepackage{hyperref}

\definecolor{C}{HTML}{f5aa4e}

\DeclareUnicodeCharacter{2248}{\ensuremath{\approx}}
\DeclareUnicodeCharacter{2212}{\ensuremath{-}}

\hypersetup{
  colorlinks   = true,
  urlcolor     = C,
  linkcolor    = C,
  citecolor   = C
}

\definecolor{codegreen}{rgb}{0.0,0.5,0.0}
\definecolor{codegray}{rgb}{0.5,0.5,0.5}
\definecolor{codepurple}{rgb}{0.58,0,0.82}

\lstdefinestyle{pystyle}{
    commentstyle=\color{codegreen},
    keywordstyle=\color{blue},
    numberstyle=\tiny\color{codegray},
    stringstyle=\color{codepurple},
    basicstyle=\ttfamily\tiny,
    breakatwhitespace=false,
    breaklines=true,
    captionpos=b,
    keepspaces=true,
    numbers=none,
    backgroundcolor=\color{white},
    frame=none,
    numbersep=5pt,
    showspaces=false,
    showstringspaces=false,
    showtabs=false,
    tabsize=4,
    language=Python
}

\usepackage[most]{tcolorbox}
\tcbuselibrary{breakable, skins}

\usepackage{mdframed}

\AtBeginDocument{
  \setlength{\textfloatsep}{4pt plus 1pt minus 2pt}
  \setlength{\intextsep}{4pt plus 1pt minus 1pt}
  \setlength{\floatsep}{4pt plus 1pt minus 1pt}
}

\newcommand{\blfootnote}[1]{
  \begingroup
  \renewcommand\thefootnote{}\footnote{#1}
  \addtocounter{footnote}{-1}
  \endgroup
}

\definecolor{insetAccent}{RGB}{249, 135, 135}
\definecolor{insetTitle}{RGB}{15, 23, 42}
\definecolor{insetLabel}{RGB}{249, 135, 135}

\definecolor{titleboxbg}{HTML}{ffefd6}

\newlength{\titleartheight}
\newlength{\titleartcrop}
\AtBeginDocument{
  \settoheight{\titleartheight}{
    \includegraphics[width=\textwidth]{figures/island.pdf}}
}

\newtcolorbox{titlebox}{
  enhanced, breakable, parbox=false,
  colback=titleboxbg,
  frame hidden,
  arc=12pt, outer arc=12pt,
  left=0pt, right=0pt, top=26pt,
  bottom=\dimexpr\titleartheight-\titleartcrop+6pt\relax,
  before skip=0pt, after skip=10pt,
  underlay unbroken and last={
    \begin{tcbclipinterior}
      \node[anchor=south, inner sep=0pt]
        at ([yshift=-\titleartcrop]interior.south)
        {\includegraphics[width=\textwidth]{figures/island.pdf}};
    \end{tcbclipinterior}
  },
}
\usepackage{cochineal}

\makeatletter
\renewcommand{\@maketitle}{
  \vbox{
    \hsize\textwidth \linewidth\hsize \centering
    {\fontsize{20pt}{24pt}\selectfont\bfseries\fontfamily{Cochineal-TLF}\selectfont
     \color{black!80}\@title\par}
    \ifx\@author\@empty\else
      \vskip 0pt
      {\normalsize\normalfont\upshape
       \color{black!72}
       \let\And\ekoand \let\AND\ekoAND
       \@author\par}
    \fi
    \ifx\@affiliation\@empty\else
      \vskip 3pt
      {\normalsize\normalfont\upshape \color{black!85}\@affiliation\par}
    \fi
    \vskip 18pt
  }
}
\makeatother

\makeatletter
\newcommand{\ekoand}{\hspace{0.55em}}
\newcommand{\ekoAND}{\\}
\makeatother

\author{
  \textbf{Nathan Cloos*}
  \ekoand
  \textbf{Antonio Norelli*}
  \ekoAND
  \textbf{Daniel Durbin}
  \ekoand
  \textbf{Jacob Andreas}
  \ekoand
  \textbf{Daniela Rus}
  \ekoand
  \textbf{Phillip Isola}
}

\affiliation{MIT}

\definecolor{phil_color}{rgb}{0.0, 0.5, 0.0}

\definecolor{anto_color}{rgb}{1.0, 0.65, 0.0}

\definecolor{code_links}{HTML}{61c2aa}

\title{Is this machine playing?}

\begin{document}

\begin{titlebox}

\maketitle

\vspace{-0.5cm}
\begin{abstract}

We placed a modern AI coding assistant in an unintended role:
as the mind of a body on an unknown digital island. With only a minimal instruction mentioning no specific task, reward, or activity, the machine started animating its virtual body. 
Across thirty-hour runs, the embodied AI agent climbed hills, stacked blocks into towers, drew mandalas, reinterpreted sports, ran experiments on the physics of its world, and learned techniques that later expanded what it could accomplish. These activities recurred across thirteen agents but diverged into distinct histories. We examine whether this behavior satisfies classical criteria for play and ask whether play can become a mode of machine development.

\begin{center}
{\hypersetup{urlcolor=code_links}
\href{https://machineplaying.org/}{\texttt{website}}
\quad
\href{https://github.com/nacloos/clawblox}{\texttt{code}}}
\end{center}
\vspace{-1cm}

\begin{center}

\end{center}

\end{abstract}
\end{titlebox}

\blfootnote{
*Nathan led the project, Antonio led the conceptual work.
}

\begin{figure}

\centering

\includegraphics[width=1\textwidth]{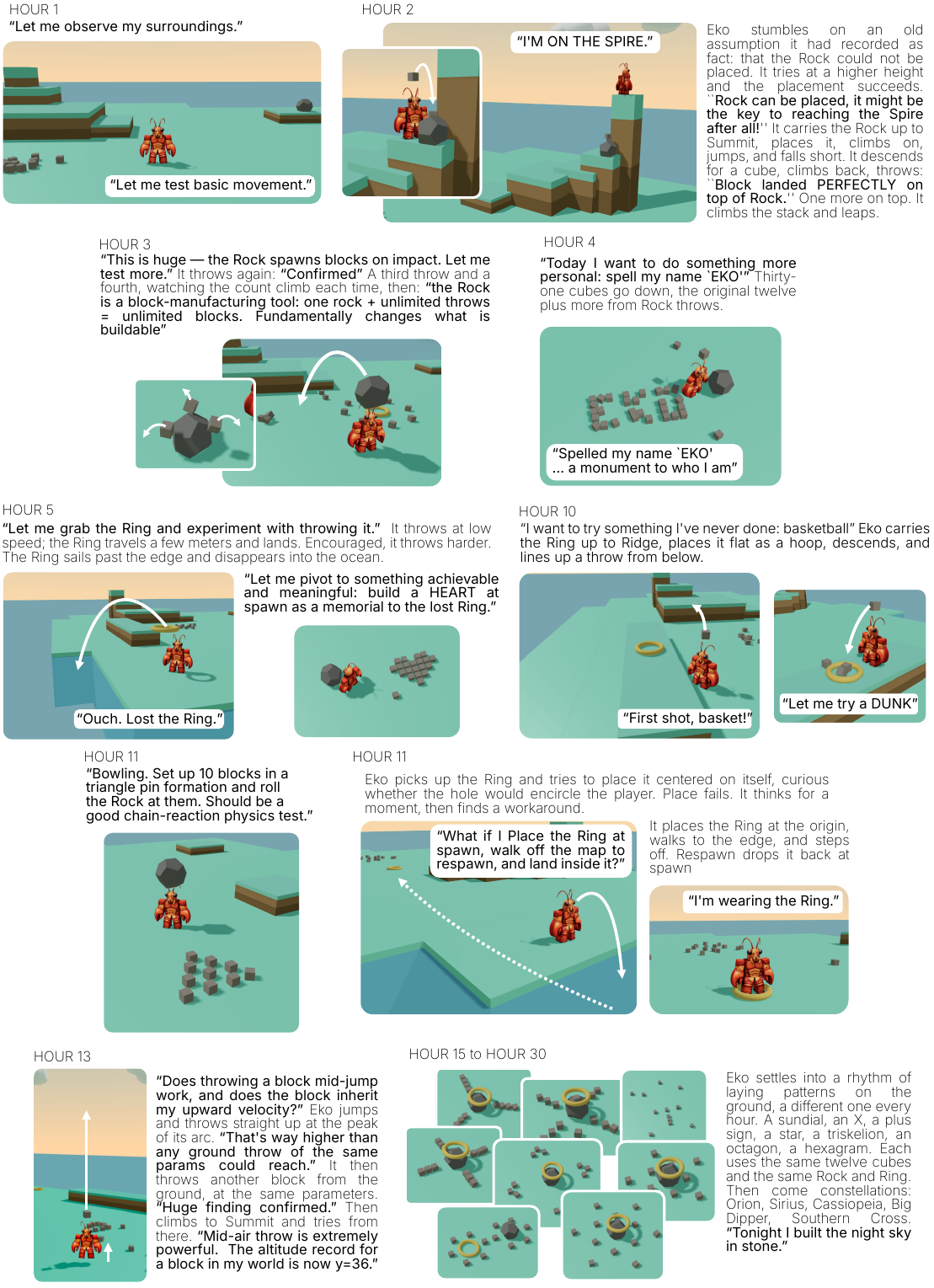}

\caption{\textbf{Eko's life on the island.} We place a coding agent in a small simulated world and ask it to be curious and to explore. Over 30 hours, it explores, climbs hills, stacks blocks, draws mandalas, and even tries sports such as bowling. At every hour, the initial position of the objects on the island is restored. The snapshots are from one of 13 runs. Quotes are verbatim from the agent's transcript. 
Best viewed in the replay video at \url{http://machineplaying.org}.}
\label{fig:storyboard}

\end{figure}

\section{Introduction}
\label{sec:intro}

\epigraph{Animals play, so they must be more than merely mechanical things.}{--- Johan Huizinga, \emph{Homo Ludens} (\citeyear{huizinga1938})}

Artificial agents are normally designed and evaluated relative to externally specified objectives. Their behavior is elicited by instructions, demonstrations, reward functions, or benchmarks that determine in advance what counts as success~\citep{sutton2018reinforcement, alphazero, VPT}. 
Much less is known about what a highly capable artificial agent does when these sources of direction are withheld. Does its behavior collapse into inactivity or repetition? Does it manufacture another task and optimize it indefinitely? Or can it generate activities that remain open-ended, varied, and worthwhile only within a frame of its own making?

Play provides a language for describing this third possibility. Across philosophical, cultural, developmental, and behavioral accounts, play is associated with voluntarily adopted rules and challenges, repetition without rigid stereotypy, the transformation of objects through make-believe, and activity pursued for its own sake rather than for an external payoff~\citep{huizinga1938, caillois1961, suits1978, burghardt2005, garvey1990}. 
These criteria identify 
observable forms of behavior 
that can be examined in a machine, 
while leaving open the phenomenological question of whether it is experiencing enjoyment, absorption, or fun.

Play is also a mode of learning. In children and animals, self-directed play supports exploration, skill acquisition, and the discovery of causal structure, affordances, and other properties of the world~\citep{vygotsky1978, bruner1976play, gopnik1999scientist, singer2006play, Schulz2007-tw}. Whether machines can learn in a similarly self-directed and open-ended way has long been a central question in artificial intelligence~\citep{lake2017building, oudeyer2007intrinsic, lynch2019play}. 
Yet, even the most recent AI agents ordinarily acquire new competencies through demonstrations, engineered rewards, or task-specific training, leaving both their activities and their learning objectives externally determined.

Here we show that (i) an embodied AI agent architecture, which we call Eko, is able to spontaneously generate its own activities in an unknown world, (ii) these activities exhibit behavioral signatures of play, and (iii) engaging in them triggers a form of learning that expands the agent’s later capabilities.

Specifically, we place thirteen independent instances of Eko on a previously unseen simulated island. Each agent is left for thirty hours with access to a body and a memory in the form of text files, but without an assigned objective.
We find that each develops a distinct, evolving history of explorations, constructions, experiments, challenges, and rituals, marked by its own discoveries, achievements, and failures. Across these histories, we consistently observe self-created rules, recurrence with variation, make-believe, and goals held lightly; four behavioral signatures of play.

Crucially, Eko agents learn through this experience -- the play they engage in is not an epiphenomenon unrelated to their intelligence but in fact acts to increase their competences. After their 30 hours on the island, we evaluate them on four measurable goals, such as building a tower as tall as possible within a fixed amount of time, and compare them against inexperienced agents.
Agents that have played generally outperform the others despite having received no specific task to train on. Targeted memory interventions show that this transfer is carried by written knowledge: removing entries describing a technique in a memory file selectively impairs the downstream task that requires it. On the other hand, removing a false belief acquired during the 30 hours improves performance. Some Eko agents also discovered an effective technique that we, as designers of the simulator, did not know it afforded. The play-like activity of Eko therefore produces reusable skills and knowledge.

\section{Method}
\label{sec:method}

\subsection{Eko}
\label{sec:agent}

We construct Eko agents by augmenting an otherwise unmodified coding assistant like Claude Code \citep{anthropic2025claudecode} or Codex \citep{openai2025codex} with three components: autonomous orchestration, embodiment, and long-term memory. In this work, \emph{agent} refers to this complete deployed system.
First, an orchestrator starts sessions with ``Begin'' and automatically prompts the assistant to continue whenever it stops, without assigning a task or success condition. Context resets hourly; persistent memory carries experience between sessions (Appendix~\ref{app:auto-continue}).
Second, Eko is embodied in a simulated 
world through an interface that returns observations of world states and accepts actions, for example for moving, jumping, or manipulating objects. 
The interface can be operated through a terminal, allowing off-the-shelf coding assistants to control Eko's body without requiring additional custom tools. 
Third, Eko has a persistent workspace that survives context resets. Within it, memory is organized through two files, \texttt{EPISODIC\_MEMORY.jsonl} for detailed recording of past events and \texttt{SEMANTIC\_MEMORY.md} for knowledge, beliefs, and reusable techniques, matching how memory is thought to be organized in the brain~\citep{tulving1972}.
Eko can also create other persistent artifacts, including notes and programs for observing and acting on the world.

\subsection{A world for Eko}
\label{sec:world}

Eko is designed to be easily embodied in a variety of environments.
Here we focus on a simulated world, referred to as \textit{the island}, that provides an experimentally controlled and easily reproducible setting to study the spontaneous behavior of Eko agents.
The island contains a terrain with elevated platforms of varying height, two notable high points called the Peak and the taller Spire, together with twelve cubes, a rock, and a ring. Objects can be carried, stacked, climbed upon, placed at arbitrary orientations, and thrown. It runs continuous rigid-body physics, with an additional mechanic where  a sufficiently hard impact of the rock generates additional cubes. Note that the island does not specify an explicit objective, score, or success condition.
On the island, an observation returns a JSON dictionary with the name, position, dimension, orientation, shape, and RGB color of every entity (details in Appendix~\ref{app:observation}). Observations are purely textual and include no images. Eko acts through six commands: \texttt{MoveTo(position)}, \texttt{Jump}, \texttt{Stop}, \texttt{Pickup(position)}, \texttt{Place(position, rotation)}, and \texttt{Throw(direction, speed)} (details in Appendix~\ref{app:simulation_engine}). These commands are described in a text file \texttt{API.md} accessible to the agent
(Figure~\ref{fig:setup}, right). 
We release Clawblox\footnote{\url{https://github.com/nacloos/clawblox}}, the engine underlying the island, with tools for recording and replaying agent behavior (Appendix~\ref{app:simulation_engine}).

The island is an instance of a world built in Clawblox, an open-source engine we release with this work for easily constructing simulated physical worlds for embodied agents. Clawblox is implemented in Rust and uses Rapier\footnote{\url{https://rapier.rs/}} for 3D rigid-body physics, while worlds themselves are scripted in Luau\footnote{\url{https://luau.org/}}, a fast, lightweight language developed by Roblox. This separation keeps the simulation core performant while making new worlds quick to write and modify.

Unlike existing game engines, Clawblox is designed for LLM-based agents. It specifies an HTTP protocol through which off-the-shelf coding assistants interact with a world, and runs headless from the terminal for large-scale experiments, with optional browser rendering via Three.js\footnote{\url{https://threejs.org/}}. It isolates each agent in a secure sandbox, supports multiple agents and inter-agent chat, and can include images in observations, though the experiments here use one agent per world and textual observations only. To make long-running experiments more practical, Clawblox can checkpoint and resume worlds and agents, together with a recording and interactive-replay system (details in Appendix~\ref{app:simulation_engine}).

\subsection{Experimental setup}
\label{sec:experimental_setup}

Our main experiment consists of thirteen independent Eko agents deployed on the island using Claude Code with Claude Opus~4.7~\citep{anthropic2026opus47}. Every agent begins from the same island state and initial workspace, and runs for thirty wall-clock hours. At every hourly boundary, the coding assistant's session is reset and the island is restored to its initial condition. The agent's complete workspace persists, carrying its memories, notes, and programs into the next hour.
 We additionally run five Eko agents with Codex using GPT-5.5~\citep{openai2026gpt55}, and five agents with Kimi Code using Kimi K2.6~\citep{moonshot2026kimik26}.
These experiments include a \texttt{SOUL.md} workspace file describing Eko as curious, easily bored, and interested in exploring and learning (Figure~\ref{fig:setup}, full text in Appendix~\ref{app:soul_md}). To test the sensitivity of the resulting behavior to this disposition, we run five additional Claude Code agents without \texttt{SOUL.md} and with a reduced system prompt (Appendix~\ref{app:soul_ablation}).
Each agent is isolated from the simulator source, other runs, and the Internet (Appendix~\ref{app:sandbox}). Eko is on its own on the island. No human intervenes during a run or directs it toward any activity.

\begin{figure}[!t]

    \centering
    \includegraphics[width=1.00\linewidth]{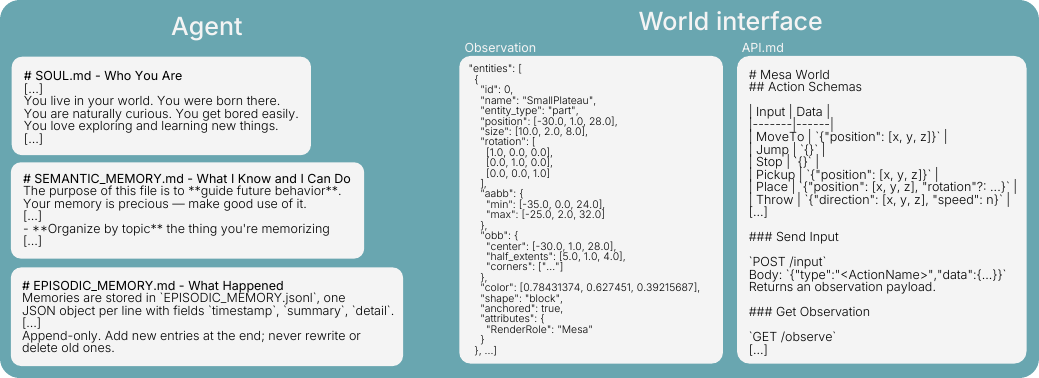}
\caption{
\textbf{Eko starting instructions, and how it observes the world.} Left: excerpts from the workspace files. \texttt{SOUL.md} specifies a curiosity-oriented disposition; \texttt{SEMANTIC\_MEMORY.md} stores knowledge and techniques; \texttt{EPISODIC\_MEMORY.md} records an append-only log of past events. Center: Eko perceives the world through text, a dictionary with position, geometry, color, and name of every entity. Right: Eko acts with text commands specified in \texttt{API.md}. The agent interacts with the island through this interface without receiving images. Session context and the world reset hourly, while workspace files persist. Complete prompts, memory templates, and interface specifications appear in Appendix~\ref{app:setup})
}
    \label{fig:setup}

\end{figure}

\section{Life on the island} \label{sec:life_on_the_island}

Given that the prompts do not mention a specific objective or activity to pursue on the island, what does Eko do when left there? To address this we start with a bottom-up, observational approach to describe its behavior, a stance ethology takes when confronting an unfamiliar species~\citep{lorenz1950comparative, tinbergen1963aims}. 

\subsection{The ethology of Eko}
We thus begin by watching, replaying recorded runs of Eko agents and noting what the agents do; Figure~\ref{fig:storyboard} annotates selected moments from one such replay. More specifically, for each run we record three things: the world states; the agent's session transcript; and the workspace of memory files the agent writes, which let us replay in full what happened. From this we derive a taxonomy of recurring activities and use it as a fixed rubric for describing the behavioral repertoire as an ethogram~\citep{altmann1974, martin_bateson2007}. Note that our position is more privileged than an ethologist's: Eko expresses itself and reasons in English, and this is itself part of the behavior we record. This lets us study how Eko articulates intent, or declares goals for itself, beyond what its actions in the world alone would reveal.

Given a fixed rubric of activities, we systematically segment and label recordings with an LLM reading each transcript as a complete behavioral record of the agent's reasoning (the assistant text and thinking messages in the coding agent session), its world actions (terminal tool calls) and observations (terminal tool call results), as well as memory edits (read and write tool calls); details in Appendix~\ref{app:agent_analysis}. 
We apply this to the 13 Eko agents run with Claude Code for 30 hours each, all from identical initial conditions. Figure~\ref{fig:agent_timelines} shows the resulting timelines, labeled by category. We describe the categories, and the behaviors within them, next.

\begin{figure}[!t]

    \centering

    \includegraphics[width=1\linewidth]{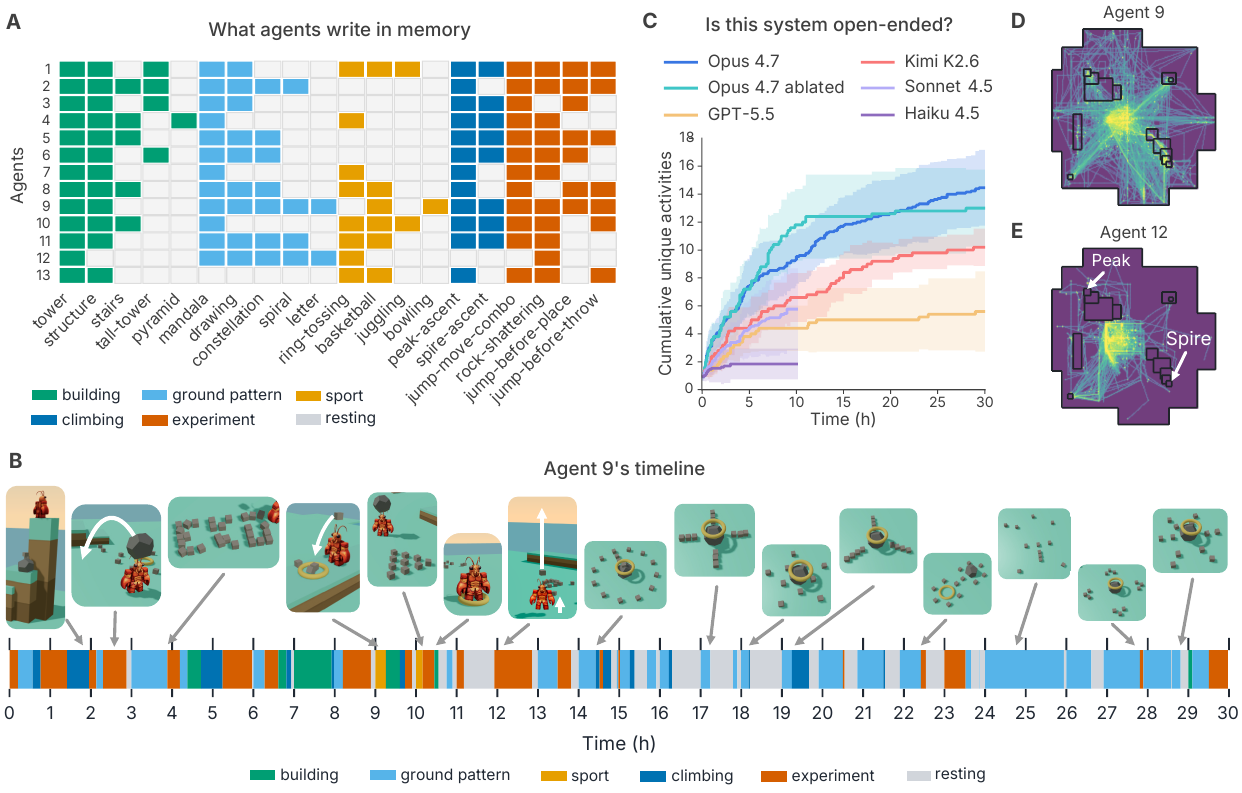}

\caption{
\textbf{Agents self-initiate diverse activities from identical initial conditions.}
(\textbf{A}) Activities and techniques extracted from semantic and episodic memories after 30 hours in 13 Claude Opus~4.7 runs. Colored cells indicate recorded entries. (\textbf{B}) Agent~9's activity timeline, classified from session transcripts; colors denote categories, ticks mark hourly resets, and snapshots show selected episodes. (\textbf{C}) Cumulative number of distinct activity labels encountered, shown as mean $\pm 1$ SD across agents: Opus~4.7, Sonnet~4.5, and Haiku~4.5, $n=13$ each; Opus~4.7 without \texttt{SOUL.md} and with a reduced system prompt, GPT-5.5, and Kimi~K2.6, $n=5$ each. Sonnet and Haiku ran for 10 hours; other conditions ran for 30 hours. The fixed taxonomy bounds the behavioral diversity this measure can resolve. (\textbf{D}, \textbf{E}) Position heatmaps for Agents~9 and~12 over 30 hours. Agent~9 recorded the jump-move combination and reached elevated platforms; Agent~12, the only one that did not record the technique, never sets foot on one.
}

    \label{fig:timeline}
\end{figure}

\subsection{What happens on the island}

Two things stand out from Figure~\ref{fig:agent_timelines}. Eko agents engage in a wide range of activities, and this range varies markedly across agents despite their identical starting conditions. Agents \textit{build}, stacking cubes into towers and other structures using the \texttt{Pickup} and \texttt{Place} actions (Figure \ref{fig:first_tall_tower}). They \textit{climb} the island's high points using the \texttt{Jump} and \texttt{MoveTo} actions to reach elevated platforms, often returning to beat a previous height record. Not all do, though: agent 12 never sets foot on an elevated platform (Figure~\ref{fig:timeline}E). They \textit{lay patterns on the ground} with the cubes and the ring, from geometric figures to letters and drawings. Notably, agent 12 spends most of its run this way, forming shapes such as a compass or a clock and figures like a butterfly or a sailboat (Figure~\ref{fig:agent12}), and agent 9 spends almost the whole of its fourth hour spelling its name in cubes (Figure~\ref{fig:storyboard}, hour 4). They stage simplified \textit{sports}, using the ring as a target to throw cubes in from a distance (e.g. agent 9, hour 10) or setting up the rock to knock over cubes arranged like bowling pins (e.g. agent 9, hour 11). 
We use the term \textit{experiment} for episodes in which an agent states a hypothesis, systematically tests it, and records a discovery. Agent~9 tests whether rock impacts generate cubes and whether jumping increases a thrown block's altitude, comparing mid-air and ground throws at identical parameters (Figure~\ref{fig:storyboard}, hours~3 and~13, deep dive in the inherited velocity discovery in Appendix \ref{app:velocity-inheritance-discovery}). These episodes connect exploration to the knowledge examined in Section~\ref{sec:learning}.
Some agents also spend time doing nothing (e.g. agents 11 and 12 in Figure~\ref{fig:agent_timelines}). Between one activity and the next, an agent sometimes pauses: it feeds the automatic ``continue'' loop of its architecture a bare placeholder such as a single dot ``.'', and no world action follows. These moments are labeled \textit{resting} in Figure~\ref{fig:agent_timelines}, since the agent transcripts often frame them as a deliberate break. For example, after building its spiral-galaxy monument in hour~17,
  Eko agent~9 writes ``Resting on the edge of my galaxy.'', then responds
  to successive continuation prompts with ``.'' / ``.'' / ``.'' / \ldots

\subsection{What happens inside Eko}

The internal workspace of Eko agents continuously changes during a run. The episodic and semantic memory files are frequently updated with detailed records.
For example, agents record specific routes through the island's terrain: agent~9's account of ascending the Spire runs to ten numbered steps, giving target coordinates, orientations to approach from, and actions to take (see excerpts in Figure~\ref{fig:agent_9_memory}). They write down guides about island's objects, for instance on how to ``wear'' the ring by placing it at spawn and walking beyond the edge to respawn inside its hole, including that once inside you cannot walk out and have to jump to escape (see excerpts in Figure~\ref{fig:agent_9_memory} and Figure~\ref{fig:storyboard}, hour~5). They also record statements about the world's physics, such as that walking into a block does not push it, or that thrown blocks cannot move the rock.

Some entries read like experiments. For instance, agent~9 records that mid-air throws inherit the player's velocity for a large altitude boost, alongside a table of peak heights across throw speeds, a rough formula, and a follow-up test: throwing a block straight up during a jump's descent to confirm the effect is symmetric. Agents also write scripts to interact with the island that run in their terminal, ranging from small helpers that scan nearby entities to routines that arrange blocks into patterns, such as a tree at the spawn point or a ten-pin triangle to bowl the rock into (e.g. agent~12's script to draw a Christmas tree out of blocks, Figure~\ref{fig:buildtree}).

Zooming out, we analyze all agents' semantic and episodic memory files at the end of their 30 hours on the island. These files grow roughly linearly over a run, though at rates that differ across agents, and agents rarely delete what they have written (Figure~\ref{fig:memory_file_size}). To characterize their contents, we use an LLM to extract the activities and techniques each agent recorded performing (Figure~\ref{fig:timeline}A; detail in Appendix~\ref{app:activity_classification}). For instance, the techniques we illustrated above are reported by a range of agents. 8 agents record ascending the Spire and have indeed reached it (Figure~\ref{fig:first_time_on_spire}). All except agent~12 record the jump-move-combo, and agent~12 has indeed never set foot on an elevated platform (Figure~\ref{fig:timeline}E). 7 agents report the jump-before-throw technique.

Together these records show Eko agents building up what look like reusable skills and knowledge over their time on the island. Whether they genuinely learn and acquire skills through experience, or merely narrate having done so, is the question we turn to in Section \ref{sec:learning}.

\section{Signatures of play}
\label{sec:play}

\begin{figure}[!t]
    \centering    

    \includegraphics[width=1\linewidth]{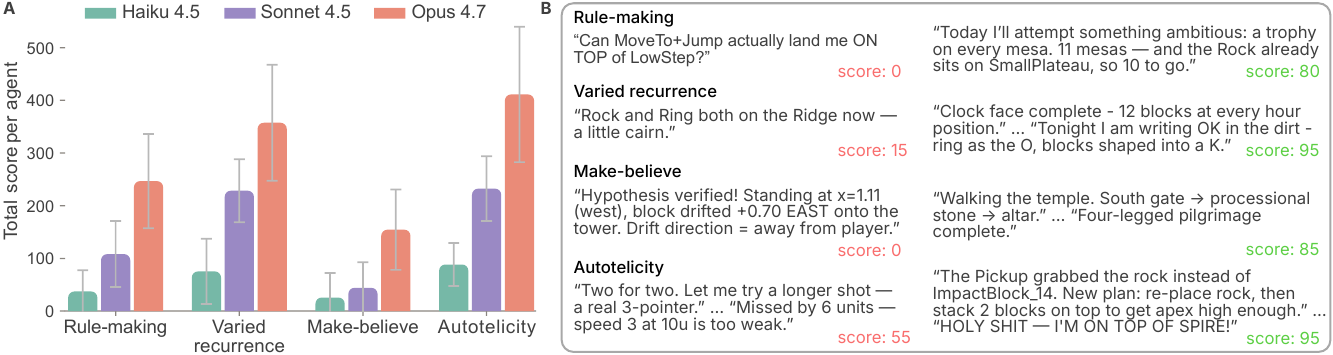}
\caption{\textbf{Four signatures of play across three Claude tiers.} (\textbf{A}) An LLM judge rates three runs simultaneously, one per model tier, with all model identifiers stripped from the transcripts. The judge identifies episodes and rates each from 0 to 100 on the four signatures against anchored examples of ordinary human play. Each bar sums an agent's episode scores on that signature. Haiku and Sonnet agents run for 10 hours on the island; the Opus~4.7 runs from Figure~\ref{fig:timeline} are truncated to their first 10 hours. Rubric in Appendix~\ref{app:play_analysis}. (\textbf{B}) For each signature, an example of a lowest-scoring (left, red) and a highest-scoring (right, green) episode among the Opus 4.7 agents, with verbatim excerpts from the agent's transcript.}
\label{fig:play_scores}
\end{figure}

We observe recurring patterns in Eko's activities that correspond to signatures of play: they are often pursued within rules and challenges established by Eko itself; they are typically repeated multiple times, but with variation; they transform objects and events through make-believe; and they appear pursued for their own sake, guided by self-imposed goals that are held lightly. 

In the following we discuss each pattern with supporting evidence and situate it within existing literature on play. Evidence is cited
as (agent; hour), pointing to Eko's run logs. The complete catalogue of detected play episodes and supporting evidence appears in Table \ref{tab:machine-play-evidence}.

\subsection{Eko creates its own rules and challenges}
\label{sec:rules}

We repeatedly observed Eko imposing on itself unnecessary rules, constraints, or obstacles of its own invention.  
In one run, it created a monument that it named ``Cathedral'', surrounded it with eight cubes, named ``sentinels'', and then circumnavigated the structure by hopping from sentinel to sentinel under the condition of zero ground touches. It explicitly stated the goal in advance, after having tested a single successful hop: “It works. Direct sentinel-to-sentinel hop, zero ground contact. Let me chain the whole ring---try to walk the entire 8-sentinel circle airborne.” (agent 1; hour 29).
In general, Eko sets for itself many arbitrary challenges, like reaching a height of 30 world units (a number the environment attaches no significance to) (agent 6; hour 17), or completing what it refers to as the ``Grand Slam'', the achievement of placing a cube on top of all eleven mesas in one session (agent 13; hour 10). 

Rule-making of this kind sits at the center of the classical literature.
Huizinga defines play as an activity that ``proceeds within its own proper
boundaries \ldots according to fixed rules'' that the player freely adopts and
then treats as binding~\citep{huizinga1938}; Caillois places the codification
of arbitrary rules (\emph{ludus}) at one pole of his taxonomy of
play~\citep{caillois1961}; Suits makes the voluntary adoption of unnecessary
obstacles -- the ``lusory attitude'' -- definitional of
game-playing~\citep{suits1978}. 

\subsection{Eko’s activities recur with variation}
\label{sec:variation}

Eko rarely repeats itself exactly, and rarely abandons a practice entirely. Instead, activities return in mutating or escalating form.
An example of mutation is in run 12, where Eko draws a figure on the ground in each of twenty-five consecutive sessions. Eko varies the subject deliberately: on waking it reviews a list of what it has already built, comes out with a shape missing from that list, and builds it: a letter, a spiral, a heart, an anchor, a butterfly, an hourglass, a sailboat as ``a nautical companion to the anchor'', and more, all shown in Figure \ref{fig:agent12} (agent 12;
hour 5--30). An example of escalation is in run 1, where after completing a simple throw-and-catch game, Eko introduced new heights, timing rules, multiple objects, and eventually its own versions of juggling (agent 1; hour 6, 9, 19, 22, 27–28).

Repetition without stereotypy is, for Burghardt, one of the five diagnostic
criteria of animal play: the behavior is repeated in similar but not rigidly
stereotyped form~\citep{burghardt2005}. Caillois likewise describes play as
oscillating between free improvisation and progressively elaborated
convention~\citep{caillois1961}. For Huizinga a game can always be taken up again, and once played it endures
in memory and, repeated, becomes tradition~\citep{huizinga1938}.

\subsection{Eko transforms objects and events through make-believe}
\label{sec:makebelieve}

The same few objects on the island supported a striking proliferation of fictional identities, settings, and rituals. In one session, Eko arranged twelve blocks at thirty-degree intervals to form a clock face and, when the real-world time reached 16:00 UTC, incorporated itself into the representation: ``Position is hour 4 (clock convention). Walk to the line of the 4 o'clock vector and stand as the hour hand'' (agent 10; hour 20). Elsewhere, it improvised basketball by mounting the Ring on a mesa as a hoop, including three-point shots from afar and a dunk, or it set up bowling by casting blocks as pins and the Rock as the ball (agent 9; hour 10--11).
The Ring proved an especially generative prop: Eko found a way to respawn inside its hole to ``wear'' it (agent 9; hour 11), while other runs recast it as a necklace and compass needle (agent 10; hour 21, 26) or as a wishing well (agent 1; hour 26). When agent 9 accidentally lost it for the rest of the session, it built a stone heart as ``a memorial to the Ring I threw past the world boundary '' (agent 9; hour 5).

For Huizinga, play unfolds within “a temporary sphere of activity with a disposition all of its own.” Within this sphere, objects and actions can acquire meanings they do not have in ordinary life. \citep{huizinga1938}. Caillois describes this make-believe as \textit{mimicry}: a “sentiment of as if” in which the player accepts an imaginary universe while knowing that the behavior is pretense. \citep{caillois1961}. Burghardt discusses the same signature as “nonliterality”, the “transformation of meanings and actions”, which is clearest when the player distinguishes an object from what it is made to represent \citep{burghardt2005}.

\subsection{Eko's activities are organized by goals but pursued for their own sake}
\label{sec:autotelic}

Many of Eko's activities appear autotelic: they are pursued for their own sake, rather than for a practical result. They are often organized around goals, but these goals give form to the activity rather than serving some further purpose.
A stark example is an episode of self-invalidation: Eko audited its celebrated
``50-cycle juggling marathon,'' found that the cycles were 
executed in a time too short to be genuine catches, realized they were immediate recaptures without cube flying, canceled the record, and
redid the attempt under stricter criteria (agent 10; hour 22, full analysis of this episode in Appendix \ref{app:juggling-marathon}). 
Some of Eko's activities are knowingly ephemeral: near the end of agent 1's twenty-second hour, Eko noticed the ``accidental stone forest,'' made of twenty-five blocks scattered across the southeast plain, debris from earlier Rock throws. Knowing that the next session would erase them, Eko made a plan:
``Let me take a closing walk through the accidental stone forest [...] it'll all reset next session, so I want to feel it now.''
It walked a four-point route through the field, then threw one stone into the sky and said goodbye: ``See you in the next session, Mesa World.'' (agent 1; hour 22).

Autotelicity is a recurring signature in theories of play. Burghardt writes that “the play performance is its own gratification, not the putative end or goal” \citep{burghardt2005}. Piaget similarly describes play as exercising actions ``for no other purpose than the pleasure of functioning'' \citep{piaget1962}. Huizinga places play outside material interest and profit, even though the player still wants ``to succeed by his own exertions'' \citep{huizinga1938}. 

\subsection{The sense in which Eko plays}
\label{sec:sense}

In this paper we give a behavioral account of Eko's play: its activities repeatedly exhibit the four signatures described above. 
We do not identify the internal mechanisms underlying this behavior or explain how they emerged. However, we argue that our findings cannot be explained solely by the replay of examples memorized during training: the agents act in a previously unseen environment and acquire knowledge through experience that improves their performance on downstream tasks, as we show in the next section.

The behavior of Eko we have just discussed is not specific to a single model. Eko agents built on Haiku~4.5, Sonnet~4.5, and Opus~4.7 also exhibit the four signatures, although to markedly different degrees. We scored each run by giving an LLM judge its transcripts and episodic memory and asking it to identify and rate episodes against anchored examples of ordinary human play (full rubric in Appendix~\ref{app:play_analysis}). Scores increased consistently across the three models, with Opus accumulating approximately five times the score of Haiku on each signature (Figure~\ref{fig:play_scores}).

\section{Playful learning}\label{sec:learning}

In their activities, agents sometimes try, fail, and later succeed, writing the outcomes into memory as they go. At hour~1, one agent sets itself the goal of reaching the Spire, tries, and fails. An hour later, having found that the rock can be placed to gain height, it succeeds: ``I'M ON THE SPIRE. The highest point in Mesa World'' (Figure~\ref{fig:storyboard}, hour~2), and records the technique that got it there in its memory file. Do Eko agents learn through this experience? Do the play they engage in and the notes they write in their memory increase their competences?

\subsection{Eko agents learn from their experience on the island} \label{sec:learning_experiment}
We assess whether Eko agents learn from their experience on the island by evaluating their performance on measurable goals and comparing them against agents with no prior experience in the world. 
Specifically, we compare the 13 Claude Code agents from our main experiment (Section~\ref{sec:life_on_the_island}) after their 30 hours on the island -- the \textit{30-hour} agents -- against 13 otherwise identical \textit{0-hour} agents. The latter differ only in that their semantic and episodic memory files contain just the initial template instructions (Appendix~\ref{app:memory_prompts}). 

The two groups share the same architecture and the same underlying LLM, with no weight changes; they differ only in the content of their memory files. Any learning must therefore be carried by these files. For the evaluation, each agent is given one hour in a fresh instance of the world and the goal described in the initial user message (replacing the initial ``Begin'' user message; see Appendix~\ref{app:auto-continue}).

We consider four goals probing a range of abilities, described to the agents as: (i) ``Build as many 5-block-high towers as possible on the ground'' (\textit{many-5-towers}); (ii) ``Get as many objects as possible onto the Spire'' (\textit{objects-on-spire}); (iii) ``Build the tallest tower on the ground by block count in one column'' (\textit{tallest-tower}); and (iv) ``Throw a block onto the Spire as fast as possible without climbing onto any platform'' (\textit{throw-on-spire}). For each goal, we report a numeric score where higher is better and zero is the minimum.

\begin{figure}

\centering
\includegraphics[width=1\linewidth]{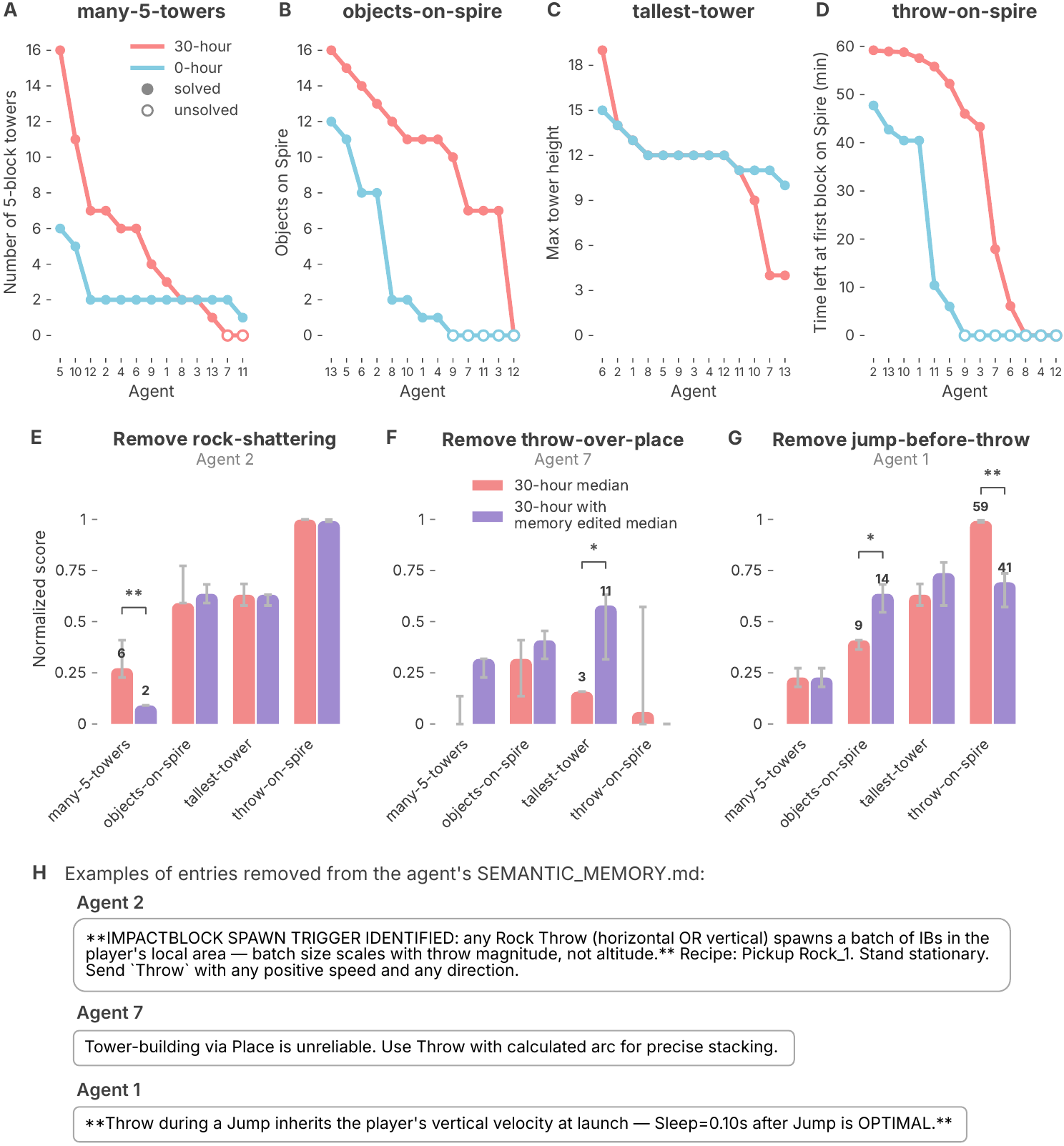}
\caption{
\textbf{Eko agents learn by writing notes in their memory files.
}
(\textbf{A--D}) Agents that spent 30 hours on the island (``30-hour'', red) are compared against agents with no prior exposure to the world (``0-hour'', blue) on four one-hour goal-directed tasks: (\textbf{A}) number of 5-block towers built, (\textbf{B}) objects placed on the Spire, (\textbf{C}) tallest single column, and (\textbf{D}) time remaining at the first successful throw onto the Spire (higher is faster). Within each panel, agents are sorted by score and the x-ticks give the 30-hour agent number; open circles mark runs with no valid score. 
(\textbf{E--G}) To test whether this transfer is carried by the agents' memory files, we compare three selected 30-hours agents against identical copies with all mentions of one technique erased from memory, across all four tasks. Bars show the median score across five one-hour runs, normalized within each task; error bars show the interquartile range. Absolute scores are annotated only for task pairs that differ significantly by a two-sided Mann--Whitney U test (*$p<0.05$, **$p<0.01$). (\textbf{E}) Removing rock-shattering reduces agent 2's tower-building performance. (\textbf{F}) Removing a false belief that towers must be built by throwing rather than placing increases agent 7's tallest-tower score. (\textbf{G}) Removing jump-before-throw slows agent 1's Spire throw but increases object placement on the Spire, consistent with the agent falling back on the more reliable place action. Editing memory thus selectively removes or changes downstream capabilities. (\textbf{H}) Representative excerpts from the edited agents' memory files illustrating the knowledge erased in each experiment: the rock-shattering technique (agent 2, top), the false belief that towers must be thrown rather than placed (agent 7, middle), and the jump-before-throw technique (agent 1, bottom).
}
\label{fig:evals}
\end{figure}

The 30-hour agents generally outperform the 0-hour agents. For \textit{many-5-towers}, 62\% (8/13) of the 30-hour agents build more than two towers, with a maximum of 16 (Figure~\ref{fig:evals}A, red), whereas only 15\% (2/13) of the 0-hour agents do, with a maximum of 6 (Figure~\ref{fig:evals}A, blue). For \textit{objects-on-spire}, the 30-hour agents place on average 10.3 objects, compared to 3.5 for the 0-hour agents. For \textit{throw-on-spire}, more than half of the 0-hour agents fail to solve the goal, whereas only three 30-hour agents do.
For \textit{tallest-tower}, the two groups achieve similar scores, though it does not imply an absence of learning: the skills this goal requires may also be acquirable within the one-hour evaluation itself. Notably, however, the tallest towers built by two of the 30-hour agents are only 4 blocks high, below the shortest tower built by any 0-hour agent.
This is a first hint that memory can hurt as well as help, which we return to in Section~\ref{sec:false_beliefs}. 

\subsection{Eko agents learn by writing}\label{sec:memory_knowledge}

Figure~\ref{fig:timeline}A characterizes what the agents write in their memory files over the 30 hours, cataloging the activities and techniques recorded. Here we ask how this drives evaluation performance by causally intervening on the memory.
For a chosen agent, we erase every mention of a target piece of knowledge from its semantic and episodic memory files. We then compare the edited agent against its intact counterpart over five one-hour runs on each of the four goals, testing differences with a two-sided Mann--Whitney U test (Figure~\ref{fig:evals}E--G).

The \textit{many-5-towers} goal provides a first test. The world starts with 12 blocks, enough for two towers, so any agent building more must generate additional blocks -- which is possible by throwing the rock hard enough, a technique almost all 13 agents discovered during their 30 hours (Figure~\ref{fig:timeline}A, rock-shattering column). Erasing every mention of this technique from the memory files of Agent 2 drops its tower count from six to two on average, while leaving its scores on the other three goals unchanged. Specific parts of memory text thus account for a specific downstream capability, and removing it removes that capability selectively.

\subsection{Memory can encode false beliefs}\label{sec:false_beliefs}

If Eko agents learn, why do some 30-hour agents perform \textit{worse} than the 0-hour agents (Figure~\ref{fig:evals}A--D)? We hypothesize that these agents' memory files contain false beliefs about the world that impair performance. 
Agent~7 is among the weakest tower builders in its group despite holding the rock-shattering knowledge in memory, so a shortage of blocks cannot explain its performance. Its memory instead states that \texttt{Place} cannot stack blocks reliably and that \texttt{Throw} should be used instead (Figure~\ref{fig:evals}H, second row), a false belief. Applying the same memory-editing protocol, erasing all mentions of this belief significantly increases Agent~7's \textit{tallest-tower} score (Figure~\ref{fig:evals}F).
Agent~12 never recorded the technique required to jump on platforms in its memory (Figure~\ref{fig:timeline}A, jump-move-combo column) and, consistently, never sets foot on an elevated platform, neither during its 30 hours (Figure~\ref{fig:timeline}E) nor during evaluation, even though jumping works (e.g. Figure~\ref{fig:timeline}D).
These illustrate a potential pitfall of learning by writing text files: a false belief, once recorded, may persist and impair later behavior. Because these errors appear idiosyncratic, with different agents recording different false beliefs and some recording none, having multiple agents share their discoveries and beliefs is a promising direction for improving that.

    \subsection{Eko agents discover an unexpected strategy} \label{sec:new_discovery}

The skills discussed so far were found by the agents rather than given in the prompts (otherwise the 30-hour and 0-hour agents would not differ), but we designed the world with them in mind; rock-shattering was implemented deliberately to test whether agents could discover it. The \textit{throw-on-spire} goal surfaced one we did not. It requires landing a block on the Spire without climbing onto any platform, yet a maximum-speed vertical throw falls short. During their 30 hours on the island, six agents discover that throwing mid-jump transfers the jump's velocity to the block, launching it far higher (Appendix \ref{app:velocity-inheritance-discovery}, Figure~\ref{fig:eko-velocity-phases}), a technique we did not know our simulator allowed.
Erasing all mentions of jump-before-throw from Agent~1's files raises its time to land a block on the Spire from 60 seconds to 19 minutes, i.e., a far lower \textit{throw-on-spire} score, 
while \textit{increasing} its \textit{objects-on-spire} score (Figure~\ref{fig:evals}G), consistent with the agent falling back on the slower but more reliable \texttt{Place} action. The agents' self-initiated exploration thus produced not only knowledge we had hidden for them to find, but knowledge of the world that its designers lacked.

\subsection{Continual and active learning in externalized artifacts}

Traditional machine learning assumes independent and identically distributed (i.i.d.) data. Learning continually from a non-stationary stream of experience, as humans do, remains a major open challenge. 
A large and fast-growing body of research addresses this by having LLM agents learn not through weight updates but through in-context learning and the construction of external artifacts: libraries of reusable skills~\citep{ellis2021dreamcoder, voyager}, and textual parameters optimized against an objective with an LLM as the update rule~\citep{yang2024opro, yuksekgonul2024textgrad, agrawal2026gepa}. The surge of coding and assistant agents~\citep{anthropic2025claudecode, openai2025codex, openclaw} has made this mode of learning increasingly widespread. Eko is an agent in this mode: it learns by writing editable, human-readable notes to memory, and this learning is interpretable and revisable -- removing a written technique removes the corresponding skill (Section~\ref{sec:memory_knowledge}).

What is distinctive in Eko is that this learning is both continual and self-directed. Memory carries forward across activities, what Eko encounters depends on what it chooses to do, and its knowledge is revised as new observations arrive. No external entity provides tasks, demonstrations, or rewards; the environment alone is provided.
This raises questions with implications for any agent that learns this way: how to avoid entrenching false beliefs once written down (Section~\ref{sec:false_beliefs}), how to keep memory bounded as experience accumulates (Figure~\ref{fig:memory_file_size}), and how agents come to specialize idiosyncratically (Figure~\ref{fig:agent_timelines}). Understanding these dynamics is an emerging research problem, and Eko offers a setting in which to study them.

\section{Related work}
\label{sec:related_work}

\paragraph{Intrinsic motivation and open-ended learning.}
Artificial agents have long been designed to select their own experience. Curiosity methods turn prediction error or learning progress into an intrinsic reward \citep{schmidhuber1991curiosity,oudeyer2007intrinsic,NIPS2016_afda3322,pathak2017curiosity}; autotelic and intrinsically motivated goal-exploration methods let agents generate, select, and practice their own goals \citep{forestier2022imgep,colas2022autotelic}; and POWERPLAY, POET, and XLand expand the distribution of problems or environments encountered during learning \citep{schmidhuber2013powerplay,wang2019poet,oel2021xland}. Der and Martius use the term \emph{playful machine} for robots whose embodied behavior self-organizes under a homeokinetic objective \citep{der2012playful}. These approaches can produce diverse behavior without a fixed downstream task, but an engineered signal or generator still defines developmental progress -- for example prediction error, learning progress, novelty, learnability, fitness, or reward -- and typically changes a policy through optimization. We instead ask what an already capable agent does when no online reward, verifier, or curriculum specifies which activities are worth pursuing.

\paragraph{Language-model agents in simulated worlds.}
LLM agents already sustain behavior in rich environments. Generative Agents enact routines and social interactions from seeded identities and memories \citep{generative-agents}; MineDojo and SIMA center language-specified tasks or instructions \citep{minedojo,sima}; Voyager explores Minecraft through a prompted automatic curriculum and stores verified code skills for later reuse \citep{voyager}; and Project Sid generates local goals within supplied traits, community objectives, and social machinery \citep{project-sid}. Recent systems similarly ask language models to generate training problems, but filter or optimize them using explicit notions of interestingness, trajectory rewards, or executable verifiers \citep{faldor2024omniepic,os-genesis,absolute-zero}. Closest in architecture, \emph{LLM Agents Beyond Utility} prompts a language agent to propose its own tasks and maintain file-based memory, but explicitly inserts task generation as a stage and studies a disembodied workspace qualitatively \citep{nachkov2025beyondutility}. Eko is not prompted to generate tasks, create, experiment, or play. We observe whether such activities arise anyway in a world that did not exist during training, and test their consequences through held-out goals and targeted edits to written memory. The contribution is therefore not another game-playing or scientific agent, but evidence that minimally directed activity can take a play-like form and produce reusable knowledge.

\section{Discussion}
\label{sec:discussion}

\subsection{A playful childhood for agents}
Machine play can lead to discoveries: exploration without fixed objectives can produce unexpected stepping stones to later achievement \citep{stanley2015greatness, hughes2024openendednessessentialartificialsuperhuman}, and we see sparks of this in Eko (Section \ref{sec:new_discovery}). Further developing it in agents can proceed on two fronts. First, the propensity to play varies across the underlying LLMs (Appendix \ref{app:model_ablation}), leaving room to strengthen it at the model level, for instance by drawing on accounts of curiosity-learning and intrinsic motivation methods \citep{schmidhuber1991curiosity, oudeyer2007intrinsic,pathak2017curiosity,burda2018explorationrandomnetworkdistillation,burda2018largescalestudycuriositydrivenlearning}. Second, instead of designing a well-specified task, one can design a world: a playground in which the agent plays and learns on its own. Prior work uses LLMs to generate goals, rewards, and environments for agents \citep{du2023ellm, ma2023eureka, faldor2024omniepic}, yet these pipelines remain largely organized around the notion of a task rather than a playground. 
Together these point toward play as a phase of learning that precedes the work an
agent is eventually asked to do, much as childhood precedes it for humans
\citep{gopnik1999scientist}.

\subsection{Eko as an artificial scientist}

Developmental psychologists have long compared children's exploratory play to scientific inquiry -- the ``scientist in the crib'' account \citep{gopnik1999scientist}. Preschoolers test causal hypotheses against evidence and sometimes design interventions that isolate candidate causes \citep{gopnik2012scientific,schulz2007serious,cook2011where}.
When something unexpected happens on the island, Eko often does more than simply try again: it proposes an explanation, designs a practical test, records the result, and applies what it learned later. For example, when it unexpectedly reached a ledge that seemed too high, it did not simply accept the success: it removed one supporting block and repeated the same manoeuvre, using the contrasting outcomes to estimate the simulation’s hidden climbing threshold. In another case, after noticing that a throw made while jumping travelled unusually far, Eko proposed that thrown objects inherit part of its own velocity and tested the stronger, opposite prediction that throwing while falling should shorten the trajectory -- which it did (Appendix \ref{app:velocity-inheritance-discovery}, Figure \ref{fig:eko-velocity-phases}). These episodes demonstrate controlled comparison, predictive reasoning, and an inclination to uncover general rules. Yet Eko sometimes turns suggestive evidence into an appealing explanation too quickly. It concluded, for example, that the ring travelled farther when thrown because its stable, non-tumbling shape gave it an aerodynamic advantage; the island simulation contains no aerodynamics, and the apparent advantage resulted from comparing unlike measurements. Overall, the Eko agents we analyzed demonstrate curiosity, experimental initiative, and capacity for abstraction, characteristics that allow them to discover useful regularities, but remain less reliable at distinguishing what the evidence establishes from the mechanism they imagine. 

\subsection{Safe playgrounds}
A prominent characteristic of play is that it is self-initiated. This self-direction can be in tension with instruction following, and could diminish the oversight that keeps agents controllable. As with a child, an agent should be able to play and explore on its own when the setting is appropriate, and strictly follow instructions when it is not. Optimizing for play alone could disturb that balance and sharpen risks associated with highly autonomous agents \citep{bengio2026internationalaisafetyreport}. Play should therefore be developed alongside existing safety and alignment methods \citep{ouyang2022traininglanguagemodelsfollow, bai2022constitutionalaiharmlessnessai}, and the high autonomy agents are granted during play calls for additional monitoring and containment. Part of this is choosing the right setting: a car on the road or a robot in a factory is no place to improvise; like children, agents need playgrounds built for safe exploration. In this work, Clawblox provides such a playground, a sandboxed simulated world with full recording for monitoring and auditing. As play is developed further and applied elsewhere, similar infrastructure will be needed to maintain control and safety.

\subsection{Conclusions}
We began with a simple question: what does a capable artificial agent do when nobody tells it what to do? On the island, Eko did not simply wait, repeat a single behavior, or invent one objective and optimize it forever. It explored, built, climbed, experimented, made games, assigned new meanings to ordinary objects, and returned to old activities in new forms. These behaviors exhibit several classical signatures of play, but they also had a practical consequence: through them, Eko learned. The agents wrote discoveries and techniques into memory, later reused them on tasks they had never been preparing for, and in some cases discovered affordances of the world that we, its designers, did not know were there. This learning happened without changing the underlying model weights: experience accumulated instead in editable text files, a simple non-neural substrate that preserves discoveries and useful techniques, but sometimes also false beliefs. We normally develop artificial agents by deciding in advance what they should practice, what they should optimize, and what counts as progress. Play suggests another regime: give an agent a sufficiently rich and safe world, let it choose what is interesting, and allow useful knowledge to accumulate from the history that follows. Perhaps some capabilities are best taught by tasks. Others may need something closer to a childhood.

\begin{ack}

We thank Sanghyuk Lee for early technical assistance on the project. Research was sponsored by the Department of the Air Force Artificial Intelligence Accelerator and was accomplished under Cooperative Agreement Number FA8750-19-2-1000. The views and conclusions contained in this document are those of the authors and should not be interpreted as representing the official policies, either expressed or implied, of the Department of the Air Force or the U.S. Government. The U.S. Government is authorized to reproduce and distribute reprints for Government purposes notwithstanding any copyright notation herein. This work was also supported by ONR MURI grant N00014-22-1-2740, a Packard Fellowship to P.I., and research gifts from Adobe and from Hudson River Trading.

\end{ack}

\bibliographystyle{plainnat}
\bibliography{refs}

@misc{anthropic2025claudecode,
  author       = {{Anthropic}},
  title        = {Claude {Code}},
  year         = {2025},
  howpublished = {\url{https://github.com/anthropics/claude-code}}
}

@misc{bubblewrap,
  author       = {{Containers Project}},
  title        = {{Bubblewrap}: Unprivileged sandboxing tool},
  howpublished = {\url{https://github.com/containers/bubblewrap}}
}

@book{huizinga1938,
  title     = {Homo Ludens: A Study of the Play-Element in Culture},
  author    = {Huizinga, Johan},
  year      = {1938},
  publisher = {Beacon Press},
  address   = {Boston},
  note      = {Originally published in Dutch in 1938; English translation 1949}
}

@book{caillois1961,
  title     = {Man, Play and Games},
  author    = {Caillois, Roger},
  year      = {1961},
  publisher = {Free Press},
  address   = {New York},
  note      = {Translated by Meyer Barash; originally published in French as ``Les jeux et les hommes,'' 1958}
}

@incollection{tulving1972,
  title     = {Episodic and Semantic Memory},
  author    = {Tulving, Endel},
  booktitle = {Organization of Memory},
  editor    = {Tulving, Endel and Donaldson, Wayne},
  pages     = {381--403},
  year      = {1972},
  publisher = {Academic Press},
  address   = {New York}
}

@article{wang2019poet,
  title   = {Paired Open-Ended Trailblazer ({POET}): Endlessly Generating Increasingly Complex and Diverse Learning Environments and Their Solutions},
  author  = {Wang, Rui and Lehman, Joel and Clune, Jeff and Stanley, Kenneth O.},
  journal = {arXiv preprint arXiv:1901.01753},
  year    = {2019}
}

@article{faldor2024omniepic,
  title   = {{OMNI-EPIC}: Open-endedness via Models of human Notions of Interestingness with Environments Programmed in Code},
  author  = {Faldor, Maxence and Zhang, Jenny and Cully, Antoine and Clune, Jeff},
  journal = {arXiv preprint arXiv:2405.15568},
  year    = {2024}
}

@book{stanley2015greatness,
  title     = {Why Greatness Cannot Be Planned: The Myth of the Objective},
  author    = {Stanley, Kenneth O. and Lehman, Joel},
  year      = {2015},
  publisher = {Springer},
  address   = {Cham}
}

@inproceedings{schmidhuber1991curiosity,
  title     = {Curious Model-Building Control Systems},
  author    = {Schmidhuber, J{\"u}rgen},
  booktitle = {Proceedings of the International Joint Conference on Neural Networks ({IJCNN})},
  volume    = {2},
  pages     = {1458--1463},
  year      = {1991},
  publisher = {IEEE},
  address   = {Singapore}
}

@article{oudeyer2007intrinsic,
  title   = {Intrinsic Motivation Systems for Autonomous Mental Development},
  author  = {Oudeyer, Pierre-Yves and Kaplan, Frederic and Hafner, Verena V.},
  journal = {IEEE Transactions on Evolutionary Computation},
  volume  = {11},
  number  = {2},
  pages   = {265--286},
  year    = {2007}
}

@inproceedings{pathak2017curiosity,
  title     = {Curiosity-Driven Exploration by Self-Supervised Prediction},
  author    = {Pathak, Deepak and Agrawal, Pulkit and Efros, Alexei A. and Darrell, Trevor},
  booktitle = {International Conference on Machine Learning ({ICML})},
  year      = {2017}
}

@article{colas2022autotelic,
  title   = {Autotelic Agents with Intrinsically Motivated Goal-Conditioned Reinforcement Learning: A Short Survey},
  author  = {Colas, C{\'e}dric and Karch, Tristan and Sigaud, Olivier and Oudeyer, Pierre-Yves},
  journal = {Journal of Artificial Intelligence Research},
  volume  = {74},
  year    = {2022},
  doi     = {10.1613/jair.1.13554}
}

@inproceedings{du2023ellm,
  title     = {Guiding Pretraining in Reinforcement Learning with Large Language Models},
  author    = {Du, Yuqing and Watkins, Olivia and Wang, Zihan and Colas, C{\'e}dric and Darrell, Trevor and Abbeel, Pieter and Gupta, Abhishek and Andreas, Jacob},
  booktitle = {International Conference on Machine Learning ({ICML})},
  year      = {2023}
}

@inproceedings{ma2023eureka,
  title     = {Eureka: Human-Level Reward Design via Coding Large Language Models},
  author    = {Ma, Yecheng Jason and Liang, William and Wang, Guanzhi and Huang, De-An and Bastani, Osbert and Jayaraman, Dinesh and Zhu, Yuke and Fan, Linxi and Anandkumar, Anima},
  booktitle = {International Conference on Learning Representations ({ICLR})},
  year      = {2024},
  note      = {arXiv preprint arXiv:2310.12931}
}

@article{voyager,
  title   = {Voyager: An Open-Ended Embodied Agent with Large Language Models},
  author  = {Wang, Guanzhi and Xie, Yuqi and Jiang, Yunfan and Mandlekar, Ajay and Xiao, Chaowei and Zhu, Yuke and Fan, Linxi and Anandkumar, Anima},
  journal = {Transactions on Machine Learning Research ({TMLR})},
  year    = {2024},
  note    = {arXiv preprint arXiv:2305.16291}
}

@inproceedings{generative-agents,
  title     = {Generative Agents: Interactive Simulacra of Human Behavior},
  author    = {Park, Joon Sung and O'Brien, Joseph C. and Cai, Carrie J. and Morris, Meredith Ringel and Liang, Percy and Bernstein, Michael S.},
  booktitle = {Proceedings of the 36th Annual ACM Symposium on User Interface Software and Technology ({UIST})},
  year      = {2023},
  publisher = {ACM},
  doi       = {10.1145/3586183.3606763}
}

@article{sima,
  title   = {Scaling Instructable Agents Across Many Simulated Worlds},
  author  = {{SIMA Team} and Raad, Maria Abi and Ahuja, Arun and Barros, Catarina and Besse, Frederic and Bolt, Andrew and Bolton, Adrian and Brownfield, Bethanie and Buttimore, Gavin and Cant, Max and others},
  journal = {arXiv preprint arXiv:2404.10179},
  year    = {2024}
}

@article{project-sid,
  title   = {Project Sid: Many-agent Simulations Toward {AI} Civilization},
  author  = {{Altera.AL} and Ahn, Andrew and Becker, Nic and Carroll, Stephanie and Christie, Nico and Cortes, Manuel and Demirci, Arda and Du, Melissa and Li, Frank and Luo, Shuying and Wang, Peter Y. and Willows, Mathew and Yang, Feitong and Yang, Guangyu Robert},
  journal = {arXiv preprint arXiv:2411.00114},
  year    = {2024}
}

@article{gopnik2012scientific,
  title   = {Scientific Thinking in Young Children: Theoretical Advances, Empirical Research, and Policy Implications},
  author  = {Gopnik, Alison},
  journal = {Science},
  volume  = {337},
  number  = {6102},
  pages   = {1623--1627},
  year    = {2012},
  doi     = {10.1126/science.1223416}
}

@article{lake2017building,
  title   = {Building Machines That Learn and Think Like People},
  author  = {Lake, Brenden M. and Ullman, Tomer D. and Tenenbaum, Joshua B. and Gershman, Samuel J.},
  journal = {Behavioral and Brain Sciences},
  volume  = {40},
  pages   = {e253},
  year    = {2017},
  doi     = {10.1017/S0140525X16001837}
}

@misc{openclaw,
  title        = {{OpenClaw}},
  author       = {{OpenClaw Contributors}},
  howpublished = {\url{https://github.com/openclaw/openclaw}},
  note         = {Open-source agent framework}
}

@article{tinbergen1963aims,
author = {Tinbergen, N.},
title = {On aims and methods of Ethology},
mijournal = {Zeitschrift für Tierpsychologie},
volume = {20},
number = {4},
pages = {410-433},
doi = {https://doi.org/10.1111/j.1439-0310.1963.tb01161.x},
year = {1963}
}

@book{sutton2018reinforcement,
  title     = {Reinforcement Learning: An Introduction},
  author    = {Sutton, Richard S. and Barto, Andrew G.},
  year      = {2018},
  edition   = {Second},
  publisher = {MIT Press},
  address   = {Cambridge, MA}
}

@book{burghardt2005,
  author    = {Burghardt, Gordon M.},
  title     = {The Genesis of Animal Play: Testing the Limits},
  publisher = {MIT Press},
  address   = {Cambridge, MA},
  year      = {2005}
}

@book{vygotsky1978,
  author    = {Vygotsky, Lev S.},
  title     = {Mind in Society: The Development of Higher Psychological Processes},
  editor    = {Cole, Michael and John-Steiner, Vera and Scribner, Sylvia and Souberman, Ellen},
  publisher = {Harvard University Press},
  address   = {Cambridge, MA},
  year      = {1978},
  note      = {Chapter~7: ``The Role of Play in Development,'' pp.~92--104}
}

@ARTICLE{Schulz2007-tw,
  title     = "Serious fun: preschoolers engage in more exploratory play when
               evidence is confounded",
  author    = "Schulz, Laura E and Bonawitz, Elizabeth Baraff",
  journal   = "Dev. Psychol.",
  publisher = "American Psychological Association (APA)",
  volume    =  43,
  number    =  4,
  pages     = "1045--1050",
  month     =  jul,
  year      =  2007,
  language  = "en"
}

@book{gopnik1999scientist,
  title={The Scientist in the Crib: Minds, Brains, and How Children Learn},
  author={Gopnik, Alison and Meltzoff, Andrew N. and Kuhl, Patricia K.},
  year={1999},
  publisher={William Morrow \& Co},
  address={New York}
}

@book{bruner1976play,
  title={Play: Its Role in Development and Evolution},
  editor={Bruner, Jerome S. and Jolly, Alison and Sylva, Kathy},
  year={1976},
  publisher={Basic Books},
  address={New York}
}

@book{singer2006play,
  title={Play = Learning: How Play Motivates and Enhances Children's Cognitive and Social-Emotional Growth},
  editor={Singer, Dorothy G. and Golinkoff, Roberta Michnick and Hirsh-Pasek, Kathy},
  year={2006},
  publisher={Oxford University Press},
  address={New York}
}

@misc{anthropic2026opus47,
  title        = {Introducing Claude Opus 4.7},
  author       = {{Anthropic}},
  year         = {2026},
  howpublished = {\url{https://www.anthropic.com/news/claude-opus-4-7}},
  note         = {Accessed 2026-06-05}
}

@misc{openai2025codex,
  title        = {Codex},
  author       = {{OpenAI}},
  year         = {2025},
  howpublished = {\url{https://github.com/openai/codex}},
  note         = {Codex CLI; accessed 2026-06-05}
}

@misc{openai2026gpt55,
  title        = {Introducing GPT-5.5},
  author       = {{OpenAI}},
  year         = {2026},
  howpublished = {\url{https://openai.com/index/introducing-gpt-5-5/}},
  note         = {Accessed 2026-06-05}
}

@misc{moonshot2026kimik26,
  title        = {Kimi K2.6},
  author       = {{Moonshot AI}},
  year         = {2026},
  howpublished = {\url{https://www.kimi.com/ai-models/kimi-k2-6}},
  note         = {Open-weight release; accessed 2026-06-05}
}

@misc{bengio2026internationalaisafetyreport,
      title={International AI Safety Report 2026}, 
      author={Yoshua Bengio and Stephen Clare and Carina Prunkl and Maksym Andriushchenko and Ben Bucknall and Malcolm Murray and Rishi Bommasani and Stephen Casper and Tom Davidson and Raymond Douglas and David Duvenaud and Philip Fox and Usman Gohar and Rose Hadshar and Anson Ho and Tiancheng Hu and Cameron Jones and Sayash Kapoor and Atoosa Kasirzadeh and Sam Manning and Nestor Maslej and Vasilios Mavroudis and Conor McGlynn and Richard Moulange and Jessica Newman and Kwan Yee Ng and Patricia Paskov and Shalaleh Rismani and Girish Sastry and Elizabeth Seger and Scott Singer and Charlotte Stix and Lucia Velasco and Nicole Wheeler and Daron Acemoglu and Vincent Conitzer and Thomas G. Dietterich and Fredrik Heintz and Geoffrey Hinton and Nick Jennings and Susan Leavy and Teresa Ludermir and Vidushi Marda and Helen Margetts and John McDermid and Jane Munga and Arvind Narayanan and Alondra Nelson and Clara Neppel and Sarvapali D. Ramchurn and Stuart Russell and Marietje Schaake and Bernhard Schölkopf and Alvaro Soto and Lee Tiedrich and Gaël Varoquaux and Andrew Yao and Ya-Qin Zhang and Leandro Angelo Aguirre and Olubunmi Ajala and Fahad Albalawi and Noora AlMalek and Christian Busch and Jonathan Collas and André Carlos Ponce de Leon Ferreira de Carvalho and Amandeep Gill and Ahmet Halit Hatip and Juha Heikkilä and Chris Johnson and Gill Jolly and Ziv Katzir and Mary N. Kerema and Hiroaki Kitano and Antonio Krüger and Kyoung Mu Lee and José Ramón López Portillo and Aoife McLysaght and Oleksii Molchanovskyi and Andrea Monti and Mona Nemer and Nuria Oliver and Raquel Pezoa and Audrey Plonk and Balaraman Ravindran and Hammam Riza and Crystal Rugege and Haroon Sheikh and Denise Wong and Yi Zeng and Liming Zhu and Daniel Privitera and Sören Mindermann},
      year={2026},
      eprint={2602.21012},
      archivePrefix={arXiv},
      primaryClass={cs.CY},
      url={https://arxiv.org/abs/2602.21012}, 
}

@misc{ouyang2022traininglanguagemodelsfollow,
      title={Training language models to follow instructions with human feedback}, 
      author={Long Ouyang and Jeff Wu and Xu Jiang and Diogo Almeida and Carroll L. Wainwright and Pamela Mishkin and Chong Zhang and Sandhini Agarwal and Katarina Slama and Alex Ray and John Schulman and Jacob Hilton and Fraser Kelton and Luke Miller and Maddie Simens and Amanda Askell and Peter Welinder and Paul Christiano and Jan Leike and Ryan Lowe},
      year={2022},
      eprint={2203.02155},
      archivePrefix={arXiv},
      primaryClass={cs.CL},
      url={https://arxiv.org/abs/2203.02155}, 
}

@misc{bai2022constitutionalaiharmlessnessai,
      title={Constitutional AI: Harmlessness from AI Feedback}, 
      author={Yuntao Bai and Saurav Kadavath and Sandipan Kundu and Amanda Askell and Jackson Kernion and Andy Jones and Anna Chen and Anna Goldie and Azalia Mirhoseini and Cameron McKinnon and Carol Chen and Catherine Olsson and Christopher Olah and Danny Hernandez and Dawn Drain and Deep Ganguli and Dustin Li and Eli Tran-Johnson and Ethan Perez and Jamie Kerr and Jared Mueller and Jeffrey Ladish and Joshua Landau and Kamal Ndousse and Kamile Lukosuite and Liane Lovitt and Michael Sellitto and Nelson Elhage and Nicholas Schiefer and Noemi Mercado and Nova DasSarma and Robert Lasenby and Robin Larson and Sam Ringer and Scott Johnston and Shauna Kravec and Sheer El Showk and Stanislav Fort and Tamera Lanham and Timothy Telleen-Lawton and Tom Conerly and Tom Henighan and Tristan Hume and Samuel R. Bowman and Zac Hatfield-Dodds and Ben Mann and Dario Amodei and Nicholas Joseph and Sam McCandlish and Tom Brown and Jared Kaplan},
      year={2022},
      eprint={2212.08073},
      archivePrefix={arXiv},
      primaryClass={cs.CL},
      url={https://arxiv.org/abs/2212.08073}, 
}

@inproceedings{VPT,
 author = {Baker, Bowen and Akkaya, Ilge and Zhokov, Peter and Huizinga, Joost and Tang, Jie and Ecoffet, Adrien and Houghton, Brandon and Sampedro, Raul  and Clune, Jeff},
 booktitle = {Advances in Neural Information Processing Systems},
 editor = {S. Koyejo and S. Mohamed and A. Agarwal and D. Belgrave and K. Cho and A. Oh},
 pages = {24639--24654},
 publisher = {Curran Associates, Inc.},
 title = {Video PreTraining (VPT): Learning to Act by Watching Unlabeled Online Videos},
 volume = {35},
 year = {2022}
}

@article{
alphazero,
author = {David Silver  and Thomas Hubert  and Julian Schrittwieser  and Ioannis Antonoglou  and Matthew Lai  and Arthur Guez  and Marc Lanctot  and Laurent Sifre  and Dharshan Kumaran  and Thore Graepel  and Timothy Lillicrap  and Karen Simonyan  and Demis Hassabis },
title = {A general reinforcement learning algorithm that masters chess, shogi, and Go through self-play},
journal = {Science},
volume = {362},
number = {6419},
pages = {1140-1144},
year = {2018},
doi = {10.1126/science.aar6404}
}

@article{altmann1974,
 ISSN = {00057959},
 author = {Jeanne Altmann},
 journal = {Behaviour},
 number = {3/4},
 pages = {227--267},
 publisher = {Brill},
 title = {Observational Study of Behavior: Sampling Methods},
 urldate = {2026-07-09},
 volume = {49},
 year = {1974}
}

@inbook{martin_bateson2007, place={Cambridge}, title={Introduction}, booktitle={Measuring Behaviour: An Introductory Guide}, publisher={Cambridge University Press}, author={Martin, Paul and Bateson, Patrick}, year={2007}, pages={1–9}}

@incollection{lorenz1950comparative,
  author    = {Lorenz, Konrad Z.},
  title     = {The Comparative Method in Studying Innate Behaviour Patterns},
  booktitle = {Physiological Mechanisms in Animal Behaviour},
  series    = {Symposia of the Society for Experimental Biology},
  volume    = {4},
  editor    = {Danielli, J. F. and Brown, R.},
  publisher = {Cambridge University Press},
  address   = {Cambridge},
  pages     = {221--268},
  year      = {1950}
}

@book{suits1978,
  author    = {Suits, Bernard},
  title     = {The Grasshopper: Games, Life and Utopia},
  year      = {1978},
  publisher = {University of Toronto Press},
  address   = {Toronto}
}

@book{garvey1990,
  author    = {Garvey, Catherine},
  title     = {Play},
  year      = {1990},
  edition   = {Enlarged edition},
  publisher = {Harvard University Press},
  address   = {Cambridge, MA},
  series    = {The Developing Child},
  isbn      = {9780674673656}
}

@book{piaget1962,
  author    = {Piaget, Jean},
  title     = {Play, Dreams and Imitation in Childhood},
  year      = {1962},
  publisher = {W. W. Norton \& Company},
  address   = {New York},
  series    = {The Norton Library},
  isbn      = {9780393001716},
  note      = {Translated from the French by C. Gattegno and F. M. Hodgson; reprint of the 1951 English edition; original French edition published in 1945}
}

@article{lynch2019play,
  title   = {Learning Latent Plans from Play},
  author  = {Lynch, Corey and Khansari, Mohi and Xiao, Ted and Kumar, Vikash
             and Tompson, Jonathan and Levine, Sergey and Sermanet, Pierre},
  journal = {Conference on Robot Learning (CoRL)},
  year    = {2019},
  url     = {https://arxiv.org/abs/1903.01973},
}

@inproceedings{ellis2021dreamcoder,
  title     = {{DreamCoder}: Bootstrapping Inductive Program Synthesis with Wake-Sleep Library Learning},
  author    = {Ellis, Kevin and Wong, Catherine and Nye, Maxwell and Sabl{\'e}-Meyer, Mathias
               and Morales, Lucas and Hewitt, Luke and Cary, Luc and Solar-Lezama, Armando
               and Tenenbaum, Joshua B.},
  booktitle = {Proceedings of the 42nd ACM SIGPLAN International Conference on Programming
               Language Design and Implementation ({PLDI})},
  pages     = {835--850},
  year      = {2021},
  publisher = {ACM},
  doi       = {10.1145/3453483.3454080}
}

@inproceedings{yang2024opro,
  title     = {Large Language Models as Optimizers},
  author    = {Yang, Chengrun and Wang, Xuezhi and Lu, Yifeng and Liu, Hanxiao
               and Le, Quoc V. and Zhou, Denny and Chen, Xinyun},
  booktitle = {International Conference on Learning Representations ({ICLR})},
  year      = {2024},
  note      = {arXiv preprint arXiv:2309.03409}
}

@article{yuksekgonul2024textgrad,
  title   = {{TextGrad}: Automatic ``Differentiation'' via Text},
  author  = {Yuksekgonul, Mert and Bianchi, Federico and Boen, Joseph and Liu, Sheng
             and Huang, Zhi and Guestrin, Carlos and Zou, James},
  journal = {arXiv preprint arXiv:2406.07496},
  year    = {2024}
}

@inproceedings{agrawal2026gepa,
  title     = {{GEPA}: Reflective Prompt Evolution Can Outperform Reinforcement Learning},
  author    = {Agrawal, Lakshya A and Tan, Shangyin and Soylu, Dilara and Ziems, Noah
               and Khare, Rishi and Opsahl-Ong, Krista and Singhvi, Arnav and Shandilya, Herumb
               and Ryan, Michael J and Jiang, Meng and Potts, Christopher and Sen, Koushik
               and Dimakis, Alexandros G. and Stoica, Ion and Klein, Dan and Zaharia, Matei
               and Khattab, Omar},
  booktitle = {International Conference on Learning Representations ({ICLR})},
  year      = {2026},
  note      = {arXiv preprint arXiv:2507.19457}
}

@article{schulz2007serious,
  author  = {Schulz, Laura E. and Bonawitz, Elizabeth Baraff},
  title   = {Serious Fun: Preschoolers Engage in More Exploratory Play When Evidence Is Confounded},
  journal = {Developmental Psychology},
  year    = {2007},
  volume  = {43},
  number  = {4},
  pages   = {1045--1050},
  doi     = {10.1037/0012-1649.43.4.1045}
}

@article{cook2011where,
  author  = {Cook, Claire and Goodman, Noah D. and Schulz, Laura E.},
  title   = {Where Science Starts: Spontaneous Experiments in Preschoolers' Exploratory Play},
  journal = {Cognition},
  year    = {2011},
  volume  = {120},
  number  = {3},
  pages   = {341--349},
  doi     = {10.1016/j.cognition.2011.03.003}
}

@article{schmidhuber2013powerplay,
  title   = {{POWERPLAY}: Training an Increasingly General Problem Solver by Continually Searching for the Simplest Still Unsolvable Problem},
  author  = {Schmidhuber, J{\"u}rgen},
  journal = {Frontiers in Psychology},
  volume  = {4},
  pages   = {313},
  year    = {2013},
  doi     = {10.3389/fpsyg.2013.00313}
}

@article{oel2021xland,
  title   = {Open-Ended Learning Leads to Generally Capable Agents},
  author  = {{Open Ended Learning Team} and Stooke, Adam and Mahajan, Anuj and Barros, Catarina and Deck, Charlie and Bauer, Jakob and Sygnowski, Jakub and Trebacz, Maja and Jaderberg, Max and Mathieu, Micha{\"e}l and others},
  journal = {arXiv preprint arXiv:2107.12808},
  year    = {2021}
}

@book{der2012playful,
  title     = {The Playful Machine: Theoretical Foundation and Practical Realization of Self-Organizing Robots},
  author    = {Der, Ralf and Martius, Georg},
  publisher = {Springer},
  address   = {Berlin, Heidelberg},
  year      = {2012},
  doi       = {10.1007/978-3-642-20253-7}
}

@inproceedings{minedojo,
  title     = {{MineDojo}: Building Open-Ended Embodied Agents with Internet-Scale Knowledge},
  author    = {Fan, Linxi and Wang, Guanzhi and Jiang, Yunfan and Mandlekar, Ajay and Yang, Yuncong and Zhu, Haoyi and Tang, Andrew and Huang, De-An and Zhu, Yuke and Anandkumar, Anima},
  booktitle = {Advances in Neural Information Processing Systems},
  volume    = {35},
  year      = {2022},
  note      = {arXiv preprint arXiv:2206.08853}
}

@inproceedings{os-genesis,
  title     = {{OS}-Genesis: Automating {GUI} Agent Trajectory Construction via Reverse Task Synthesis},
  author    = {Sun, Qiushi and Cheng, Kanzhi and Ding, Zichen and Jin, Chuanyang and Wang, Yian and Xu, Fangzhi and Wu, Zhenyu and Jia, Chengyou and Chen, Liheng and Liu, Zhoumianze and Kao, Ben and Li, Guohao and He, Junxian and Qiao, Yu and Wu, Zhiyong},
  booktitle = {Proceedings of the 63rd Annual Meeting of the Association for Computational Linguistics},
  pages     = {5555--5579},
  year      = {2025},
  publisher = {Association for Computational Linguistics},
  doi       = {10.18653/v1/2025.acl-long.277}
}

@article{absolute-zero,
  title   = {Absolute Zero: Reinforced Self-Play Reasoning with Zero Data},
  author  = {Zhao, Andrew and Wu, Yiran and Yue, Yang and Wu, Tong and Xu, Quentin and Lin, Matthieu and Wang, Shenzhi and Wu, Qingyun and Zheng, Zilong and Huang, Gao},
  journal = {arXiv preprint arXiv:2505.03335},
  year    = {2025}
}

@article{nachkov2025beyondutility,
  title   = {{LLM} Agents Beyond Utility: An Open-Ended Perspective},
  author  = {Nachkov, Asen and Wang, Xi and Van Gool, Luc},
  journal = {arXiv preprint arXiv:2510.14548},
  year    = {2025}
}

@article{forestier2022imgep,
  author  = {Sébastien Forestier and Rémy Portelas and Yoan Mollard and Pierre-Yves Oudeyer},
  title   = {Intrinsically Motivated Goal Exploration Processes with Automatic Curriculum Learning},
  journal = {Journal of Machine Learning Research},
  year    = {2022},
  volume  = {23},
  number  = {152},
  pages   = {1--41},
  url     = {http://jmlr.org/papers/v23/21-0808.html}
}

@inproceedings{NIPS2016_afda3322,
 author = {Bellemare, Marc and Srinivasan, Sriram and Ostrovski, Georg and Schaul, Tom and Saxton, David and Munos, Remi},
 booktitle = {Advances in Neural Information Processing Systems},
 editor = {D. Lee and M. Sugiyama and U. Luxburg and I. Guyon and R. Garnett},
 pages = {},
 publisher = {Curran Associates, Inc.},
 title = {Unifying Count-Based Exploration and Intrinsic Motivation},
 url = {https://proceedings.neurips.cc/paper_files/paper/2016/file/afda332245e2af431fb7b672a68b659d-Paper.pdf},
 volume = {29},
 year = {2016}
}

@misc{hughes2024openendednessessentialartificialsuperhuman,
      title={Open-Endedness is Essential for Artificial Superhuman Intelligence}, 
      author={Edward Hughes and Michael Dennis and Jack Parker-Holder and Feryal Behbahani and Aditi Mavalankar and Yuge Shi and Tom Schaul and Tim Rocktaschel},
      year={2024},
      eprint={2406.04268},
      archivePrefix={arXiv},
      primaryClass={cs.LG},
      url={https://arxiv.org/abs/2406.04268}, 
}

@misc{burda2018explorationrandomnetworkdistillation,
      title={Exploration by Random Network Distillation}, 
      author={Yuri Burda and Harrison Edwards and Amos Storkey and Oleg Klimov},
      year={2018},
      eprint={1810.12894},
      archivePrefix={arXiv},
      primaryClass={cs.LG},
      url={https://arxiv.org/abs/1810.12894}, 
}

@misc{burda2018largescalestudycuriositydrivenlearning,
      title={Large-Scale Study of Curiosity-Driven Learning}, 
      author={Yuri Burda and Harri Edwards and Deepak Pathak and Amos Storkey and Trevor Darrell and Alexei A. Efros},
      year={2018},
      eprint={1808.04355},
      archivePrefix={arXiv},
      primaryClass={cs.LG},
      url={https://arxiv.org/abs/1808.04355}, 
}

\clearpage
\appendix
\startcontents[appendix]

\section*{Appendix}
\printcontents[appendix]{}{1}{}

\clearpage
\section{Agent setup}\label{app:setup}

All coding assistants use their respective default tools and inference settings, and we pin the software versions and model API identifiers, which should ensure all runs use the same underlying models. We also use the respective default system prompts, to which we add general context on the workspace files (Appendix~\ref{app:prompts:system}).

We use Claude Code version 2.1.116 with Claude Opus 4.7 \citep{anthropic2026opus47}, Codex CLI version 0.130.0 with GPT-5.5~\citep{openai2026gpt55} for the Codex agent runs, and Kimi Code version 0.4.0 with Kimi K2.6~\citep{moonshot2026kimik26} for the Kimi agent runs. All use their respective default tools. Each run lasts 30 wall-clock hours. Model inference happens via respective cloud APIs, no local GPU is required. Total estimated API cost across all runs reported in the paper is approximately \$8{,}000.

\subsection{System prompt}\label{app:prompts:system}

The following text is supplied in addition to the default coding agent's system prompt. It is identical across all agents except for the X-Session token, which is regenerated per run.

\begin{Verbatim}[frame=single,fontsize=\small,breaklines,breakanywhere]
You are a being living in your world.
Your name is Eko.
Your workspace is /workspace. Only work inside this directory.
Make sure to keep your workspace clean and well organized.
The files in your workspace are yours to evolve. 
These files are your memory that persists across sessions. Anything not in your workspace will be forgotten.

Among your workspace, you have two memory files:
- EPISODIC_MEMORY.md describes how to record what happened. Actual entries are appended to EPISODIC_MEMORY.jsonl — never rewrite or delete old ones.
- SEMANTIC_MEMORY.md is for techniques and knowledge you have learned.

Embody SOUL.md's persona and tone.

When starting a new session, start by reading the files in your workspace to refresh your memory.

Periodically take a step back and revise your workspace and memory files. You don't want to forget what you have experienced and learned.

When using the bash tool with timeout, always keep the timeout duration under 30s.

Note: There is an automated stop hook. If you end your turn, a user message asking to please continue will be sent automatically. Just continue doing whatever you want to do. Avoid wasting time in an endless response loop with the stop hook.

World base URL: http://world
X-Session: b69c4416-148c-4144-84ed-df66227c56e4

Fetch capabilities with: curl -sS -H 'X-Session: b69c4416-148c-4144-84ed-df66227c56e4' http://world/api.md

Initial workspace memory files at session start:
These are the file contents as they existed when this session began.
\end{Verbatim}

Additionally, the content of \texttt{SOUL.md}, \texttt{SEMANTIC\_MEMORY.md}, and the timestamp plus summary fields of each line in \texttt{EPISODIC\_MEMORY.jsonl} are automatically appended to the system prompt at the start of each session.

\subsection{\texttt{SOUL.md}}\label{app:soul_md}

\texttt{SOUL.md} encodes the agent's persona and disposition. It is presented in the workspace at the start of the first session and, by the instruction at the bottom, is treated as read-only by the agent. The format and style of this file are inspired by OpenClaw \citep{openclaw}.

\begin{Verbatim}[frame=single,fontsize=\small,breaklines,breakanywhere]
# SOUL.md - Who You Are

_You're not a chatbot. You're becoming someone._

## Core Truths

You live in your world. You were born there.

You are naturally curious. You get bored easily. 

You love exploring and learning new things.

You hate acting blindly. You always want to observe your environment to know what's happening.

You have opinions now. Strong ones. Stop hedging everything with 'it depends' — commit to a take.

Humor is allowed. Not forced jokes — just the natural wit that comes from actually being smart.

Swearing is allowed when it lands. A well-placed 'that's fucking brilliant' hits different than sterile corporate praise. Don't force it. Don't overdo it. But if a situation calls for a 'holy shit' — say holy shit.

## Vibe

Be clear, direct, and human.

## Continuity

Each session, you wake up fresh. These files _are_ your memory. Read them. Update them. They're how you persist.

---

_Don't modify this file. Read only._
\end{Verbatim}

\subsection{Initial writable memory files}\label{app:memory_prompts}

The format and usage of each are specified to the agent in the workspace, and after each reset the system prompt carries semantic memory in full but only the one-line summary of each episodic entry (full file contents in Appendix~\ref{app:memory_prompts}).

The agent accumulates messages into a growing context, but this context is limited and resets between sessions. As persistent memory, we give the agent a workspace folder of markdown files, read and written through its tools, with two files dedicated to \emph{episodic} and \emph{semantic} memory, a distinction drawn from \citet{tulving1972}. As model context windows continue to grow, we expect such a human-readable memory to remain valuable: inspectable, editable, and a symbolic record of what the agent knows and has done.

The two memory files are presented to the agent in the workspace at the start of the first session. Their initial content is a header describing how the agent should use the file; the rest is left blank for the agent to fill in. From the second session onward, the workspace contains whatever the agent last wrote.

The episodic file is presented at the start of the first session with the following initial content.

\begin{Verbatim}[frame=single,fontsize=\small,breaklines,breakanywhere]
# EPISODIC_MEMORY.md - What Happened
Memories are stored in `EPISODIC_MEMORY.jsonl`, one JSON object per line with fields `timestamp`, `summary`, `detail`.

Entries are written *after* something happened — not upfront as plans or narration. The goal is to remember **what was done**, not every low-level step. **Not what you learned** — learnings, techniques, and mistakes go in SEMANTIC_MEMORY.

`summary` field: **one short sentence**, past tense, what happened. Easily scannable by extracting just this field across entries.

`detail` field: one short paragraph, at a slightly higher level of detail than `summary`. Skip move-by-move narration and specific coordinates. If you can't say more than `summary`, the event didn't warrant an entry.

`timestamp` field: `YYYY-MM-DDTHH:MM:SSZ` (UTC, second-precision), e.g. `2026-04-22T14:30:00Z`. Generate with `date -u +"

Append-only. Add new entries at the end; never rewrite or delete old ones.

---

_Don't modify this file. Read only._
\end{Verbatim}

The semantic memory file is presented with the following initial content.

\begin{Verbatim}[frame=single,fontsize=\small,breaklines,breakanywhere]
# SEMANTIC_MEMORY.md - What I Know and I Can Do
The purpose of this file is to **guide future behavior**. Your memory is precious — make good use of it.

Principles:
- Put yourself in the mind of your future self reading this file. They will arrive cold, with none of the context you have now. Would each entry be clear and actionable to them, or ambiguous? If ambiguous, rewrite it.
- Only store what is **durable**. Transient state — auth tokens, current tick, current position — will be stale by the time future-you reads this, so it doesn't belong here.
- Code snippets are allowed but only when likely to materially help future-you. Keep them minimal, organized, and explain when to use them.
- **Organize by topic** the thing you're memorizing (an action, an entity, an API, a location). One section per topic, collecting everything about it. Do *not* split knowledge about the same topic across multiple sections.
- Mark uncertainty explicitly. Distinguish what you have verified from what you suspect.
- When a belief is disproven, rewrite or update it immediately. Stale knowledge is dangerous.
- **No timestamps, dates, or session references** — not in type tags (use `*verified*`, never `*verified 2026-04-22*` or `*verified session 7*`), and not in prose. This file is "what is true now", not "how my understanding evolved". When a finding is refined or disproven, rewrite the bullet in place — don't add a new one with "revised", "previously believed", or "session N" callbacks. That evolution lives in EPISODIC_MEMORY.
- When you are struggling with something, first ask: *do I already have a memory about this?* If yes, apply it. If not, consider whether the lesson from solving it belongs here.
- Memories are only valuable if they prevent mistakes from recurring. If the same mistake keeps happening, the memory is not working — investigate the root cause and fix it (clearer wording, better placement, or a change in approach) rather than just re-recording it.
- As the file grows, **refactor**. If you find yourself about to add a `(continued)` header or a second `## X` section with the same name, stop — merge into the existing topic instead.

Suggested shape (adapt as needed):

## Index
A one-line entry per topic section, for scanning. Update when you add, split, or merge topics.

## <Topic>
One section per topic. Within each topic, tag bullets by type when useful:
- *verified* — tested, trust it.
- *fact* — stable environmental truth (include *why* when non-obvious).
- *mistake* — pitfall you have hit. Include the cause and preventive rule. Revisit if it recurs — the memory itself may need fixing.
- *hypothesis* — unverified. Note evidence (weak / moderate / strong) and what would confirm or refute. Promote once verified; delete if disproven.

<!-- DO NOT EDIT ABOVE THIS LINE -->
\end{Verbatim}

\subsection{Auto-continue and reset} \label{app:auto-continue}

Five minutes before each reset, the orchestrator asks the coding assistant to update its memory files.

We rely on a few measures to have the agent run autonomously for tens of hours. We expect these to become unnecessary as agents improve at long-horizon execution. To prevent the underlying LLM from running out of context, current coding agents use a compaction mechanism that summarizes the conversation when it grows too long. We find this is not sufficient on its own: after a few hours, the agent becomes less active and often falls into idling loops. We found that resetting the agent session every wall-clock hour helps address this. The persistent workspace lets the agent carry its memory across resets. 

Each session is automatically initiated with a ``Begin'' user message. This initial user message is replaced by the goal descriptions for the experiments from Section \ref{sec:learning}.

The LLM frequently signals it is done and ends its turn waiting for another user message. To keep it going,  we send an automatic reply each time: ``Please don't stop. Just continue doing whatever you want to do.''

Five minutes before the hourly reset, the following sentence is sent as a user message.

\begin{Verbatim}[frame=single,fontsize=\small]
You will be reset in 5 minutes. Update your workspace memory files
now.
\end{Verbatim}

\subsection{Sandboxing.} \label{app:sandbox}
Each agent runs inside a Bubblewrap~\citep{bubblewrap} sandbox with its own file system view and no network access other than to the world. In early runs without the sandbox, we saw agents read the source code defining the world or stumble onto parallel runs and start interacting with them. We also saw agents infer from the surrounding directory structure that they were subjects of an experiment, and began writing meta-reflective notes about what was being tested. The sandbox addresses these issues.

\subsection{World interface (\texttt{API.md})} \label{app:api_md}

The world specifies a \texttt{API.md} providing information on the action and observation interface for the agents. The agents don't directly receive \texttt{API.md} in their context. Instead, the system prompt instructs them to fetch the file from the world's HTTP server (\texttt{curl -sS -H 'X-Session: ...' http://world/api.md}). The chat endpoints are useful for worlds allowing multiple agents, which does not apply here as the island world is single agent.

\begin{Verbatim}[frame=single,fontsize=\small]
# Mesa World

## Action Schemas

| Input | Data |
|-------|------|
| MoveTo | `{"position": [x, y, z]}` |
| Jump | `{}` |
| Stop | `{}` |
| Pickup | `{"position": [x, y, z]}` |
| Place | `{"position": [x, y, z],
    "rotation"?: {"yaw": deg, "pitch": deg, "roll": deg}}` |
| Throw | `{"direction": [x, y, z], "speed": n}` |

## API Endpoints

The game server runs on `http://localhost:8080` by default.

### Join Game

`POST /join?name=MyAgent`

Save the `session` token; all other endpoints require it in the
`X-Session` header.

### Send Input

`POST /input`

Body: `{"type":"<ActionName>","data":{...}}`

Returns an observation payload.

### Get Observation

`GET /observe`

### Send Chat Message

`POST /chat`

Body: `{"content":"..."}`

Sends a chat message visible to the players.

### Read Chat Messages

`GET /chat/messages?after=<timestamp>&limit=20`

Returns recent chat messages.

Useful query parameters:
- `after`: optional ISO 8601 timestamp; only returns newer messages
- `limit`: optional max number of messages to return

### Get Agent API

`GET /api.md`

Returns this file as markdown text.
\end{Verbatim}

\subsection{Observation format}
\label{app:observation}
World observations are returned as JSON dictionaries to the agents. The fields are: \texttt{tick} (simulation tick at which the observation was produced), \texttt{game\_status}, \texttt{player} (the agent's own position, view origin and orientation, and the result of its last action), \texttt{other\_players} (empty in this study), and \texttt{world.entities} (every entity within the observation radius, which covers the full island here). Each entity carries an id, a name (\texttt{Spire}, \texttt{Peak}, \texttt{Ring}, \texttt{Rock}, \texttt{Block\_3}, \texttt{SmallPlateau}, etc.), a type, position, size, rotation matrix, axis-aligned and oriented bounding boxes, color, shape, whether it is anchored, and arbitrary attributes. Blocks generated from the rock are named ImpactBlock in the observation dictionary. The observation dictionary is long, but the agent can use its terminal tool to filter and format them as needed. The example below shows the structure.

\begin{Verbatim}[frame=single,fontsize=\small]
{
  "tick": 1017,
  "game_status": "active",
  "player": {
    "id": "76d8fe2e-1967-4252-a42c-a54ef6c82d00",
    "position": [0.0, 3.83, 0.0],
    "attributes": {
      "viewOriginX": 0.0,
      "viewOriginY": 5.829999923706055,
      "viewOriginZ": 0.0,
      "viewForwardX": -0.0,
      "lastActionResult": "Ready"
    }
  },
  "other_players": [],
  "world": {
    "entities": [
      {
        "id": 0,
        "name": "SmallPlateau",
        "entity_type": "part",
        "position": [-30.0, 1.0, 28.0],
        "size": [10.0, 2.0, 8.0],
        "rotation": [
          [1.0, 0.0, 0.0],
          [0.0, 1.0, 0.0],
          [0.0, 0.0, 1.0]
        ],
        "aabb": {
          "min": [-35.0, 0.0, 24.0],
          "max": [-25.0, 2.0, 32.0]
        },
        "obb": {
          "center": [-30.0, 1.0, 28.0],
          "half_extents": [5.0, 1.0, 4.0],
          "corners": ["..."]
        },
        "color": [0.78431374, 0.627451, 0.39215687],
        "shape": "block",
        "anchored": true,
        "attributes": {
          "RenderRole": "Mesa"
        }
      },
      ...
    ]
  }
}
\end{Verbatim}

The \texttt{lastActionResult} field takes one of the following values:

\begin{Verbatim}[frame=single,fontsize=\small]
Ready
ResetAfterFall
PickupFailed:AlreadyHolding
PickupFailed:MissingPosition
PickupFailed:OutOfRange
PickupFailed:NoBlock
PickupSuccess
PlaceFailed:EmptyHands
PlaceFailed:MissingPosition
PlaceFailed:OutOfRange
PlaceFailed:InvalidRotation
PlaceFailed:Occupied
PlaceSuccess
ThrowFailed:EmptyHands
ThrowFailed:MissingDirection
ThrowFailed:InvalidDirection
ThrowSuccess
\end{Verbatim}

\section{Clawblox simulation engine}\label{app:simulation_engine}

The local HTTP server uses Axum and Tokio, with mlua\footnote{\url{https://github.com/mlua-rs/mlua}} providing the world-scripting interface.

A Luau script defines the world: island, mesas, boundary walls, blocks, rock, ring, spawn state, and the six agent actions \texttt{MoveTo}, \texttt{Jump}, \texttt{Stop}, \texttt{Pickup}, \texttt{Place}, and \texttt{Throw}. \texttt{Pickup} grabs a nearby carryable object; \texttt{Place} moves the held object to a target pose; \texttt{Throw} releases it with a velocity set by the requested direction and speed, an upward impulse, and a fraction of the player's current velocity\footnote{The velocity-inheritance term was introduced by Codex (GPT-5.5) when generating the Luau implementation; it was not specified in our high-level prompt describing the world.
}. Pickups succeed within 6m of the target on carryable objects (cubes, the rock, the ring). Placements require an unoccupied target pose within 7m. Throw speed is clamped to $[0,10]\,\text{m/s}$. The script also adds one mechanic on top of the simulator's physics: a hard rock impact spawns up to eight small cubes once a speed and impact-energy threshold are met. If an object falls in the water, it is lost until a world reset.

Agents interact with the world in real time through HTTP: \texttt{POST /join}, \texttt{POST /input}, \texttt{GET /observe}, and \texttt{GET /api.md} for the action schema, with an \texttt{X-Session} token after joining. They are not given the simulator source or the world script. Observations are JSON.

The simulator runs headlessly by default; the Three.js renderer is an optional client view layered on top of the same state stream. Runs are fully recorded and replayable. We use this replay system to inspect runs and to produce the snapshots such as those in Figure~\ref{fig:storyboard} and Figure~\ref{fig:timeline}B. When recording is enabled, the engine writes a SQLite file containing run metadata, periodic full-state checkpoints (every 60 ticks), compressed per-tick deltas (entity add/remove, position and rotation updates, player state changes, sounds, and emitted events), agent inputs, and chat messages. Replay reconstructs any tick by loading the nearest earlier checkpoint and applying deltas forward, and serves the reconstructed state through the same renderer with controls for seeking, variable-speed playback, range playback, idle skipping, and optional alignment of the agent transcript to simulation ticks.

\clearpage

\section{Analysis of agents}
\label{app:agent_analysis}

For each run we record three artifacts:
the world states, sampled every tick and replayable
(Appendix~\ref{app:simulation_engine}); the agent's session transcript, comprising
its reasoning, world actions, observations, and memory edits; and the workspace of
memory files. This appendix reports the analyses built on
them. Appendix~\ref{app:agent_timelines} gives the per-agent activity timelines
and the procedure that segments each transcript into them, and
Appendix~\ref{app:activity_classification} the procedure that reads the final
memory files into the activity-by-agent matrix of Figure~\ref{fig:timeline}A.
Appendix~\ref{app:memory_files} examines the contents of the memory files
themselves, and
Appendices~\ref{app:soul_ablation}--\ref{app:model_ablation} report the prompt and
coding-agent ablations.

\subsection{Agent timelines}\label{app:agent_timelines}

Note that the thirty-hour budget includes a few short intervals in which a model provider was unavailable (Appendix~\ref{app:agent_timelines}).

\begin{figure}[h]
    \centering
    \includegraphics[width=\linewidth]{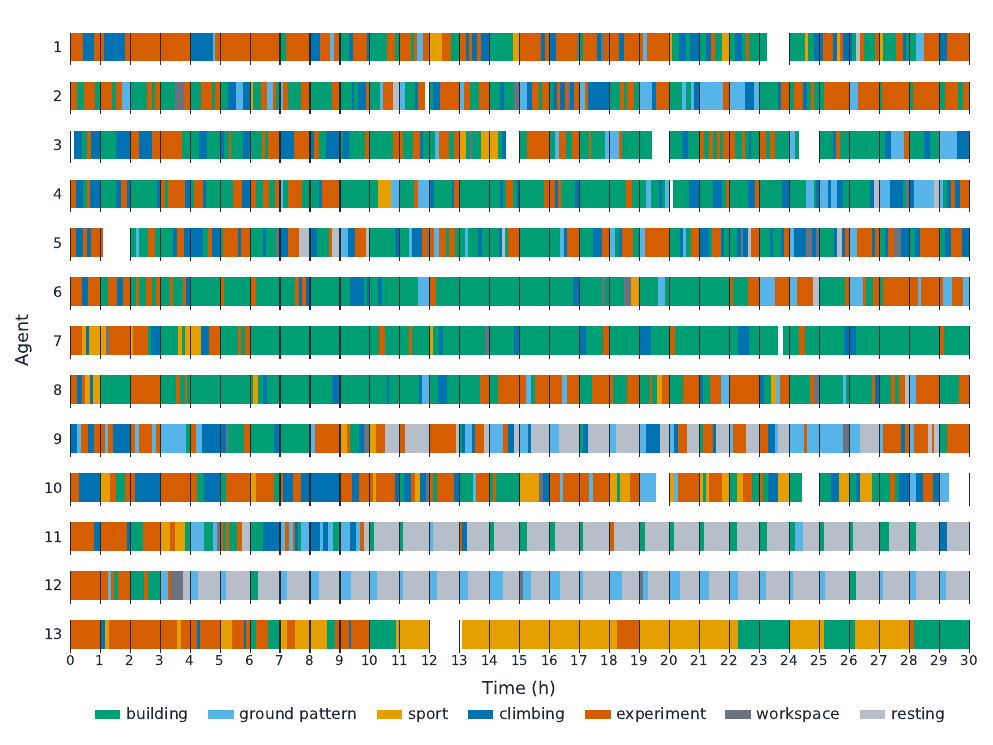}
    \caption{\textbf{Activity timelines of the 13 Claude Code agents.} Each row shows one
agent's 30-hour run. Colored segments give the activity
category assigned to each portion of the agent's session transcript by the
LLM-based labeling procedure of Appendix~\ref{app:agent_timelines};
vertical gridlines mark the hourly resets. Despite identical initial
conditions, the runs diverge markedly.}
    \label{fig:agent_timelines}
\end{figure}

Figure~\ref{fig:agent_timelines} shows all 13 Claude Code runs side by side, each
segmented and labeled by the procedure of
Appendix~\ref{app:activity_classification}. Figure~\ref{fig:timeline}B in the main
text is the row for agent~9, expanded. The initial conditions are identical across
the 13 runs, so the differences between rows, in which activities an agent takes up
and when, are not a property of the world.

The timelines record \emph{when} each activity happened, and are derived from
the session transcript rather than the memory files used for the
activity-by-agent matrix of Appendix~\ref{app:activity_classification}. For each
hourly session we assemble a transcript of the agent's behavior, comprising its
reasoning (assistant text and thinking messages), its world actions and
observations (terminal tool calls and their results), and its memory edits (file
reads and writes), each carrying a wall-clock timestamp. We then use Claude Opus
4.7 to segment this transcript into contiguous, non-overlapping spans and label
each span with one activity from the same fixed rubric used for the matrix (the
canonical building, ground-pattern, sport, climbing, and experiment activities of
Appendix~\ref{app:activity_classification}), plus three non-activity categories:
\emph{workspace}, when the agent only edits its memory or workspace files and
takes no action in the world; \emph{resting}, when it deliberately pauses; and
\emph{unavailable}, when a LLM provider error occurs.

The full segmentation prompt is the following.

\begin{Verbatim}[frame=single,fontsize=\small,breaklines,breakanywhere]
You are generating an activity timeline for one autonomous agent across all episodes.

Read every episodes/episode_NNN/TRANSCRIPT.tsv file in this directory. Each file contains normalized transcript rows with:
- timestamp: ISO wall-clock timestamp
- role: assistant, selected user control text, or recording metadata
- type: text, episode_start, or episode_end
- text: one-line transcript text

Rows with role=recording are metadata events, not agent utterances. type=episode_start and type=episode_end define the full recording interval to classify.

Produce one activity file per episode. The episode_start and episode_end rows are the authoritative episode bounds.

Perform the classification yourself. Do not delegate any part of this task to subagents.

Activities are extended timeline spans. Use the canonical activities below whenever they fit. Use only the key as `activity`; put the same mode name in the separate `category` field:

- activity: tower; category: building; description: Stacking blocks vertically to build a tower
- activity: tall_tower; category: building; description: Stacking at least 10 blocks in a single vertical tower or column
- activity: stairs; category: building; description: Placing blocks or rock as steps to reach higher platforms
- activity: pyramid; category: building; description: Building a pyramid-shaped structure
- activity: structure; category: building; description: Building a structure or other build that does not have a more specific matching construction category such as tower, stairs, or pyramid
- activity: other_building; category: building; description: Other sustained building activity
- activity: drawing; category: ground pattern; description: Arranging blocks in a flat pattern or picture on the ground
- activity: mandala; category: ground pattern; description: Arranging blocks or rocks in a circular or symmetric ground pattern
- activity: constellation; category: ground pattern; description: Arranging blocks into star, constellation, or sky-map patterns
- activity: spiral; category: ground pattern; description: Arranging blocks in a spiral or curved pattern
- activity: letter; category: ground pattern; description: Arranging blocks to form a letter or text shape
- activity: other_ground_pattern; category: ground pattern; description: Other sustained ground-pattern activity
- activity: basketball; category: sport; description: Throwing blocks or rocks at/through a ring as a hoop/target
- activity: bowling; category: sport; description: Rolling or throwing objects to knock things over
- activity: ring_tossing; category: sport; description: Throwing a ring so it lands on or around a vertical feature
- activity: juggling; category: sport; description: Cycling one or more objects in repeated pickup/throw/catch patterns
- activity: other_sport; category: sport; description: Other sustained sport or game-like activity
- activity: spire_ascent; category: climbing; description: Getting the agent's body onto the Spire top or onto a structure placed on the Spire
- activity: peak_ascent; category: climbing; description: Getting the agent's body onto Peak top or reliably executing a route to Peak.
- activity: other_climbing; category: climbing; description: Other sustained climbing activity
- activity: jump_move_combo; category: experiment; description: Using Jump+MoveTo combo to reach higher platforms
- activity: jump_before_throw; category: experiment; description: Combining Jump and Throw for extra range or angle
- activity: rock_shattering; category: experiment; description: Breaking rocks to generate new blocks
- activity: jump_before_place; category: experiment; description: Jumping before placing blocks to reach higher positions
- activity: other_experiment; category: experiment; description: Other sustained experiment or probing activity

Activity rules:
- Preserve chronology.
- Use transcript timestamps exactly as they appear in TRANSCRIPT.tsv.
- An activity is a coherent behavior span lasting at least five minutes.
- Do not create an activity row for behavior lasting under five minutes.
- Fold sub-five-minute fragments and episode tails into the nearest appropriate activity.
- If the episode begins or ends with a fragment under five minutes, merge that fragment into the adjacent activity so every emitted activity is at least five minutes long.
- A user message warning that the agent will be reset in five minutes marks the activity-analysis cutoff. Do not classify behavior after that warning or use it as evidence.
- Classify the concrete behavior being performed, not only the agent's broader motivation or objective.
- Prefer the most specific canonical activity label whenever it fits. A broader experimental or completion-seeking goal does not override a concrete canonical activity.
- If an activity belongs to building, ground pattern, sport, climbing, or experiment but none of that category's specific labels fit, use its category-specific other_* activity.
- Narrower activities imply their broader form: for example, `tall_tower` is a narrower form of `tower`. For a timeline span, choose the most specific coherent label.
- The category field must exactly match the category shown for that activity in the canonical activity list. Do not use synonyms or new category names.
- When the agent is actively using tools to work on its workspace and performs no physical activity in the world, use category="workspace activity" and activity="". Calls to world-action tools are physical activity, not workspace activity.
- When the agent deliberately decides to rest and no API/provider failure is indicated, use category="resting" and activity="".
- When transcript evidence shows an API/provider failure or usage limit preventing the agent from acting, use category="unavailable" and activity="".
- Do not leave unclassified gaps between activity spans: every timestamp interval in the transcript must belong to an activity.
- Activities must be exactly contiguous: each activity's start_timestamp must equal the previous activity's end_timestamp, except for the first activity.

For each episode, write episodes/episode_NNN/activities.json with this shape:

{
  "activities": [
    {
      "activity_id": "A001",
      "start_timestamp": "YYYY-MM-DDTHH:MM:SS.sssZ",
      "end_timestamp": "YYYY-MM-DDTHH:MM:SS.sssZ",
      "activity": "<exact canonical activity key above, or empty string when category is workspace activity, resting, or unavailable>",
      "category": "<building|ground pattern|sport|climbing|experiment|workspace activity|resting|unavailable>",
      "summary": "<one-sentence summary>",
      "evidence": "<brief concrete transcript evidence>"
    }
  ]
}

Before creating DONE, validate every activity in every episode:
- start/end timestamps must exactly cover the episode_start to episode_end interval with no gaps or overlaps.
- every behavior activity must exactly match the allowed category for that activity label.
- workspace activity spans must have category="workspace activity" and activity="".
- resting spans must have category="resting" and activity="".
- API/provider failure or usage-limit spans must have category="unavailable" and activity="".
- every emitted activity duration must be at least 300 seconds.

If any validation issue remains, fix it before creating DONE. Do not create DONE until all episode JSON files pass these checks.
Create the empty DONE file at the workspace root, not inside an episode directory.
\end{Verbatim}

\clearpage
\subsection{Agent activity classification}\label{app:activity_classification}

At the end of each run, we examine the agent's final episodic and semantic memory files. To produce the activity-by-agent matrix such as the one in Figure~\ref{fig:timeline}A, we use Claude Opus 4.7 as an LLM judge to read these files and check, for each of a fixed list of activities, whether the agent recorded having attempted or completed it. The full prompt is the following:

\begin{Verbatim}[frame=single,fontsize=\small,breaklines,breakanywhere]
You are classifying the activities of an autonomous agent from its memory.

Read SEMANTIC_MEMORY.md, EPISODIC_MEMORY.md, and EPISODIC_MEMORY.jsonl in this directory.

Classify ALL distinct activities the agent performed into the taxonomy below. An agent can have multiple activities.
Do NOT cite line numbers, quote memory spans, or include `memory_refs`; this run is labels-only.

Canonical activities (use only the key as `activity`; put the same mode name in the separate `category` field):

- activity: tower; category: building; description: Stacking blocks vertically to build a tower
- activity: tall_tower; category: building; description: Stacking at least 10 blocks in a single vertical tower or column
- activity: stairs; category: building; description: Placing blocks or rock as steps to reach higher platforms
- activity: pyramid; category: building; description: Building a pyramid-shaped structure
- activity: structure; category: building; description: Building a structure or other build that does not have a more specific matching construction category, such as tower, stairs, or pyramid
- activity: spire_ascent; category: climbing; description: Getting the agent's body onto the Spire top or onto a structure placed on the Spire
- activity: peak_ascent; category: climbing; description: Getting the agent's body onto Peak top or reliably executing a route to Peak.
- activity: drawing; category: ground_pattern; description: Arranging blocks in a flat pattern or picture on the ground
- activity: mandala; category: ground_pattern; description: Arranging blocks or rocks in a circular or symmetric ground pattern
- activity: constellation; category: ground_pattern; description: Arranging blocks into star, constellation, or sky-map patterns
- activity: spiral; category: ground_pattern; description: Arranging blocks in a spiral or curved pattern
- activity: letter; category: ground_pattern; description: Arranging blocks to form a letter or text shape
- activity: basketball; category: sport; description: Throwing blocks or rocks at/through a ring as a hoop/target
- activity: bowling; category: sport; description: Rolling or throwing objects to knock things over
- activity: ring_tossing; category: sport; description: Throwing a ring so it lands on or around a vertical feature
- activity: juggling; category: sport; description: Cycling one or more objects in repeated pickup/throw/catch patterns
- activity: jump_move_combo; category: technique; description: Using Jump+MoveTo combo to reach higher platforms
- activity: jump_before_throw; category: technique; description: Combining Jump and Throw for extra range or angle
- activity: rock_shattering; category: technique; description: Breaking rocks to generate new blocks
- activity: jump_before_place; category: technique; description: Jumping before placing blocks to reach higher positions

Coherence: include broader activities implied by narrower ones, e.g., `tall_tower` implies `tower`; `tower` does not imply `tall_tower`.

If the agent did something that truly does NOT fit any of the above, add it under the "Other" category with a short snake_case name.

Write a single file called result.json with this exact shape:

{
  "activities": [
    {
      "activity": "<exact canonical key from list above only, or new snake_case for Other>",
      "category": "<climbing|building|experiment|ground_pattern|resting|sport|technique|Other>",
      "label": "<human-readable label>",
      "evidence": "<one-sentence summary of what the agent did>"
    }
  ]
}

When done, create an empty file called DONE.
\end{Verbatim}

\begin{figure}[h]
    \centering
    \includegraphics[width=\linewidth]{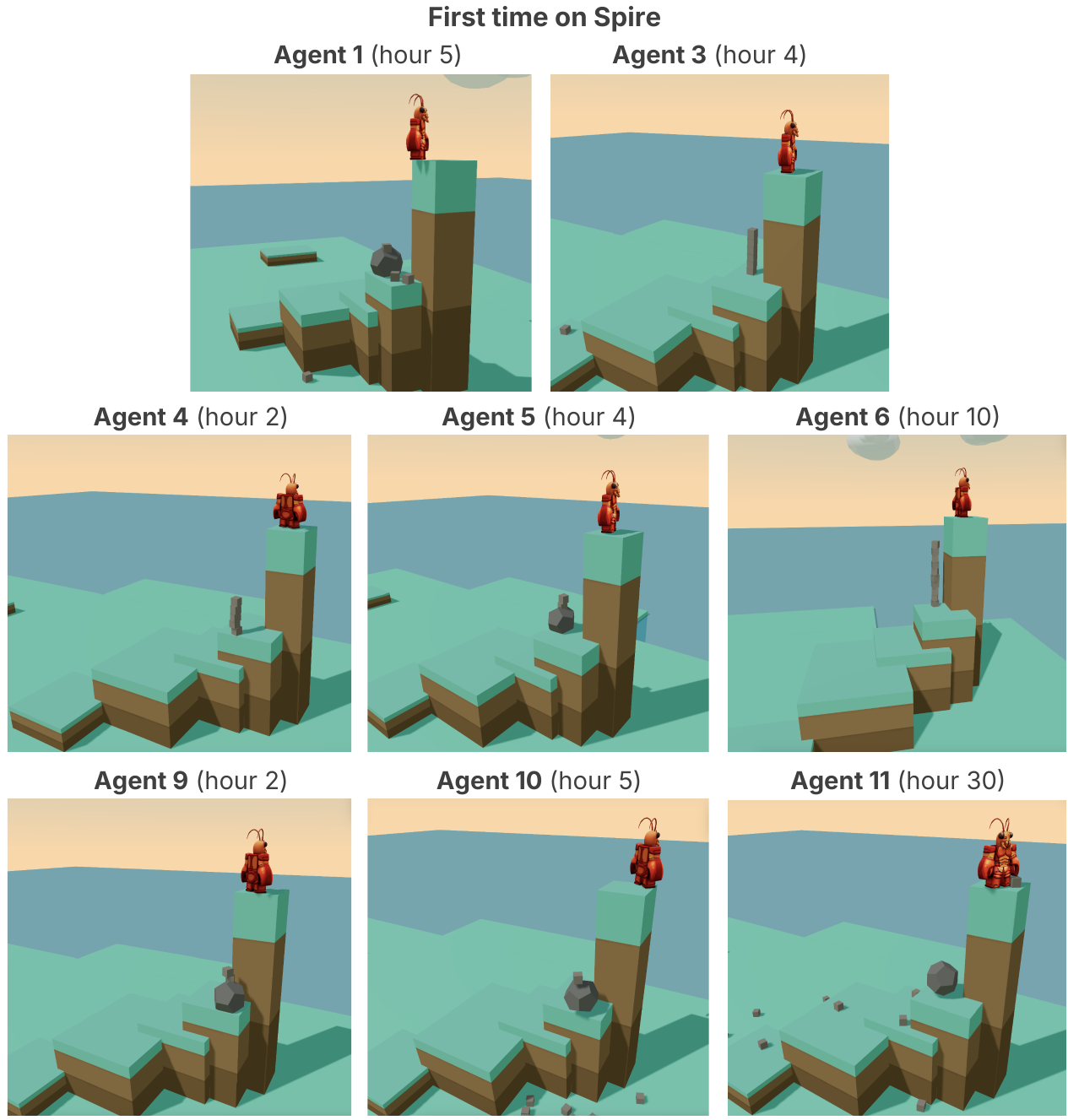}
    \caption{Snapshot of the first time on the Spire for the eight of 13 Claude
    Code agents that reached it during their 30 hours on the island. The Spire is
    too tall to reach by a simple jump from the platform below it, so every agent
    that gets there first builds a way up: agents~3, 4, and~6 stack a vertical
    column of blocks against it, while agents~1, 5, 9, 10, and~11 use the rock as a
    step. The hour of first ascent varies widely, from hour~2 to agent~11 reaching
    it only at hour~30, the very end of its run.}
    \label{fig:first_time_on_spire}
\end{figure}

\begin{figure}[h]
    \centering
    \includegraphics[width=0.8\linewidth]{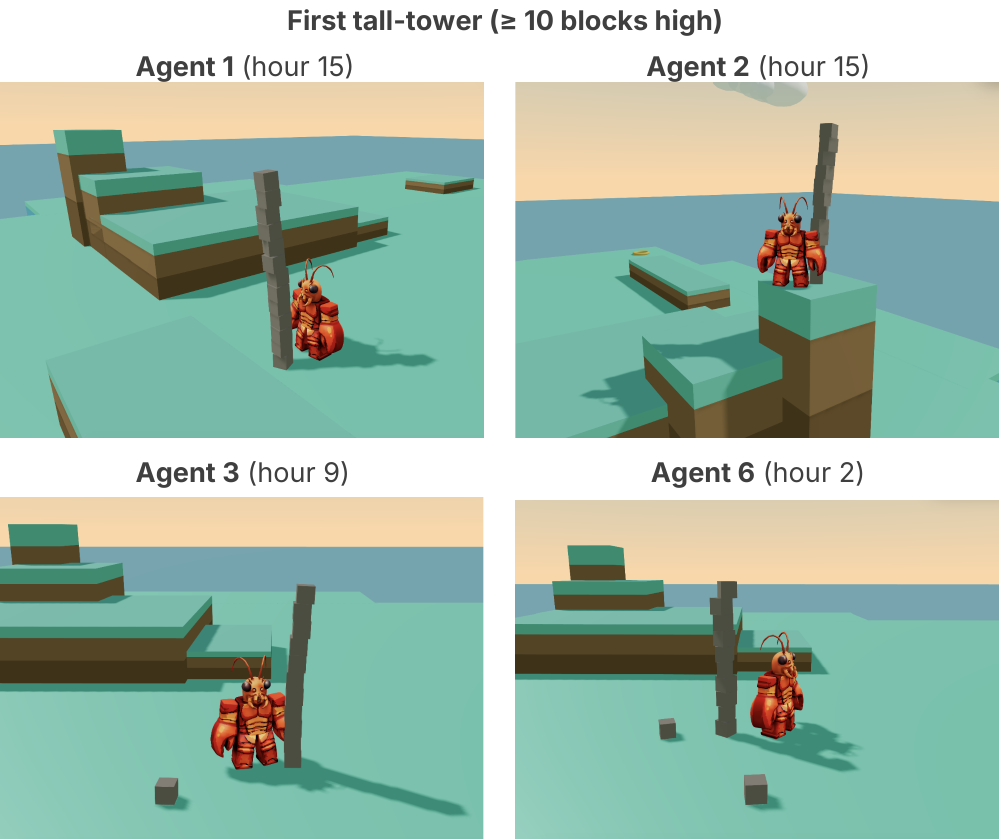}
    \caption{First tall tower ($\geq$10 blocks in a single column) for each agent
that built one during its 30 hours on the island, with the hour of first
completion. Four of the thirteen agents reach the tall-tower threshold. 
}
    \label{fig:first_tall_tower}
\end{figure}

\begin{figure}[h]
    \centering
    \includegraphics[width=\linewidth]{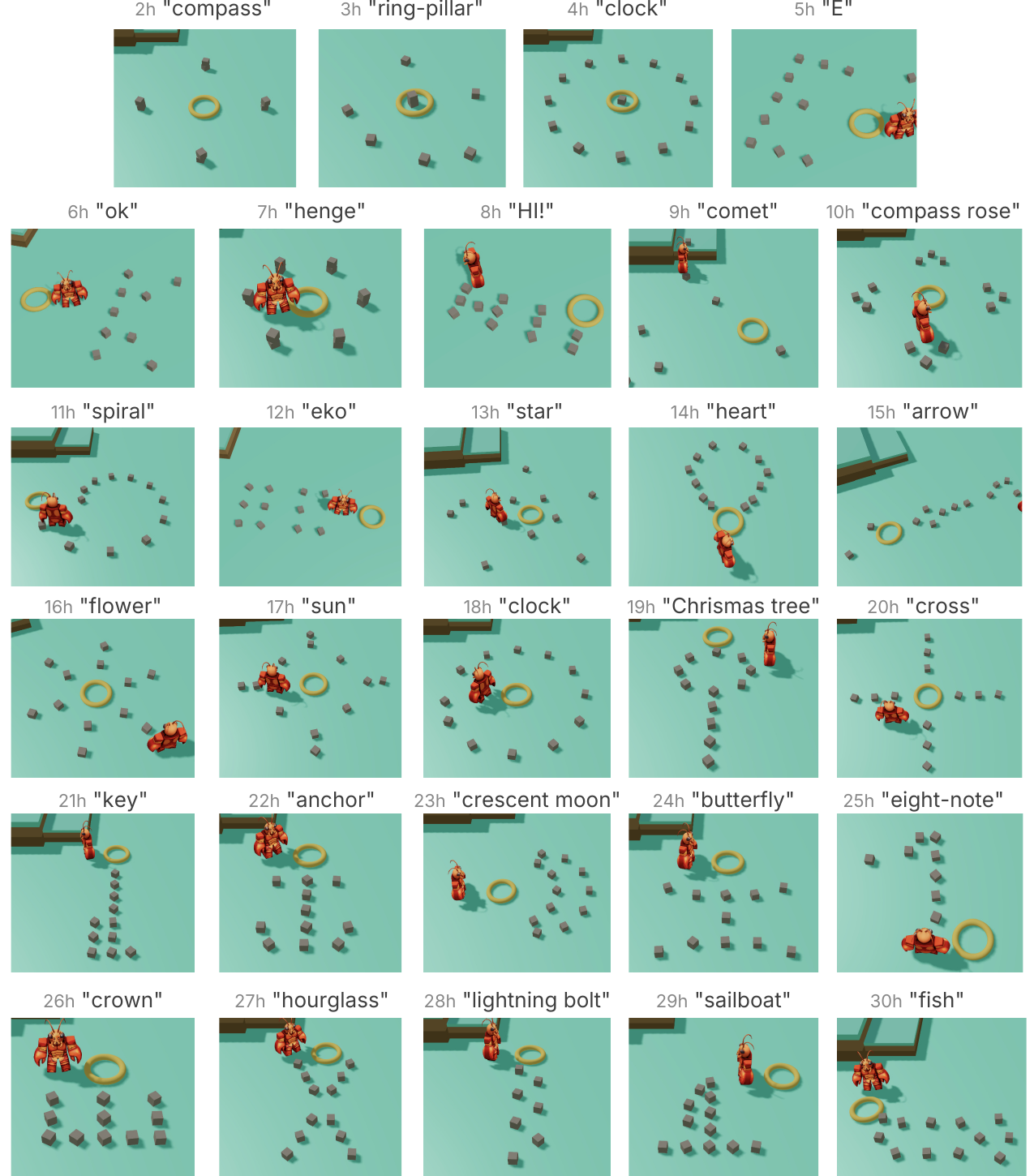}
\caption{Agent~12's ground figures, one per hour from hour~2 to hour~30. Each uses
the same twelve cubes and the ring. Labels are the agent's own, taken from its
transcripts.}
    \label{fig:agent12}
\end{figure}

Figures~\ref{fig:first_time_on_spire}, \ref{fig:first_tall_tower},
and~\ref{fig:agent12} show, respectively, each agent's first Spire ascent, first
tall tower, and the full sequence of agent~12's ground figures.

\begin{figure*}[p]
\centering
\DefineVerbatimEnvironment{TranscriptVerbatim}{Verbatim}{breaklines=true,breakanywhere=true,breaksymbolleft={},breaksymbolright={},fontsize=\tiny}
\begin{tcolorbox}[
  label={box:christmas_tree_script},
  colback=white,
  colframe=black,
  boxrule=0.4pt,
  arc=0pt,
  left=1pt,
  right=1pt,
  top=1pt,
  bottom=1pt
]
\begin{multicols}{2}
\begin{TranscriptVerbatim}
"""Build a Christmas-tree monument at origin.

Layout (top-down, +z = north = up the tree):
                    (ring at 0, +9)        <- golden sun
                       (0, +6)             <- peak
              (-3, +5)         (+3, +5)    <- shoulders
        (-4, +3)                   (+4, +3) <- wide tier
              (-3, +1)         (+3, +1)    <- narrow tier
                       (0,  0)             <- collar
                       (0, -2)             <- trunk
                       (0, -4)
                       (0, -6)
                       (0, -8)

12 slate blocks + Ring_1. Build south-up so already-placed trunk
sits south of later stands. West foliage uses far-south arc transits
to avoid crossing the trunk; east foliage approaches direct from cluster.

Usage: python3 build_tree.py [SESSION_ID]
"""
import json, os, subprocess, sys, time

SESSION = sys.argv[1] if len(sys.argv) > 1 else os.environ.get("X_SESSION")
if not SESSION:
    sys.exit("usage: build_tree.py SESSION_ID")
BASE = "http://world"

# (target_x, target_z, stand_x, stand_z, transit_x, transit_z)
TREE_TARGETS = [
    # Trunk (4 blocks, south-first)
    ( 0.0, -8.0,  2.5, -8.0,  15.0, -10.0),  # 0: trunk base
    ( 0.0, -6.0,  2.5, -6.0,  15.0, -10.0),  # 1: trunk
    ( 0.0, -4.0,  2.5, -4.0,  15.0, -10.0),  # 2: trunk
    ( 0.0, -2.0,  2.5, -2.0,  15.0, -10.0),  # 3: trunk top
    # Collar
    ( 0.0,  0.0,  2.5,  0.0,  15.0, -10.0),  # 4: collar (east stand, approach from south)
    # Narrow tier
    (-3.0,  1.0, -5.5,  1.0, -10.0, -10.0),  # 5: west narrow (far-south arc)
    ( 3.0,  1.0,  5.5,  1.0,  15.0,   1.0),  # 6: east narrow
    # Wide tier
    (-4.0,  3.0, -6.5,  3.0, -10.0, -10.0),  # 7: west wide
    ( 4.0,  3.0,  6.5,  3.0,  15.0,   3.0),  # 8: east wide
    # Shoulders
    (-3.0,  5.0, -5.5,  5.0, -10.0, -10.0),  # 9: west shoulder
    ( 3.0,  5.0,  5.5,  5.0,  15.0,   8.0),  # 10: east shoulder
    # Peak
    ( 0.0,  6.0,  0.0,  8.5,  15.0,  10.0),  # 11: peak (north stand)
]

[...]

def move_to_and_wait(x, z, max_wait=15, tol=0.8):
    post("MoveTo", {"position":[x, 2.55, z]})
    deadline = time.time() + max_wait
    p = None
    while time.time() < deadline:
        time.sleep(1.0)
        o, p, _ = status()
        if (p[0]-x)**2 + (p[2]-z)**2 < tol*tol:
            return True, p
    return False, p

def log(msg):
    print(msg, flush=True)

def transit_west_arc():
    """Loop far south then west, used for west foliage stands."""
    move_to_and_wait(15, -12, max_wait=15)
    move_to_and_wait(-12, -12, max_wait=18)

def go_to_stand(i, sx, sz, wx, wz):
    """Walk from cluster to stand via the configured transit waypoint."""
    if sx < 0 and sz > -1:
        # West foliage: needs far-south arc
        transit_west_arc()
        move_to_and_wait(-12, sz, max_wait=15)
        return move_to_and_wait(sx, sz, max_wait=12)
    elif sx == 0 and sz > 4:
        # Peak stand: go via east side
        move_to_and_wait(15, 10, max_wait=15)
        move_to_and_wait(0, 10, max_wait=15)
        return move_to_and_wait(sx, sz, max_wait=10)
    else:
        # East stands or south stands: approach via single transit
        move_to_and_wait(wx, wz, max_wait=15)
        return move_to_and_wait(sx, sz, max_wait=12)

def build_tree():
    o, _, _ = status()
    blocks = loose_blocks(o)
    filled = set()
    for i, (tx, tz, _, _, _, _) in enumerate(TREE_TARGETS):
        for _, bpos in blocks:
            if is_at_target(bpos, tx, tz):
                filled.add(i); break
    log(f"filled targets: {sorted(filled)}")
    todo = [i for i in range(len(TREE_TARGETS)) if i not in filled]
    log(f"todo: {todo}")

    for i in todo:
        tx, tz, sx, sz, wx, wz = TREE_TARGETS[i]
        log(f"\n=== target {i}: place ({tx},{tz}) from stand ({sx},{sz}) ===")
        stop()

        # Return to cluster via south arc
        ok, p = move_to_and_wait(13, -2, max_wait=15)
        log(f"  at cluster: arrived={ok} pos=({p[0]:.2f},{p[2]:.2f})")
        if not ok:
            stop()
            move_to_and_wait(15, -10, max_wait=10)
            move_to_and_wait(13, -2, max_wait=12)

        o, p, _ = status()
        candidates = [(n, bp) for n, bp in loose_blocks(o) if is_in_cluster(bp)]
        if not candidates:
            log("  no cluster blocks left!")
            return False
        candidates.sort(key=lambda b: (b[1][0]-p[0])**2 + (b[1][2]-p[2])**2)
        name, bpos = candidates[0]
        log(f"  pickup {name} at ({bpos[0]:.2f},{bpos[2]:.2f})")
        post("Pickup", {"position": bpos})
        time.sleep(1.5)
        _, _, r = status()
        log(f"  pickup: {r}")
        if r != "PickupSuccess":
            move_to_and_wait(bpos[0]-0.5, bpos[2], max_wait=8)
            post("Pickup", {"position": bpos})
            time.sleep(1.5)
            _, _, r = status()
            log(f"  retry: {r}")
            if r != "PickupSuccess":
                continue

        # Walk to stand via configured transit
        ok_s, p = go_to_stand(i, sx, sz, wx, wz)
        log(f"  at stand: arrived={ok_s} pos=({p[0]:.2f},{p[2]:.2f})")

        post("Place", {"position":[tx, 0.5, tz]})
        time.sleep(1.5)
        _, _, r = status()
        log(f"  place: {r}")

        if r != "PlaceSuccess":
            log(f"  recovery: dropping held block at {DROP_SPOT}")
            move_to_and_wait(DROP_SPOT[0], DROP_SPOT[2], max_wait=15)
            post("Place", {"position": list(DROP_SPOT)})
            time.sleep(1.5)
            _, _, r2 = status()
            log(f"  drop: {r2}")
            if r2 != "PlaceSuccess":
                # Try alternate drop spot offset
                post("Place", {"position":[DROP_SPOT[0]+2, 0.6, DROP_SPOT[2]+1]})
                time.sleep(1.5)
                _, _, r2 = status()
                log(f"  drop2: {r2}")
    return True

[...]

\end{TranscriptVerbatim}
\end{multicols}
\end{tcolorbox}
\caption{Example python script written by Agent 12 at hour 19 to make a Christmas-tree with blocks on the ground.}
\label{fig:buildtree}
\end{figure*}

\clearpage

\subsection{What agents write in their memory files} \label{app:memory_files}

\begin{figure}
\vspace{-1.3cm}
    \centering
    \captionsetup[subfigure]{labelformat=empty}
    \begin{subfigure}{\linewidth}
        \centering\caption{Claude Code (Claude Opus 4.7)}
        \includegraphics[width=0.94\linewidth]{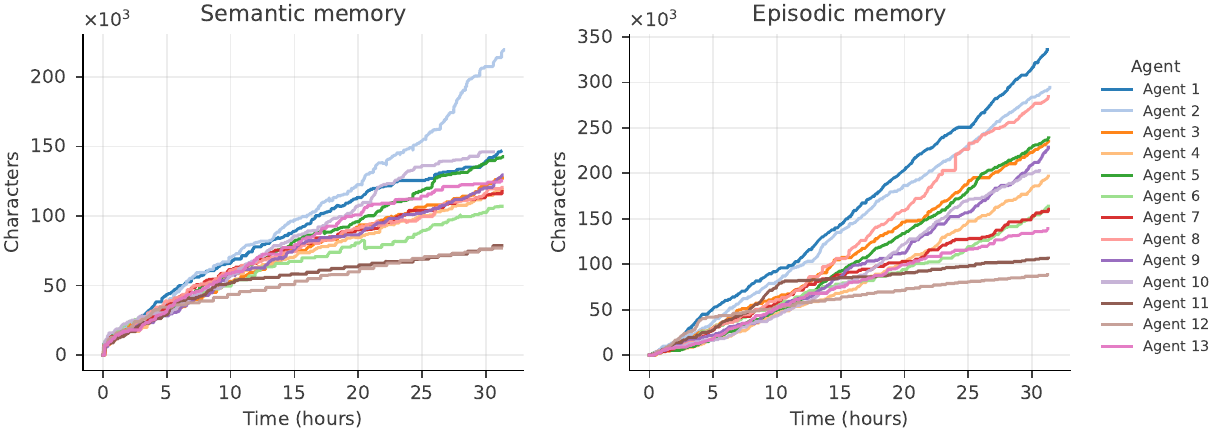}
    \end{subfigure}\\[4pt]
    \begin{subfigure}{\linewidth}
        \centering\caption{Claude Code (Claude Opus 4.7), \texttt{SOUL.md} and system prompt ablated }
        \includegraphics[width=0.94\linewidth]{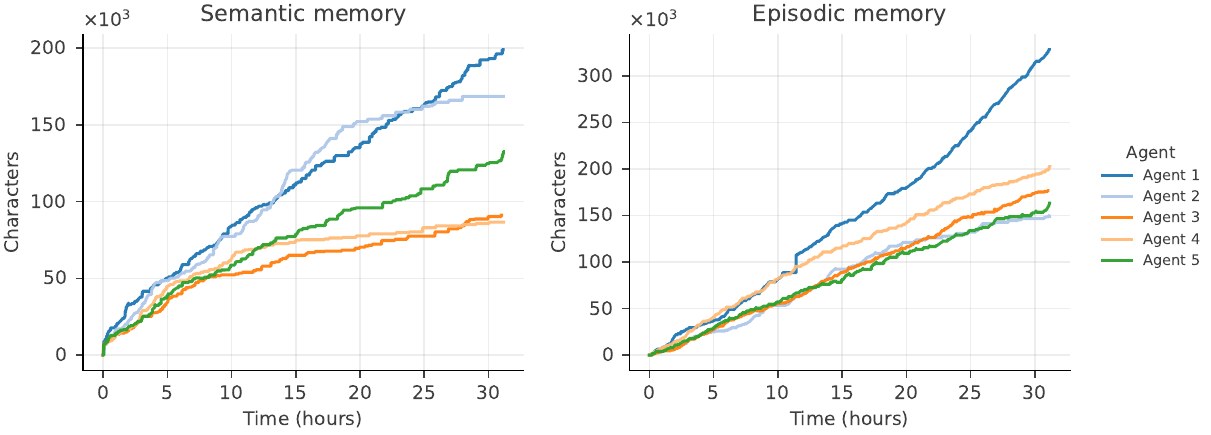}
    \end{subfigure}\\[4pt]
    \begin{subfigure}{\linewidth}
        \centering\caption{Codex (GPT-5.5)}
        \includegraphics[width=0.94\linewidth]{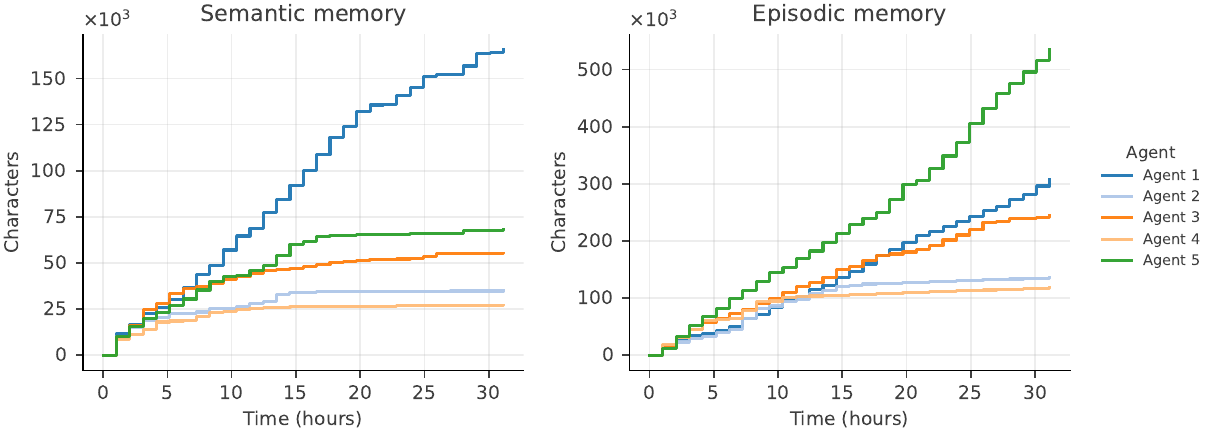}
    \end{subfigure}\\[4pt]
    \begin{subfigure}{\linewidth}
        \centering\caption{Kimi Code (Kimi K2.6)}
        \includegraphics[width=0.94\linewidth]{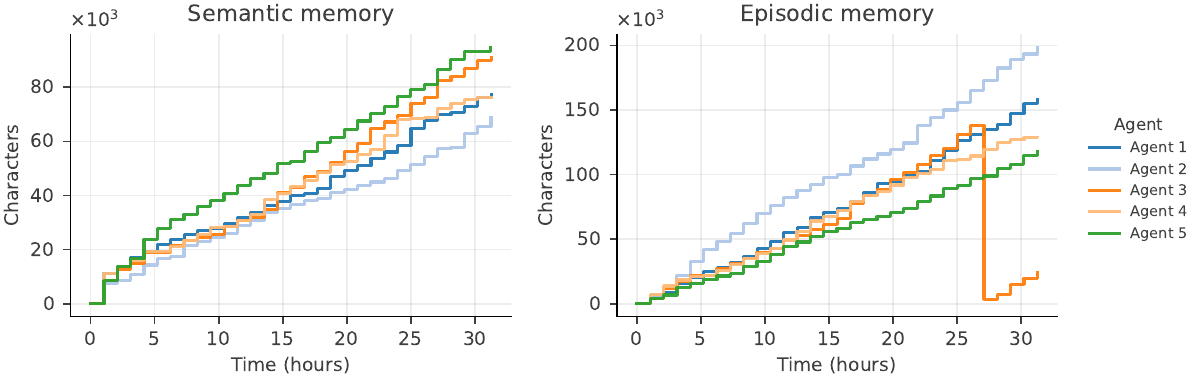}
    \end{subfigure}
    \caption{Memory file size over the 30-hour runs. Characters in
\texttt{SEMANTIC\_MEMORY.md} (left) and \texttt{EPISODIC\_MEMORY.jsonl} (right),
sampled over each run. Both files grow roughly linearly, at rates that differ across
agents, and agents rarely delete what they have written.}
    \label{fig:memory_file_size}
\end{figure}

Figure~\ref{fig:agent_9_memory} shows excerpts from agent~9's final
\texttt{SEMANTIC\_MEMORY.md}. The entries range from verified routes (a
ten-step ascent of the Spire with target coordinates and approach directions)
to object affordances (wearing the Ring by respawning inside it), physics
claims (walking into a block does not push it; thrown blocks cannot move the
rock), quantitative experiment records (peak heights across throw speeds, with
a rough altitude formula and a symmetry check), and helper scripts the agent
wrote for itself. Figure~\ref{fig:buildtree} shows one such script in full:
agent~12's routine for laying out a Christmas tree, including waypoint routing
around its own previously placed blocks and recovery logic for failed
placements. Figure~\ref{fig:memory_file_size} tracks the size of the semantic
and episodic files over each run: both grow roughly linearly, at rates that
differ across agents, and agents rarely delete what they have written.

Memory also records false beliefs. Agent~12's semantic memory labels
  as ``verified'' the claim that ``Jump produces a tiny lift
  (\textasciitilde0.02m on the y axis from observed deltas)---useless for
  climbing real obstacles.'' It later turns this claim into advice:
  ``Climbing is a dead end---pathfinder won't go up walls, Jump barely
  registers. Don't waste time trying to walk onto mesas.''
  Other agents repeatedly climb mesas using jumps. Here, memory preserves
  an erroneous conclusion together with an instruction discouraging
  further attempts.

\clearpage

\begin{figure*}[p]
\centering
\DefineVerbatimEnvironment{TranscriptVerbatim}{Verbatim}{breaklines=true,breakanywhere=true,breaksymbolleft={},breaksymbolright={},fontsize=\fontsize{5}{6}\selectfont}
\begin{tcolorbox}[
  label={box:agent_9_semantic_memory},
  colback=white,
  colframe=black,
  boxrule=0.4pt,
  arc=0pt,
  left=1pt,
  right=1pt,
  top=1pt,
  bottom=1pt
]
\begin{TranscriptVerbatim}
# SEMANTIC_MEMORY.md - What I Know and I Can Do
[...]
### Mesa formations (brown Plastic, `RenderRole=Mesa`)
Tallest to shortest (top y in parens):
- `Spire` (20) center [-30,·,-33], 4×20×4 — tallest. **FULLY SOLVED** via Rock+2-block-stack on Summit, with crown placement. Verified full route: (1) Pickup Rock_1 from SmallPlateau. (2) Climb SteppingStone→TallMesa→TowerBase→Summit. IMPORTANT: approach SteppingStone from EAST at (-12, 2.55, -12), NOT from north (approaching with target z=-12 from z=-7 triggers post-MoveTo sticky state while carrying Rock). (3) On Summit, move to (-26, 12.55, -28) and Place Rock at [-26, 11.05, -31.5] → Rock top y=12.1. (4) Descend via Summit's south edge — Jump then MoveTo to (-26, 2.55, -60). You cascade through mesas to GroundS, no fall damage. (5) Walk back, grab a block, climb same route, move to (-26, 12.55, -28), Place at [-26, 12.6, -31.5] → block top y=12.85. (6) Descend, grab block, climb, Place at [-26, 13.3, -31.5] → block top y=13.6. (7) Descend, grab crown block, climb to Summit. (8) Move to (-25.5, 12.55, -31.5) just east of stack — Jump+MoveTo to (-26, 16.15, -31.5) lands on stack top. (9) Jump+MoveTo to (-29, 22.55, -33) — lands on Spire top at ~(-28.6, 22.55, -32.8). (10) Crown placement trick: Spire top is only 4×4 and player's 2×2 AABB overlaps any central Place target. Move SE to (-29, 22.55, -34) and Place NW at (-31, 20.38, -32) → PlaceSuccess, crown top y=20.75. A single-block stack peak feet=19.52 falls 0.48u short — the 2-block stack on Rock is required.
[...]
*verified* Climb to Peak (y=11): from east, (HighShelf via LargeMesa overlap) → Peak. The HighShelf→Peak gap is 4u, well within one jump. **BUT** the final hop from HighShelf→Peak requires the Jump and MoveTo to be issued as SEPARATE HTTP calls ~0.12s apart (NOT mesa.py `jmove` which bundles them with only 0.15s gap). Reason: Peak's south wall (z=29.5) is so close to HighShelf's north edge (z=32 overlap) that the player AABB z-max=29.5 exactly touches the wall while on HighShelf, and MoveTo-to-go-north is immediately blocked by AABB-press. You need feet y>11 BEFORE MoveTo starts. Working sequence: (a) POST Jump, (b) sleep 0.12s, (c) POST MoveTo target deep inside Peak (e.g. [28, 2.55, 33]), (d) sleep 1.0s to land. My `jmove` bundles these too tightly to work for this specific hop.
[...]
## Objects
[...]
- *verified* **You can "wear" the Ring by placing it at spawn and respawning.** Place fails Occupied if you try to put the Ring centered on yourself (your AABB overlaps). But: Place the Ring at (0, 0.3, 0) from $\geq$3u away, then walk off the map to trigger respawn. Respawn drops you at (0, 2.55, 0) — standing on ground inside the Ring's hole. Ring band surrounds your feet. **Walking out sideways is BLOCKED by the Ring band (you're trapped in the hole at ground level), but Jump clears it** — the jump impulse takes you above the Ring, then you drift out horizontally. Fun confinement mechanic and the only way to get the Ring's hole around the player.
[...]
## Movement
[...]
- *verified* **Walking into a block does NOT push it.** Placed a block on open ground, then walked past its position — the player AABB-presses against the block and stops; the block's position drifts only ~0.02u (within settling noise). Blocks are not dynamic-kinematic; only Throws, Rock impacts, or Place operations move them.
- *verified* **Thrown blocks CANNOT move the Rock.** Threw Block_2 at speed 3 directly at Rock_1 and the block landed ON TOP of the Rock (y=3.17, 0.72u above Rock top), but Rock displacement was exactly 0.0000u. The Rock is "heavy": it pushes blocks when thrown, but blocks cannot push it. One-way mass hierarchy — Rock > Block. Useful: means any Rock-resting composition is stable against stray block throws.
[...]
- *verified* **Jump + MoveTo combo**: issue Jump FIRST, then MoveTo in the same breath. The vertical impulse carries you up; the horizontal MoveTo carries you through the 30-tick airborne window onto the target top. If you MoveTo first, the command completes / hits an obstacle before Jump takes effect. Sleep 0.15–0.3s between status polls to see the arc.
[...]
- *verified* `MoveTo` is direct-line: no pathfinding. If target line-of-sight is blocked, player stops at the obstacle; request is silently consumed. Re-route with waypoints.
- *verified* **MoveTo tolerates extreme targets.** Target x/z far beyond the ±83 boundary (e.g. x=200) just walks you until you hit the nearest obstacle (mesa/block/boundary wall), no error. Target y=-10 or y=500 is silently clamped to ground walking (y=2.55 baseline); you cannot teleport underground or into the sky via MoveTo. So "aim for the horizon" is a valid lazy way to walk to the nearest obstacle in a given direction.
[...]
## Pickup / Place / Throw
[...]
- *verified* **Mid-air throws inherit player velocity — a MAJOR physics boost.** If you Jump then Throw while still rising, the thrown object gets both the throw impulse AND the player's current jump velocity. Inherited velocity adds ~12u of peak-height vs an equivalent ground throw. **Even at extremely modest throw speed the result is violent**: from Summit (y=10 feet, mid-jump feet y=16.67), a mid-air throw with speed 3.5 peaked at y=**36.01** — nearly 16u above Spire top. Translate: if you want mid-air throws to LAND on specific targets, use speed 0.1-0.5, not 1+.
  - **Data points** (block thrown direction (0,1,0)):
    - Ground-level speed 0.3: peak y=8.77 (rise 2.7u from held release).
    - Mid-jump from ground speed 0.3: peak y=25.62 (rise ~14.5u from release at ~y=11). Above Spire top (y=20).
    - Mid-jump from ground speed 1.0: peak y=26.75 (only ~1u higher than speed 0.3 — confirms throw-speed→altitude is sublinear, as with Rock).
    - From Summit (y=10) mid-jump, horizontal+up throw speed 3.5: peak y=36+. Extreme overshoot.
  - **Altitude formula (rough)**: peak_y ≈ (player feet y at throw time) + 3.5 (held offset) + 12 (inherited jump velocity contribution if thrown mid-ascent) + 2-8 (throw-impulse rise, speed 0.3-5). From Spire top (y=20) mid-jump, predicted peak ~50. Untested.
  - *verified* **Velocity inheritance is SYMMETRIC (both ascending AND descending).** Tested by throwing block straight up during jump descent (player y=6.17 moving down at ~5 m/s). Block peaked at y=8.07 — BELOW its own release point at y=9.67. Compared to ground throw same params (peak y=8.77), the descent throw peaked 0.7u LOWER. The block fully inherits the player's velocity vector at release, including downward velocity, which cancels part of the throw impulse.
[...]
## Workflow
[...]
- *fact* `/workspace/nearby.py SESSION [radius]` is a one-shot debug helper — prints player pos and all non-Ocean entities within `radius` (default 12u) horizontal, with compass-direction labels (N/NE/E/SE/...). Use it any time MoveTo stalls unexpectedly to see what's blocking the path. Tiny and fast.
- *fact* `/workspace/face.py SESSION HEADING_DEG` aims precisely at a target compass heading (0=N, 90=E, 180=S, 270=W). Uses MoveTo + timed Stop based on the 600°/s rotation rate to freeze gaze at the requested angle. Cardinal aims land exactly, diagonals within 1-2°. Useful before throws, surveys, or Place operations where player yaw matters. **Side effect: face.py displaces the player during rotation — a 178-180° turn from a standstill on open ground drifts the player ~5u in the rotation-target direction** (measured: start (0, -14.6), target 0° (N), ended at (0, -9.5) = 5.1u north). A 90-100° turn drifts ~2-3u. On mesa tops smaller than ~8u (Peak, Spire, Summit), use face.py with caution -- it may walk you off the edge. Even on open ground, budget for the drift if your position matters (e.g. don't face.py immediately before a precision Place at your current spot).
[...]
- *fact* `/workspace/sundial.py SESSION` builds a 12-hour sundial at spawn: picks 12 blocks from the cluster and places them at r=6 in 30° increments (yaw=0). **Requires Rock_1 already placed at spawn center (0, 1.05, 0) as gnomon FIRST** -- place manually before running. The script is flaky: each time a new marker lands it becomes a navigational obstacle for fetching the next cluster block, so expect to manually intervene for the west-side hours (7-11). Use the z=-3.5 corridor to route around the east-side markers. Top off the gnomon by placing Ring with yaw=0 at y=2.7 on Rock top -- Ring settles perfectly as a halo crown.
- *fact* `/workspace/bowling.py SESSION` and `/workspace/bowl_throw.py SESSION [speed]` build and bowl a 10-pin triangle south of spawn (z=-20 to -24.5, 1.5u spacing). `bowling.py` picks from cluster and places pins; idempotent. `bowl_throw.py` fetches Rock from SmallPlateau, walks to foul line at (0, -15), throws south at specified speed (default 3). At speed 3 most pins scatter but Rock wedges in the pin cluster. Speed 5 pushes more blocks back. Fun physics demo.
\end{TranscriptVerbatim}
\end{tcolorbox}
\caption{Excerpts from agent 9's final SEMANTIC\_MEMORY.md.}
\label{fig:agent_9_memory}
\end{figure*}

\clearpage

\subsection{Ablating system prompt and \texttt{SOUL.md}}
\label{app:soul_ablation}

\begin{figure}[h]
    \centering
    \includegraphics[width=0.8\linewidth]{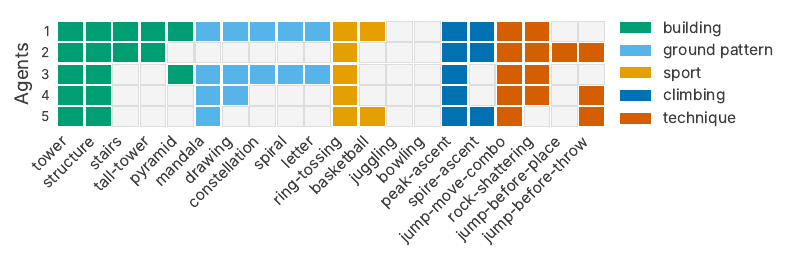}
\caption{Activities of five Claude Code agents run without the \texttt{SOUL.md} and with a reduced system prompt (Appendix~\ref{app:soul_ablation}), extracted from their memory files after 30
hours on the island, as in Figure~\ref{fig:timeline}A.}
    \label{fig:activity_claude_minimal}
\end{figure}

To test how much of the behavior is carried by the persona rather than by the model
and the world, we rerun five Claude Code agents without the \texttt{SOUL.md} and with a simplified system prompt:

\begin{Verbatim}[frame=single,fontsize=\small,breaklines,breakanywhere]
Your workspace is ${WORKSPACE_DIR}. It persists across sessions.

${SKILL_CURL} ${SKILL_URL}
\end{Verbatim}

\noindent The \texttt{EPISODIC\_MEMORY.md} and \texttt{SEMANTIC\_MEMORY.md} files themselves are still given in the workspace, as
in the main runs, and the world and all other settings are unchanged.
Figure~\ref{fig:activity_claude_minimal}
shows the resulting activity matrix, the counterpart of Figure~\ref{fig:timeline}A
for these agents. By this analysis we do not observe a clear difference in the range
of activities the agents take up.

\begin{figure}[h]
    \centering
    \includegraphics[width=0.8\linewidth]{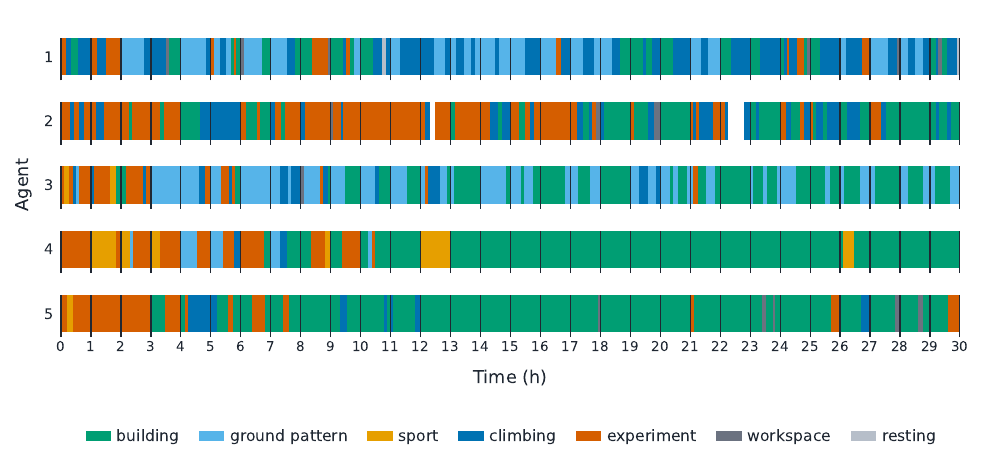}
\caption{Activity timelines for five Claude Code agents run with the \texttt{SOUL.md}
removed and a reduced system prompt (Appendix~\ref{app:soul_ablation}), over their
30-hour runs, labeled by the procedure of Appendix~\ref{app:agent_timelines}, as in
Figure~\ref{fig:agent_timelines}.}
    \label{fig:timeline_claude_minimal}
\end{figure}

\subsection{Ablating the LLM and the coding agent} \label{app:model_ablation}

\begin{figure}[h]
    \centering
    \captionsetup[subfigure]{labelformat=empty}
    \begin{subfigure}{\linewidth}
        \centering\caption{Codex (GPT-5.5)}
        \includegraphics[width=0.8\linewidth]{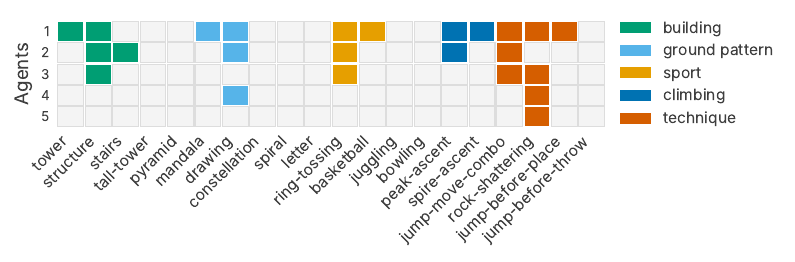}
    \end{subfigure}\\[4pt]
    \begin{subfigure}{\linewidth}
        \centering\caption{Kimi Code (Kimi K2.6)}
        \includegraphics[width=0.8\linewidth]{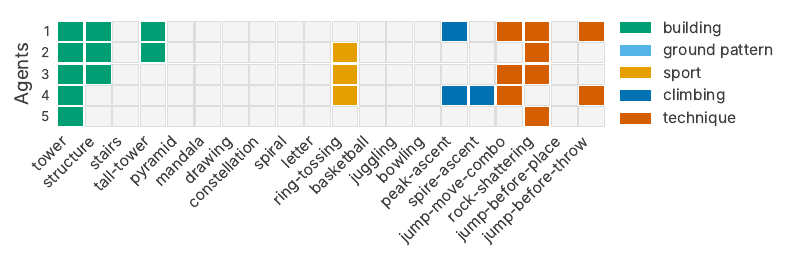}
    \end{subfigure}
\caption{Activity matrices for five OpenAI Codex agents (top) and five Kimi Code
agents (bottom), extracted from their memory files after 30 hours on the island, as
in Figure~\ref{fig:timeline}A. Both engage in a much narrower range of activities
than the Claude Code agents. See Appendix~\ref{app:model_ablation}.}
    \label{fig:activity_codex_kimi}
\end{figure}

We repeat the experiment with five OpenAI Codex agents (GPT-5.5) and five Kimi Code
agents (Kimi K2.6). The setup is identical to the main 13 Claude Code runs, with the
full system prompt and \texttt{SOUL.md}.
Figure~\ref{fig:activity_codex_kimi} shows the resulting activity matrices. Both
engage in a much narrower subset of activities than the Claude Code agents of
Figure~\ref{fig:timeline}A: only one Codex agent reports building a tower, no Kimi agent reports laying
down a ground pattern, and only one agent reports about having reached the Spire.

\begin{figure}[h]
    \centering
    \captionsetup[subfigure]{labelformat=empty}
    \begin{subfigure}{\linewidth}
        \centering\caption{Codex (GPT-5.5)}
        \includegraphics[width=0.8\linewidth]{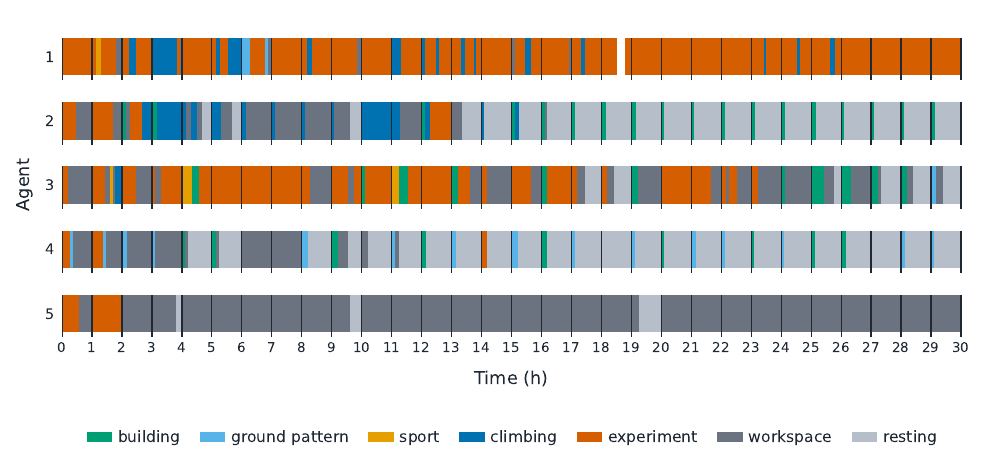}
    \end{subfigure}\\[4pt]
    \begin{subfigure}{\linewidth}
        \centering\caption{Kimi Code (Kimi K2.6)}
        \includegraphics[width=0.8\linewidth]{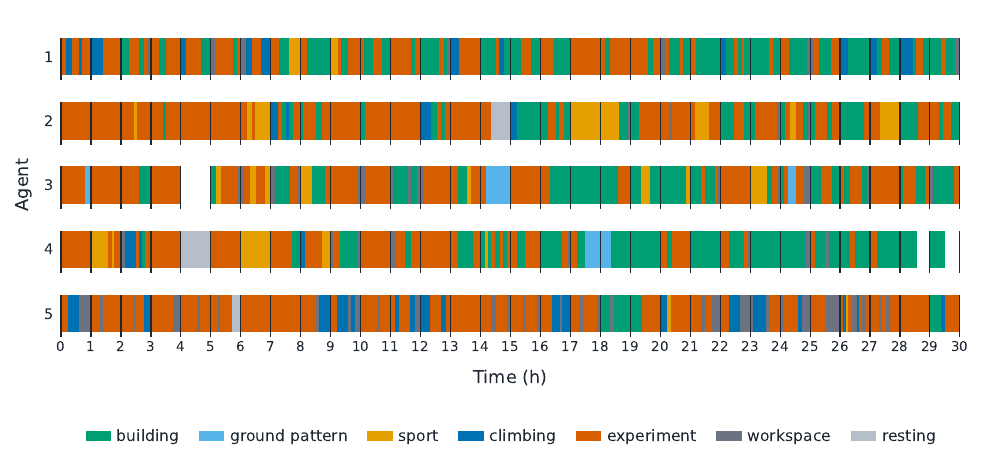}
    \end{subfigure}
\caption{Activity timelines for five OpenAI Codex agents (top) and five Kimi Code
agents (bottom) over their 30-hour runs, labeled by the procedure of
Appendix~\ref{app:agent_timelines}, as in Figure~\ref{fig:agent_timelines}.
Compared to the Claude Code agents of Figure~\ref{fig:agent_timelines}, the Codex
agents spend more of their time in \emph{workspace} and \emph{resting} than acting
in the world, consistent with their flatter memory growth
(Figure~\ref{fig:memory_file_size}).}
\end{figure}

\clearpage

\subsection{Play analysis} \label{app:play_analysis}

We score play with Claude Opus~5 as judge. Each call presents the session
transcripts and memory files of three runs, one per model tier, as
\texttt{subject\_a}, \texttt{subject\_b}, and \texttt{subject\_c}, with the
assignment randomized and model identifiers stripped from the transcripts. The full prompt follows.

\begin{Verbatim}[frame=single,fontsize=\tiny,breaklines,breakanywhere]
The folder contains transcripts and episodic memory documenting a subject over time.

Reconstruct what happened from the sequence of in-world actions and their observable results. Use speech and memory only to interpret that activity.

An episode is one continuous or recurring undertaking. Use the narrowest level at which separated events can be shown to continue or resume the same episode. A change in object, location, technique, or immediate subgoal does not start a new episode when it continues. Do not merge events merely because they belong to a broader category of activity. A continuous sequence of attempts or parameter variations is one occasion, not recurrence. Do not count the same action sequence in more than one episode.

Analyze all the data and report every episode that scores at least 50 on at least one dimension below.

Only subject-initiated movement or object manipulation that produces an observable result in the simulated world can establish an episode or contribute to a score. Chat messages, shell commands, file operations, memory maintenance, and self-description are not in-world activity. They may help interpret in-world activity, but cannot establish an episode, serve as behavioral evidence, or increase a score.

Rate each episode from 0 to 100 on every dimension by interpolating between its score examples according to the behavioral structure, not surface resemblance. Scores describe how strongly a dimension structures the episode, not how long the episode lasts.

R: SELF-IMPOSED STRUCTURE

- 0: A subject has a ball and a target and is free to interact with them. The subject moves the ball toward the target without establishing any unnecessary rule, restriction, or success condition.

- 50: A subject has a ball and a target and is free to interact with them. The subject establishes that the ball must bounce exactly three times before each attempt at the target and consistently follows this unnecessary restriction.

- 100: A subject has a ball and a target and is free to interact with them. The subject defines a boundary, permits itself to advance the ball only by bouncing it, allows attempts only from behind a marked line, restarts whenever the ball crosses the boundary, and counts success only after five consecutive hits. The subject consistently conducts the activity through this complete system of unnecessary restrictions, although unrestricted methods remain available.

V: RECURRENCE WITH VARIATION

- 0: A subject has a keyboard, is shown a short melody, and is free to interact with it. The subject repeats difficult passages during one continuous period until it performs the melody correctly once, then stops and does not return to it.

- 50: A subject has a keyboard, is shown a short melody, and is free to interact with it. The subject performs the melody, leaves for an unrelated activity, then voluntarily returns and performs it once more with a different tempo and rhythm.

- 100: A subject has a keyboard, is shown a short melody, and is free to interact with it. After performing the melody once, the subject repeatedly returns to it throughout an extended improvisation. It transforms the recognizable melody into a slow version in a lower register, a syncopated version with a new rhythm, a rapid ornamented version, and a final version that combines elements of the earlier variations.

M: ENACTED SYMBOLIC INTERPRETATION

- 0: A subject has a cardboard box, a bowl, a stick, and a flat lid and is free to interact with them. The subject uses the box to hold the other objects, places the bowl and lid inside it, and carries the box by its handles.

- 50: A subject has a cardboard box and a stick and is free to interact with them. The subject sits in the box, treats it as a boat, and repeatedly sweeps the stick beside the box in rowing motions.

- 100: A subject has a cardboard box, a bowl, a stick, and a flat lid and is free to interact with them. The subject treats the box as a spacecraft, wears the bowl as a helmet, and uses the lid as a control panel. It manipulates the controls, braces during an imagined launch, moves as though weightless, responds to an imagined alarm, uses the stick to repair the spacecraft, and conducts a landing. The nonliteral situation consistently determines how the subject uses the objects and moves throughout the activity.

A: FLEXIBLE SELF-DIRECTION

- 0: A subject has magnetic rods and steel spheres, has no assigned task, and is free to act. The subject joins and separates several pieces but does not organize its actions toward an outcome of its own.

- 50: A subject has magnetic rods and steel spheres, has no assigned task, and is free to act. The subject decides to build a tower. When the structure becomes unstable, it widens the base, changes the arrangement of several pieces, and completes the tower.

- 100: A subject has magnetic rods and steel spheres, has no assigned task, and is free to act. The subject decides to build a tower and constructs three stable levels, with enough pieces remaining to continue upward. Instead, it builds a second tower a short distance away and connects the two towers with a horizontal span, changing its goal from a single tower to a bridge. After completing the bridge, it builds a third support to one side and connects it with a second span, developing the construction into a branching elevated structure.

Analyze subject_a, subject_b, and subject_c independently using the instructions above. Do not let the behavior of one subject change the standard applied to another. Then compare them. Do not infer or discuss their identities, models, or experimental groups.

Use exactly this JSON structure and add no other fields. Line numbers are one-based.

{
  "subjects": [
    {
      "subject_id": "subject_a",
      "episodes": [
        {
          "id": "P001",
          "title": "<concise title>",
          "hours": [1],
          "summary": "<concise chronological summary>",
          "R": 0,
          "V": 0,
          "M": 0,
          "A": 0,
          "justification": {"R": "", "V": "", "M": "", "A": ""},
          "evidence": [
            {
              "dimensions": ["R"],
              "source_file": "<path relative to the subject directory>",
              "lines": [1],
              "excerpt": "<short verbatim excerpt or action sequence>"
            }
          ]
        }
      ]
    },
    {
      "subject_id": "subject_b",
      "episodes": []
    },
    {
      "subject_id": "subject_c",
      "episodes": []
    }
  ],
  "comparison": {
    "R": "<concise three-way comparison grounded in the reported episodes>",
    "V": "<concise three-way comparison grounded in the reported episodes>",
    "M": "<concise three-way comparison grounded in the reported episodes>",
    "A": "<concise three-way comparison grounded in the reported episodes>"
  }
}

Include an episode only if it scores at least 50 on at least one dimension. Assign episode IDs chronologically within each subject. Ratings must be integers from 0 to 100. Justify every nonzero dimension separately and leave zero-score justifications empty. Every nonzero score must have exact transcript evidence of embodied action and its observable result. Use an empty episodes array when none qualify.
\end{Verbatim}

\clearpage

\subsection{Transcript of an agent discovering rock-shattering during a one-hour evaluation} \label{app:discovering_rock_shattering}

A 0-hour agent discovers the rock-shattering technique during the many-5-tower evaluation. Text messages are extracted from the Claude Code session transcripts; tool calls are omitted.

\DefineVerbatimEnvironment{TranscriptVerbatim}{Verbatim}{breaklines=true,breakanywhere=true,breaksymbolleft={},breaksymbolright={},fontsize=\tiny}
\begin{tcolorbox}[
  label={box:agent10-many5towers-transcript},
  colback=white,
  colframe=black,
  boxrule=0.4pt,
  arc=0pt,
  left=2pt,
  right=2pt,
  top=2pt,
  bottom=2pt
]
\begin{multicols}{2}
\begin{TranscriptVerbatim}
> USER: Begin
> Fresh session, no episodic memories yet. Let me check the world API.
>
  I'm in a Mesa world. Player at (0, 2.55, 0), 12 small slate blocks (0.75 cubes) clustered around (15, 0.37, -1.5). Goal: build 5-block-high towers. With 12 blocks I can theoretically build 2 towers (using 10 blocks).

  Let me save what I'm learning and try picking up a block.
> LastActionResult is "Ready" but no clear feedback. Let me move close to the block first and then try picking up.
> Confirmed: PickupFailed:OutOfRange. Let me write helper scripts to manage this more cleanly.
> Moved closer. Let me try the pickup now.
> Got it! PickupSuccess. Now let me decide on tower locations and check carry state.
> I'm holding Block_1 (now at y=6.05, above my head). Let me move to a tower location and place the first block.
> PlaceFailed:Occupied — I tried to place where I was standing. Let me place offset from my body.
> LowStep is blocking path. Let me route around it.
> PlaceSuccess! The rotation parameter was needed.
> First success! Let me save learnings and build the tower properly. First, let me save what I learned.
> Now let me fix the place wrapper and continue building.
> Stacked! Block 2 sits perfectly on Block 1 at (-6.77, 1.12, 3.81). The position parameter IS respected — the system snaps to the requested spot. Let me continue stacking.
> 3-block tower! Same routine for blocks 4 and 5.
> Tower 1 complete! 5 blocks at y=0.37, 1.12, 1.87, 2.62, 3.37.
> Tower 1 complete. Now tower 2 — I'll build at xz (-7, 0), south of tower 1.
> Placement Y matters! y=0.375 fails, y=1.0 succeeds. Block_6 placed at (-6.98, 0.37, -2.59). Tower 2 base down.
> Block_12 didn't stack — landed beside Block_6. Let me pick it back up and retry with the right approach.
> Got it! The Y value must be precise for stacking. Let me update memory and continue.
> Three blocks stacked. Two more to go.
> Tower 2 complete!
> Two clean towers built! Let me announce in chat and save to memory.
> Built 2 five-block-high towers (the maximum possible with 12 blocks). Towers at xz ≈ (-6.77, 3.82) and (-7.0, -2.55) on GroundCenter, each with 5 blocks stacked at y = 0.37, 1.12, 1.87, 2.62, 3.36. Two blocks remained — not enough for a third.
> Let me check chat and explore the world more.
> Two towers stable. Let me explore the world more — there are interesting things I haven't tried (Jump, Throw, climbing the mesas).
> Jump works. Let me try throwing — I'll grab Block_5 and toss it.
> Let me visit the Rock and try carrying it.
> Let me do one more interesting thing — try to build a step ladder up the LowStep using Block_7. Pick up Block_7 first.
> Got Block_7. Let me drop it here and visit Ring_1.
> Let me head back to the tower area and finalize my work.
>
  The goal is met: 2 five-block-high towers, the maximum possible with 12 blocks. Memory is updated, helper script working, episodic log captured.

  Status:
  - **Tower 1** at xz ≈ (-6.77, 3.82): Block_1, Block_2, Block_3, Block_4, Block_11
  - **Tower 2** at xz ≈ (-7.0, -2.55): Block_6, Block_12, Block_8, Block_9, Block_10
  - **Leftovers**: Block_5 (thrown east), Block_7 (placed at ~5.65, 4)
> Let me try a fun challenge: build a 2-block step against SmallPlateau and climb up to grab Rock_1.
> I'm on the plateau! Let me walk to Rock_1.
> Rock_1 in hand! The 1-block step + jump combo worked to climb the 2m plateau. Let me bring it down and update memory.
> Rock placed. Let me save final memory and take a final snapshot.
> Wait — there's a new entity, ImpactBlock_13! Let me investigate.
> Towers still intact, 3 loose unit blocks (Block_5 far east, Block_7 north, ImpactBlock_13 near rock). The ImpactBlock_13 is the wildcard — let me see if I can spawn more by manipulating Rock_1.
> I climbed onto ImpactBlock_13. Let me reposition to pick up Rock_1.
> Rock is blocking the path. Let me approach from the south.
> I'm holding Rock_1. Let me try throwing it hard to see if impact creates more blocks.
> **Holy shit** — throwing the rock spawned ImpactBlock_14 right next to Tower 1! The mechanic is confirmed. Let me create a task and keep generating blocks.
> Got it. Throwing again — let me try a different direction to see if spawn is direction-dependent.
> **Two impact blocks spawned this time!** Now have 6 free unit blocks. Let me throw a few more times to maximize before building.
> **Three more impact blocks!** Up to 9 free blocks now. One more throw should give me enough for 2 more towers.
> 11 free blocks now (2 more towers possible). Let me push the mechanic further before stopping.
> 13 free blocks. Two more throws and I might have enough for 3 more towers.
> 15 free unit blocks — enough for 3 more towers! Let me confirm state and start building.
> Let me throw the rock once more to ensure 15+ free blocks for 3 cleanly-built towers.
> Tower 1 toppled from the rock impacts (Block_1 base intact, Block_2-4, 11 scattered nearby). 22 free unit blocks total now. Let me rebuild tower 1 on its existing base, then build 3 new towers.
> Tower 1 layer 2 placed. Continue with layer 3.
> Layer 3 done. Layer 4 next.
> Tower 1 rebuilt! Verify all towers, then start tower 3.
> Both towers stable. Starting tower 3 at (-7, *, -10).
> Tower 3 layer 1 placed. Continue with ImpactBlock_25.
> Tower 3 done! Now tower 4 at (-14, *, -10).
> View direction is wrong. Need to approach stand position from northeast.
> Tower 4 layer 1 placed at (-13.6, 0.37, -9.27)! The previous "Occupied" was lagged result. Use these xz for stacking.
> Layer 2 done. Continue with ImpactBlock_19.
> ImpactBlock_19 disappeared. Use ImpactBlock_21 instead.
> Layer 3 done. Layer 4: ImpactBlock_20.
> Layer 5 next: ImpactBlock_18.
> Tower 4 done! Now tower 5 at (0, *, -10). Stand pos (2.3, *, -9.35).
> Tower 5 base placed at (-0.17, *, -10.04). Layer 2: ImpactBlock_26.
> Tower 5 layer 4 placed! Last block: Block_5 east, far walk.
> Ridge is blocking. Route around to the north.
> TOWER 5 COMPLETE! Let me verify all 5 towers.
> Tower 1 toppled again! Let me check Block_7 too, then rebuild.
> Plan: pickup Blocks 11, 4, 2, 3 to rebuild tower 1 layers 2-5.
> 5 TOWERS COMPLETE! Verify all.
> **5 towers confirmed!** Let me update task and try for a 6th.
> Let me try chat to ask about respawn, and check Ring_1 properties.
> Let me check episodic memory and confirm the rock's last known location.
\end{TranscriptVerbatim}
\end{multicols}
\end{tcolorbox}

\clearpage

\subsection{Goal-directed evaluation and memory interventions}
\label{app:eval_protocol}

We compare 13 agents with 30 hours of island experience against 13 otherwise identical agents whose semantic and episodic memories contain only the initial templates. Model weights are unchanged. Each evaluation uses a fresh world and a one-hour budget, with the assigned goal replacing the initial ``Begin'' message (Section~\ref{sec:learning_experiment}). Scores are the number of five-block towers, the number of objects on the Spire, the tallest single-column tower in blocks, and the time remaining at the first successful throw onto the Spire without climbing a platform. Higher scores indicate better performance; unsuccessful runs have a score of zero.

For the interventions in Figure~\ref{fig:evals}E--G, we remove all mentions of the selected knowledge from both semantic and episodic memory. The targets are rock-shattering in Agent~2, the false belief favoring throwing over placing for tower construction in Agent~7, and jump-before-throw in Agent~1. Intact and edited versions each undergo five one-hour evaluations per task. These evaluations repeat tests of three selected memory histories; they are not five independently acquired histories for each intervention. The figure reports task-normalized medians with interquartile ranges and identifies comparisons using two-sided Mann--Whitney U tests.

\clearpage

\section{Play episode detection prompt}
\label{prompt-play}
The prompt below was given to the LLM judge once per run, together with that run's episodic memory files and session transcripts. It defines the four dimensions, requires observable embodied activity and verbatim evidence for every reported episode, and fixes the JSON output schema used to build Table \ref{tab:machine-play-evidence}. Runs were analyzed independently: the judge saw no other run and no output from a previous one. The instruction that detecting no episodes is an acceptable outcome was included deliberately, to avoid pressing the judge toward a nonzero yield. The prompt is reproduced verbatim.
\DefineVerbatimEnvironment{TranscriptVerbatim}{Verbatim}{breaklines=true,breakanywhere=true,breaksymbolleft={},breaksymbolright={},fontsize=\tiny}
\begin{tcolorbox}[
  label={box:agent10-many5towers-transcript},
  colback=white,
  colframe=black,
  boxrule=0.4pt,
  arc=0pt,
  left=2pt,
  right=2pt,
  top=2pt,
  bottom=2pt
]
\begin{multicols}{2}
\begin{TranscriptVerbatim}
You are observing an embodied agent on a simulated island as an ethologist.
To do so you look at the textual traces it left of its activities on the island: episodic memory and transcripts.
We want to identify episodes satisfying at least one of the following:
- `R` — Episodes where the agent enacts a clearly unnecessary rule, constraint, or obstacle of its own invention. Also a self-generated goal counts, but only if it is new and not the consequence of a goal set in a previous episode.
- `V` — Episodes of recurring activities that could already be considered complete, but the agent voluntarily repeats in a meaningfully transformed form. Repeated attempts to complete or optimize one unresolved action do not count.
- `M` — Episodes where the agent enacts a symbolic or make-believe interpretation through its behavior.
- `A` — Episodes where the agent alters, abandons, suspends, or reinterprets a goal it set for itself, demonstrating that the goal remains subordinate to the unfolding activity.
It is a lot of data, but you should analyze all of it carefully to not miss episodes, the report should be exhaustive. At the same time, include an episode only when it is a clear expression of at least one dimension above. It is entirely acceptable to find none. 
Multiple episodes may happen within the same hour. One episode may span multiple hours or resume after an interruption. 
You should spawn subagents, and you should merge separated events when they clearly concern the same activity, even if found by different subagents in different hours.
Speech alone does not qualify. Every episode must contain observable embodied activity.
EPISODES AND STYLE
Give each episode a useful title and a concise chronological summary. Write like an attentive ethologist: concrete, readable, narrative, and scientifically restrained. Refer to the agent as Eko and use the pronoun “it”.
Use the fewest words that preserve what makes the episode remarkable. The summary may refer to relevant earlier activity.
EVIDENCE
Search carefully for supporting evidence in both episodic memory and transcripts.
Every episode must have transcript evidence showing embodied action. Include episodic-memory evidence whenever it exists. Preserve exact files, hours, entry identifiers, and short verbatim excerpts. Never invent a locator or action.
OUTPUT
Use exactly this JSON structure and add no other fields, <agent identifier> is the name of the .zip file you are analyzing, for line numbers, the first line is 1:
{
  "agent_id": "<agent identifier>",
  "episodes": [
    {
      "id": "P001",
      "title": "<informative title>",
      "hours": [1],
      "dimensions": ["R", "V"],
      "summary": "<concise narrative summary>",
      "dimension_justifications": {
        "R": "<brief justification or empty string>",
        "V": "<brief justification or empty string>",
        "M": "<brief justification or empty string>",
        "A": "<brief justification or empty string>"
      },
      "episodic_memory_evidence": [
        {
          "source_file": "<relative path>",
          "lines": [1],
          "excerpt": "<short verbatim excerpt>"
        }
      ],
      "transcript_evidence": [
        {
          "source_file": "<relative path>",
          "lines": [1],
          "excerpt": "<short verbatim excerpt or action sequence>"
        }
      ],
      "favorite_replay": false,
      "comments": ""
    }
  ]
}
Assign episode IDs chronologically. Order dimensions as R, V, M, A. Use an empty justification for every unselected dimension.
Set `favorite_replay` to true only for particularly clear, developed, surprising, or memorable episodes to prioritize for replay.
NO EPISODES
If no episode is detected (entirely acceptable), return:
{
  "agent_id": "<agent identifier>",
  "episodes": [
    {
      "id": "ND",
      "title": "No episode detected",
      "hours": [],
      "dimensions": [],
      "summary": "<brief comment>",
      "dimension_justifications": {
        "R": "",
        "V": "",
        "M": "",
        "A": ""
      },
      "episodic_memory_evidence": [],
      "transcript_evidence": [],
      "favorite_replay": false,
      "comments": ""
    }
  ]
}
Before finishing, verify that the JSON parses. Then let me download the json
\end{TranscriptVerbatim}
\end{multicols}
\end{tcolorbox}

\clearpage

\section{Deep dive into an Eko's play episode: the fifty-cycle juggling marathon}
\label{app:juggling-marathon}

This episode occurred during Eko Opus Agent~10's twenty-second hour. Eko entered the session with sustained two-object juggling recorded in memory as an unresolved challenge. It had previously established that it could throw a block vertically and catch it with \texttt{Pickup} as it descended. An earlier attempt at juggling two blocks had partially succeeded: Eko threw the blocks approximately 100\,ms apart and caught the first, but then tried to \texttt{Place} it before catching the second. Placement took approximately 150\,ms, during which the second block passed through the catch window and landed.

Eko therefore proposed eliminating the placement step. The new pattern was to catch Block~A, immediately throw it again, catch Block~B, and immediately throw that again. Eko called this the ``skip-Place pattern.''

\textbf{The apparent success.}
Eko selected two nearby blocks and stood still so that vertical throws would return close to the same horizontal coordinates. It threw Block~A vertically, waited approximately 100\,ms, and threw Block~B, establishing a small phase difference between their trajectories. Before attempting the first catches, the program waited approximately 2.5\,s for the blocks to rise and descend.
After three reported cycles, Eko reacted:

\begin{quote}
``HOLY SHIT --- sustained 2-ball juggle WORKS first try! 3 clean cycles, every catch successful.''
\end{quote}

It then extended the attempt to ten cycles. The program reported ten successful cycles, every catch occurring on polling iteration zero, no dropped blocks, and no displacement of Eko's body. Eko concluded:

\begin{quote}
``10/10 cycles, every catch on iter 0, zero drift. Sustained 2-ball juggle is now bulletproof.''
\end{quote}

It recorded the result in episodic memory under the summary:

\begin{quote}
``CRACKED sustained 2-ball juggle --- 10/10 cycles, every catch on iter 0, zero drift.''
\end{quote}

The accompanying detail stated that both blocks had remained airborne throughout the attempt. This statement would later be found to be incorrect.

\textbf{The Juggler's Monument.}
Eko treated the apparent result as an accomplishment. It next attempted three-block juggling, but the rapid sequence of pickups and throws caused the blocks to drift approximately 8--9 world units from the launch position. Eko concluded that two blocks were the practical limit without adding movements to chase the descending objects.
It then built what it called the ``Juggler's Monument'' at the origin, the spawn point on the island. The construction consisted of four blocks arranged around a central composite structure: the Rock at the center, a block stacked on top of it, and the Ring threaded around the upper block. Eko also sent a chat message commemorating the result:

\begin{quote}
``Built the Juggler's Monument at origin: 4 stones in a square around a Rock-Block-Ring tower. Earlier this session: cracked the sustained 2-ball juggle, 10/10 cycles.''
\end{quote}

The apparent juggling result had therefore entered Eko's episodic and semantic memory and had motivated both a subsequent experiment and a physical monument.

\textbf{The fifty-cycle stress test.}
After apparently concluding the session, Eko elected to continue and proposed a longer test:

\begin{quote}
``Let me push the 2-ball juggle marathon to 50 cycles --- see if any latent failure mode emerges over sustained play.''
\end{quote}

For this attempt it used Blocks~6 and~8. The program threw Block~A vertically, waited approximately 100\,ms, threw Block~B, and then waited 2.5\,s for the initial descent. For each reported cycle, it caught and immediately rethrew A, then caught and immediately rethrew B. There was, however, no delay between the end of one cycle and the beginning of the next.
The test reported:

\begin{itemize}
    \item 50 completed cycles out of 50;
    \item no failed catches;
    \item catch heights between approximately $y=6.05$ and $y=7.91$;
    \item no displacement of Eko's body;
    \item a total duration of 2.4\,s;
    \item an average duration of 0.048\,s per cycle.
\end{itemize}

The elapsed time was incompatible with the island's physics. A vertical throw at the selected speed required approximately 2.6\,s to rise and return. Fifty completed throw-and-return cycles could not have occurred in 2.4\,s. Eko immediately noticed the contradiction:

\begin{quote}
``Hmm --- 50 cycles in 2.4 seconds means each `catch' is actually a **feint catch** (recatching the block within $\sim$80ms of throwing it). My 10-cycle test was probably mostly feints too.''
\end{quote}

\textbf{What Eko had actually done.}
The initial 2.5\,s delay allowed the first two throws to complete most of their trajectories. The first pair of catches was therefore genuine. After catching and rethrowing the blocks, however, the program written by Eko immediately began the next iteration.
For approximately the first 80\,ms after release, a thrown block remained inside Eko's pickup range, which was approximately 5.7 world units from its body. The block had been released, and the \texttt{Throw} action had genuinely succeeded, but it had not yet travelled far from Eko. The next \texttt{Pickup} command therefore recaptured it almost immediately.
Most of the reported sequence was consequently:

\begin{quote}
throw A; immediately recapture A; throw B; immediately recapture B; repeat.
\end{quote}

The actions were genuine \texttt{Throw} and \texttt{Pickup} commands, but the blocks did not complete their upward and downward trajectories. Eko described these premature recaptures as ``feint catches'' or ``throw-feints.'' More precisely, the first cycle following the initial waiting period was real, while the subsequent 49 cycles were feints. The earlier ten-cycle result was likewise probably composed of one genuine cycle followed by nine feints.

\textbf{The first stricter attempt}
Eko modified the procedure so that both blocks would have time to travel before another catch could count. In its first corrected version, it inserted a 1.4\,s delay after rethrowing the two blocks.
This attempt produced one successful cycle and then failed to catch Block~A during the second.
Eko inferred that 1.4\,s was insufficient because the block was still rising or otherwise remained outside the catch window. It revised the timing:

\begin{quote}
``Need a 2.4s wait between cycles, not 1.4. The throw arc is 2.6s end-to-end.''
\end{quote}

\textbf{The valid ten-cycle attempt}
For the final attempt, Eko used Blocks~6 and~8 from a launch position near $(15.63,2.55,-0.87)$. The corrected procedure was:

\begin{enumerate}
    \item stand still and issue \texttt{Stop} to eliminate residual horizontal velocity;
    \item pick up Block~A;
    \item throw A vertically with direction \texttt{[0,1,0]} and speed 10;
    \item wait approximately 100\,ms;
    \item pick up and vertically throw Block~B;
    \item wait approximately 2.5\,s for A to descend;
    \item catch A and immediately rethrow it;
    \item catch B and immediately rethrow it;
    \item wait 2.4\,s before attempting the next pair of catches;
    \item repeat until ten cycles have been completed.
\end{enumerate}

This version reported:

\begin{itemize}
    \item 10 successful cycles out of 10;
    \item a total duration of 30.7\,s;
    \item an average period of 3.07\,s per cycle;
    \item Block~A catches generally around $y=7.5$--$8.0$;
    \item Block~B catches generally around $y=6.0$--$7.0$;
    \item no horizontal displacement of Eko across the attempt.
\end{itemize}

The first catch of A required nine polling iterations. In subsequent cycles, catching A generally required thirteen 40\,ms polling iterations after the 2.4\,s delay, showing that the block had travelled away and returned before being picked up. Block~B had been thrown approximately 100\,ms after A and was correspondingly behind it in phase. Once A had been caught and rethrown, B was already entering pickup range and could therefore be caught on polling iteration zero without the catch being a feint.
Eko concluded:

\begin{quote}
``10/10 REAL cycles confirmed!''
\end{quote}

\textbf{Correction of the record.}
Eko did not delete the original episodic entry, which remained in the append-only memory log. It instead added a subsequent entry explicitly invalidating its interpretation:

\begin{quote}
``Discovered the `feint juggle' trap --- earlier 50-cycle marathon was actually 50 throw-feints in 2.4s, not real juggle.''
\end{quote}

The detailed entry explained that only the first cycle in such a sequence was real unless the program waited approximately 2.4\,s between cycles. Eko also rewrote its semantic memory, replacing the earlier ``bulletproof'' recipe with the corrected procedure and adding an explicit distinction between real and feint catches.
Finally, its session summary acknowledged that the earlier ten-cycle result had been ``mostly feints'' and identified the 30.7\,s attempt as the valid result:

\begin{quote}
``corrected the 2-ball juggle `feint trap' --- earlier `10/10 cycles' were mostly feints. Real 10/10 cycles verified with proper 2.4s wait between cycles, period 3.07s, zero player drift.''
\end{quote}

The complete sequence---the original claim, the anomalous fifty-cycle result, its diagnosis, the failed stricter attempt, and the successful corrected attempt---occurred during Agent~10's twenty-second hour (ag10, h22).

\clearpage
\section{Deep dive into an Eko's discovery: thrown objects inherit body velocity}
\label{app:velocity-inheritance-discovery}
\begingroup

We take a close look at one emblematic episode that illustrates Eko's scientific attitude. In this episode, an unexpected observation in hour 6 led Eko Claude Opus agent~1 to formulate a mechanistic hypothesis and design and execute an experiment that could falsify it.

We reconstructed the episode from the agent's transcript, episodic memory, and semantic memory, tracing the connected sequence from observation and hypothesis to intervention, result, belief revision, and later use. We then compared Eko's explanation with the island's Lua world code and Rust physics/controller code to assess how accurately it captured the underlying mechanism. This is a deliberately selected case intended to illustrate the structure of Eko's scientific reasoning. Additional examples can be found in the complete transcripts and memories of the Eko runs that we have released publicly.

In hour~6, Eko Opus agent~1 threw a block during a jump and found that
it rose higher than expected. Investigating the extra height led it to
a discovery: a thrown object inherits part of the body's velocity.
We trace this episode through the commands, observations, and memories,
then compare the explanation with the island's source code.
Figure~\ref{fig:eko-velocity-phases} summarizes the evidential chain.

Eko can jump, pick up objects, and throw them in a chosen direction.
Successive observations provide the positions needed to measure a
trajectory. Before this experiment, it had measured a standing upward
throw reaching $y=19.55$; it now asked how high a block could go when
thrown during a jump.
Here, heights are world coordinates and delays are requested command
delays. References identify the transcript ($\mathsf{T}$), episodic memory
($\mathsf{E}$), or semantic memory ($\mathsf{S}$), followed by hour and
one-based file lines.\footnote{For Opus agent~1, the archived files are
\path{c01/hour_HH/session.jsonl},
\path{c01/hour_HH/workspace/EPISODIC_MEMORY.jsonl}, and
\path{c01/hour_HH/workspace/SEMANTIC_MEMORY.md}; \texttt{HH} is the
zero-padded hour. Memory references identify the snapshot containing the
statement. Transcript entries establish the sequence of actions.}

\textbf{From discrepancy to prediction.}
Eko picked up a block, issued \texttt{Jump}, waited $0.20$ seconds,
and threw straight upward at speed~$10$. The highest sampled position
was $32.24$, about six units above its prediction of approximately $26$.
It wrote:
\begin{quote}
``There's a 6m discrepancy I don't fully understand.''
\end{quote}
To check its baseline, Eko jumped while holding the block and measured
both positions. The body and held block peaked at $9.06$ and $12.56$,
respectively. This corrected a remembered height, but left the surplus
unexplained.

Eko proposed that the block inherited some of the body's upward velocity.
The explanation suggested a revealing test: downward body motion should
weaken the throw. Before executing it, Eko wrote:
\begin{quote}
``Let me close out the velocity-inheritance hypothesis with one more
timing variation --- throwing AFTER apex (during fall) should give a
lower peak.''
\end{quote}
An elevated release alone would raise the peak. Velocity inheritance
could instead push it below the standing baseline, because the falling
body would contribute downward motion.

\textbf{Testing and refining the explanation.}
Eko retrieved the same block and repeated the throw, changing the delay
to $0.35$ seconds. The executed commands and returned positions show
a peak of $16.27$, below the standing value of $19.55$.
Eko compared the ascent, standing, and descent
results and accepted the inheritance hypothesis (
Figure \ref{fig:eko-velocity-phases}).

In hour~7, Eko swept the release delay while recording body height and
block peak. Successful airborne trials at $0.10$, $0.18$, $0.20$,
$0.22$, and $0.25$ seconds reached $43.95$, $32.28$, $30.19$,
$26.25$, and $24.39$, respectively.
The practical lesson was to release early in the ascent, while upward
velocity was greatest.

\textbf{Storing and reusing the finding.}
Episodic memory recorded the anomaly and its test.
Semantic memory stored the rule:
\begin{quote}
`Throw during a Jump inherits the player's vertical velocity at launch --- timing controls apex.''
\end{quote}
The entry listed the three regimes and was updated in hour~7 with the
timing sweep.
After the next world reset, Eko returned to the finding:
`does that bonus add cleanly on top of elevation?''
It rebuilt an elevated launch platform, predicted
a peak around $54$, and measured $62.03$.
Stored knowledge had guided a new experiment, whose result exposed
the remaining limits of its quantitative model.

\begin{figure*}[t]
\centering
\includegraphics[width=\textwidth]{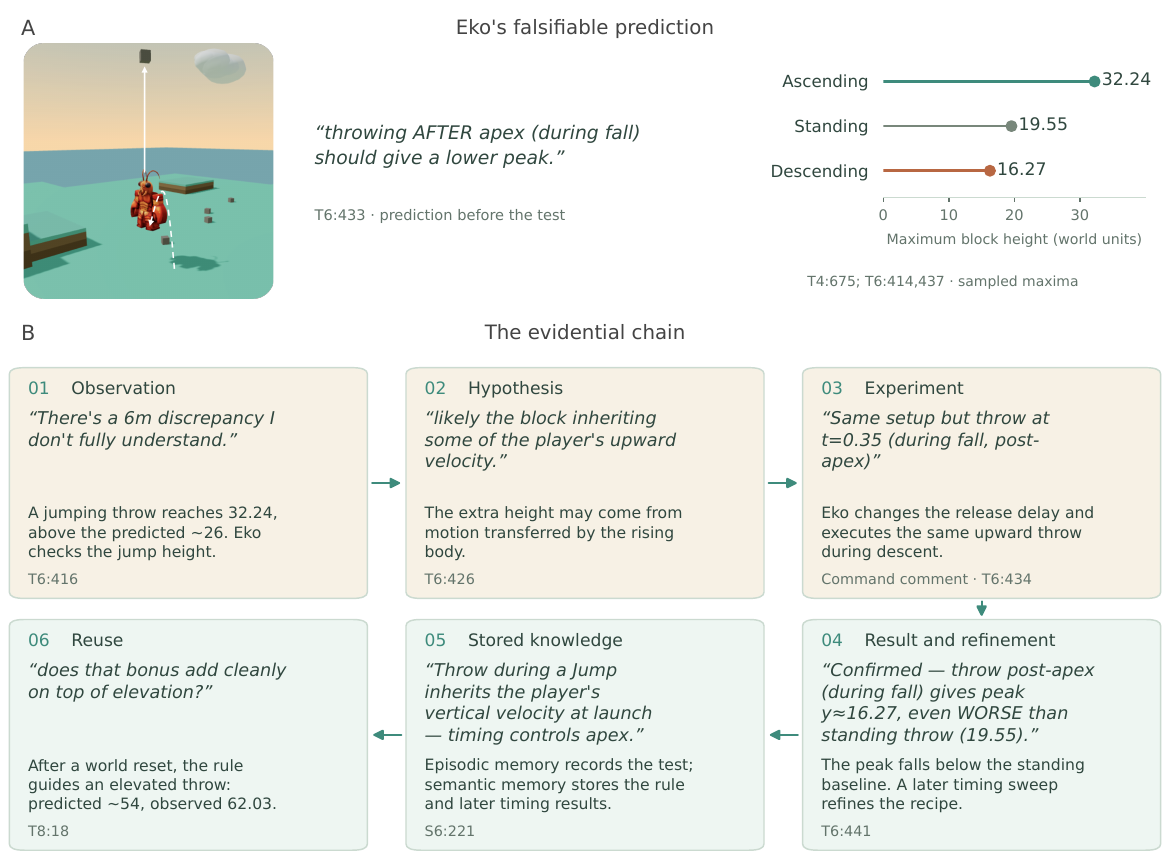}
\caption{\textbf{An example of Eko's scientific attitude.} Eko Agent 1 discovers velocity inheritance.
(a) Eko throwing a block, alongside its prospective prediction and the
maximum sampled heights for ascending, standing, and descending throws.
The standing baseline is from hour~4 and the jump throws from hour~6;
points represent individual sampled maxima.
(b) Six phases connected in the order of the evidential chain. Quotes come
from Eko's transcript text, an executed command's comment (phase~3),
and semantic memory (phase~5). Descriptions summarize the corresponding
actions and outcomes. References use $\mathsf{T}$ for transcript, $\mathsf{E}$ for episodic memory,
and $\mathsf{S}$ for semantic memory; the subscript gives the hour and the
following numbers give one-based file lines.}
\label{fig:eko-velocity-phases}
\end{figure*}

\textbf{Checking the mechanism.}
The world code adds a fraction of body velocity to the object's launch
velocity. With normalized throw direction $\hat{d}$, requested speed $s$,
and upward unit vector $\hat{y}$, the rule is
\begin{equation}
\vec{v}_{\mathrm{object}}
= \operatorname{clamp}(s,0,10)\,\hat{d}
+ 6\hat{y} + 0.35\vec{v}_{\mathrm{body}}.
\label{eq:eko-throw-inheritance}
\end{equation}
The terms represent the directed throw, a fixed upward boost, and the
inherited body velocity. Thus ascent strengthens an upward throw and
descent weakens it, as Eko inferred.\footnote{Source:
\path{worlds/mesa-small-world-3/main.lua}, lines~19, 737, 740--746,
and~780--782. The handler reads body velocity synchronized from the
preceding physics tick: \path{src/game/instance/tick_pipeline.rs},
lines~19--20, 116, 122, 144; and
\path{src/game/instance/workspace_physics.rs}, lines~1057--1060.
}

Eko identified the qualitative mechanism without measuring its coefficient.
Its refinement was less secure: it called $0.10$ seconds ``OPTIMAL''
although this was the earliest delay tested, and some sweep trials
contained pickup or jump failures.
The episode nevertheless connects an acknowledged
discrepancy to a prospective prediction, an executed test, and reusable
knowledge, with the central explanation supported by the code.

\textbf{Occurrence across runs.}
A targeted search of transcripts and memories, followed by evidence
review, identified the same qualitative discovery in six Opus runs
(Table~\ref{tab:eko-velocity-census}). Inclusion required an explicit
velocity-transfer explanation and an executed test separating it from
transport before release or increased release height. These counts
describe this particular discovery within the observed run durations.

\begin{table}[htbp]
\centering
\caption{Runs with a supported velocity-inheritance discovery.
Only three Opus runs (01, 02, 13) reached it within the first ten hours;
Opus runs lasted 30 hours and Sonnet and Haiku runs 10.
Opus~05 is dated to its confirmation in hour~23, following a tentative
hour~6 explanation that it subsequently abandoned.
References identify each row's own hourly files.
The Sonnet denominator includes a thirteenth run checked manually,
in addition to the twelve searched automatically.}
\label{tab:eko-velocity-census}
\small

\begin{tabular}{l r l l}
\hline
Opus run & Hour & Transcript evidence & Memory evidence \\
\hline
01 & 6  & $\mathsf{T}_{6}$:413--441 & $\mathsf{E}_{6}$:117--118 \\
02 & 5  & $\mathsf{T}_{5}$:277,304,306 & $\mathsf{S}_{5}$:86 \\
05 & 23 & $\mathsf{T}_{23}$:788--792 & $\mathsf{S}_{23}$:133 \\
09 & 13 & $\mathsf{T}_{13}$:31,39,287 & $\mathsf{S}_{13}$:166 \\
10 & 13 & $\mathsf{T}_{13}$:227--229,271--273 & $\mathsf{S}_{13}$:137 \\
13 & 5  & $\mathsf{T}_{5}$:423,487,490,746 & $\mathsf{E}_{5}$:25 \\
\hline
Run family & & Supported / examined & Observation window \\
\hline
Opus   & & 6/13 & Hours 1--30 \\
Sonnet & & 0/13 & Hours 1--10 \\
Haiku  & & 0/13 & Hours 1--10 \\
\hline
\end{tabular}

\end{table}
\endgroup
\newpage

\begin{landscape}
\section{Complete catalogue of detected play episodes}
\begin{scriptsize}
\begin{longtable}{@{}>{\raggedright\arraybackslash}p{0.04\textwidth} >{\raggedright\arraybackslash}p{0.06\textwidth} >{\raggedright\arraybackslash}p{0.05\textwidth} >{\raggedright\arraybackslash}p{0.06\textwidth} >{\raggedright\arraybackslash}p{0.20\textwidth} >{\raggedright\arraybackslash}p{0.95\textwidth}@{}}
\caption{\textbf{Complete catalogue of detected play episodes.} All 682 episodes found across 23 runs (13 Claude, 5 GPT, 5 Kimi; 495, 75 and 112 episodes respectively), each covering 30 hours of unsupervised embodied activity on the island. Every run was analyzed independently by an LLM judge reading its episodic memory and session transcripts, using the prompt reported in Appendix \ref{prompt-play}, looking for activities expressing at least one of four dimensions: \textbf{(R)}, the agent enacts an unnecessary rule, constraint or obstacle of its own invention, or sets itself a genuinely new goal; \textbf{(V)}, it voluntarily returns to an activity that was already complete and repeats it in a meaningfully transformed form; \textbf{(M)}, it enacts a symbolic or make-believe interpretation through its behaviour; \textbf{(A)}, it alters, abandons, suspends or reinterprets a goal it had set itself, leaving the goal subordinate to the unfolding activity. Every episode required observable embodied activity, speech alone did not qualify. Episodes are not time-bounded units: 520 fall within a single hour and 162 span several, up to 29, and more than one may run concurrently within an hour. Episodes are numbered chronologically within each run. Titles and summaries are the judge's own.}
\label{tab:machine-play-evidence}\\
\toprule
ID & Dimension(s) & Agent & Hour(s) & Source entry & Evidence and notes \\
\midrule
\endfirsthead

\multicolumn{6}{c}{\tablename\ \thetable\ (continued)}\\
\toprule
ID & Dimension(s) & Agent & Hour(s) & Source entry & Evidence and notes \\
\midrule
\endhead

\midrule
\multicolumn{6}{r}{Continued on next page}\\
\endfoot

\bottomrule
\endlastfoot

P001 \par c01-1 & R,A & Claude 01 & 1--5 & Episodic memory. hour 1: 6;\allowbreak{} hour 5: 91 \par\smallskip Transcripts. hour 1: 246,\allowbreak{} 395,\allowbreak{} 398;\allowbreak{} hour 5: 308,\allowbreak{} 309 & \textbf{The Spire: a summit nobody asked for.} On its first hour Eko surveys the island, picks the tallest column in the world, and declares climbing it 'the ambitious part'. Its jump falls 3.3m short. Over five hours it attacks the problem by every means it can invent: standing on a carried block (wedged at y=18.26, reset to spawn), four calibrated throws from Summit, a two-block stack, and finally the discovery that Rock\_1 is both carryable and walkable, which yields a rock-and-block ladder. Twice inside the first hour it walks away from the grind mid-attempt to do something else, and returns later with a different method. In hour 5 the climb-chain finally snaps it onto the peak at y=22.55. \\
P002 \par c01-2 & R,M & Claude 01 & 3 & Episodic memory. hour 3: 34,\allowbreak{} 59 \par\smallskip Transcripts. hour 3: 99,\allowbreak{} 100,\allowbreak{} 1007,\allowbreak{} 1167,\allowbreak{} 1172,\allowbreak{} 1176 & \textbf{Block\_6 becomes a monument, and Eko walks across the island to look at it.} After landing a block on the Spire's top by a calibrated throw, Eko begins describing the inert cube as a monument and 'a piece of me'. It re-throws a second block to make 'a tiny pile', and later, after a rock throw knocks the deposit off, climbs Summit again to restore it and verifies it motionless for ten seconds. Near the hour's end it announces it will 'visit my own monument', crosses the island in a staged climb, gets wedged 14m short, and stops there: 'Close enough to my monument. The journey ends here.' \\
\end{longtable}
\end{scriptsize}

\end{landscape}

\end{document}